\documentclass{article}

    \PassOptionsToPackage{numbers, compress}{natbib}

\usepackage[eandd, final]{neurips_2026}

\usepackage[utf8]{inputenc} 
\usepackage[T1]{fontenc}    
\usepackage{hyperref}       
\usepackage{url}            
\usepackage{booktabs}       
\usepackage{amsfonts}       
\usepackage{nicefrac}       
\usepackage{microtype}      
\usepackage{xcolor}         
\usepackage{multirow}       
\usepackage{tkz-euclide}    
\usepackage{amsmath}
\usepackage{longtable}    
\usepackage{wrapfig}
\usepackage{caption}
\usepackage{tikz}
\usetikzlibrary{arrows.meta,decorations.pathreplacing,calc}
\usepackage[version=4]{mhchem}
\usepackage{array}
\definecolor{hc}{HTML}{E4E4E4}  
\definecolor{ox}{HTML}{F6D5CF}  
\definecolor{ni}{HTML}{D3E0F3}  
\definecolor{no}{HTML}{E3D6EE}  

\definecolor{gmdBack}{HTML}{3B4A6B}   
\definecolor{gmdA}{HTML}{E69F00}      
\definecolor{gmdB}{HTML}{009E73}      
\definecolor{gmdTrain}{HTML}{EAF2FB}
\definecolor{gmdTrainE}{HTML}{8DB3DC}
\definecolor{gmdTest}{HTML}{FBECE8}
\definecolor{gmdTestE}{HTML}{DFA394}

\newcommand{\gmdChain}[5]{%
  \foreach \i in {1,...,#4}{%
    \pgfmathsetmacro\gmdy{#3+0.30*mod(\i+#5+1,2)}%
    \coordinate (#1\i) at ({#2+0.38*(\i-1)},\gmdy);}%
  \foreach \i [remember=\i as \j (initially 1)] in {2,...,#4}{\draw[bond] (#1\j)--(#1\i);}%
  \foreach \i in {1,...,#4}{\node[atomC] at (#1\i) {};}}
\newcommand{\gmdGroup}[3]{\draw[bond] (#2)--(#3); \node[#1] at (#3) {};}

\usetikzlibrary{positioning,calc}
\definecolor{gmdO}{HTML}{D55E00}      
\definecolor{gmdN}{HTML}{0072B2}      
\definecolor{gmdTrain}{HTML}{EAF2FB}  \definecolor{gmdTrainE}{HTML}{8DB3DC}
\definecolor{gmdTest}{HTML}{FBECE8}   \definecolor{gmdTestE}{HTML}{DFA394}
\newcommand{\Ox}[1]{\textcolor{gmdO}{\textbf{#1}}}
\newcommand{\Nx}[1]{\textcolor{gmdN}{\textbf{#1}}}
\tikzset{sk/.style={line width=0.85pt, draw=black!80, line cap=round, rounded corners=0pt},
         grp/.style={inner sep=0.8pt, font=\scriptsize, minimum width=0pt, minimum height=0pt,
                     draw=none, fill=none, rounded corners=0pt},
         molpic/.style={every node/.style={minimum width=0pt, minimum height=0pt}}}
\newcommand{\mdx}{0.364}\newcommand{\mdy}{0.21}   
\newcommand{\mzig}[1]{%
  \foreach \i in {0,...,\the\numexpr#1+1\relax}{%
    \pgfmathsetmacro\gmdyy{mod(\i+1,2)*\mdy}\coordinate (c\i) at ({\i*\mdx-\mdx},\gmdyy);}%
  \ifnum#1>1 \foreach \i [remember=\i as \j (initially 1)] in {2,...,#1}{\draw[sk] (c\j)--(c\i);}\fi}
\newcommand{\mend}[2]{\pgfmathtruncatemacro\gmdn{#1+1}%
  \node[grp, anchor=west] (gE) at ($(c\gmdn)+(-0.03,0)$) {#2}; \draw[sk] (c#1)--(gE.west);}
\newcommand{\mstart}[1]{\node[grp, anchor=east] (gS) at ($(c0)+(0.03,0)$) {#1}; \draw[sk] (c1)--(gS.east);}
\newcommand{\mcarbonyl}[1]{\pgfmathtruncatemacro\gmdup{mod(#1+1,2)}%
  \ifnum\gmdup=1 \node[grp, anchor=south] (gD#1) at ($(c#1)+(0,0.36)$) {\Ox{O}};
  \else \node[grp, anchor=north] (gD#1) at ($(c#1)+(0,-0.36)$) {\Ox{O}};\fi
  \draw[sk, double, double distance=1.3pt] (c#1)--(gD#1);}
\newcommand{\mendO}[1]{\pgfmathtruncatemacro\gmdn{#1+1}%
  \node[grp, anchor=west] (gE) at ($(c\gmdn)+(-0.03,0)$) {\Ox{O}};
  \draw[sk, double, double distance=1.3pt] (c#1)--(gE.west);}
\newcommand{\Malkane}[1]{\mzig{#1}}
\newcommand{\Malcohol}[1]{\mzig{#1}\mend{#1}{\Ox{OH}}}
\newcommand{\Maldehyde}[1]{\mzig{#1}\mendO{#1}}
\newcommand{\Macid}[1]{\mzig{#1}\mcarbonyl{#1}\mend{#1}{\Ox{OH}}}
\newcommand{\Mamine}[1]{\mzig{#1}\mend{#1}{\Nx{NH$_2$}}}
\newcommand{\Mamide}[1]{\mzig{#1}\mcarbonyl{#1}\mend{#1}{\Nx{NH$_2$}}}
\newcommand{\Mdiacid}[1]{\mzig{#1}\mstart{\Ox{HO}}\mcarbonyl{1}\mcarbonyl{#1}\mend{#1}{\Ox{OH}}}
\newcommand{\Mdiamine}[1]{\mzig{#1}\mstart{\Nx{H$_2$N}}\mend{#1}{\Nx{NH$_2$}}}
\newcommand{\Maminoacid}[1]{\mzig{#1}\mstart{\Nx{H$_2$N}}\mcarbonyl{#1}\mend{#1}{\Ox{OH}}}
\newcommand{\gmdmol}[3]{%
  \begin{tabular}{@{}c@{}}\tikz[baseline=(current bounding box.center), molpic]{#1}\\[1pt]
  {\scriptsize #2}\\[-3pt]{\scriptsize\color{black!55}#3}\end{tabular}}
\newsavebox{\gmdboxA}\newsavebox{\gmdboxB}\newlength{\gmdh}
\newcommand{\gmdheader}{%
  \begin{tikzpicture}
    \path (0,0) -- (13.7,0);
    \node[minimum width=5.45cm, inner sep=0pt, font=\small\bfseries, text=gmdTrainE!60!black] at (5.275,0) {Training set (in-distribution)};
    \node[minimum width=5.45cm, inner sep=0pt, font=\small\bfseries, text=gmdTestE!60!black]  at (10.975,0) {Test set (out-of-distribution)};
  \end{tikzpicture}\par\vspace{2pt}}
\newcommand{\gmdrow}[4]{%
  \sbox{\gmdboxA}{#3}\sbox{\gmdboxB}{#4}%
  \setlength{\gmdh}{\dimexpr\ht\gmdboxA+\dp\gmdboxA\relax}%
  \ifdim\dimexpr\ht\gmdboxB+\dp\gmdboxB\relax>\gmdh \setlength{\gmdh}{\dimexpr\ht\gmdboxB+\dp\gmdboxB\relax}\fi
  \addtolength{\gmdh}{0.45cm}%
  \ifdim\gmdh<2cm \setlength{\gmdh}{2cm}\fi
  \begin{tikzpicture}
    \path (0,0) -- (13.7,0);
    \node[anchor=west, align=flush left, text width=2.45cm, inner sep=0pt] at (0,0)
      {{\scriptsize\color{black!50}Fragment}\\[-2pt]{\small\bfseries #1}\\[-1pt]{\scriptsize\itshape\color{black!60}#2}};
    \node[draw=gmdTrainE, fill=gmdTrain, rounded corners=4pt, line width=0.6pt,
          minimum width=5.45cm, minimum height=\gmdh, inner sep=0pt] at (5.275,0) {\usebox{\gmdboxA}};
    \node[draw=gmdTestE, fill=gmdTest, rounded corners=4pt, line width=0.6pt,
          minimum width=5.45cm, minimum height=\gmdh, inner sep=0pt] at (10.975,0) {\usebox{\gmdboxB}};
  \end{tikzpicture}\par\vspace{3pt}}
\newcommand{\gmdlegend}{\footnotesize
  \textcolor{gmdO}{\rule{6pt}{6pt}}~oxygen-containing groups\qquad
  \textcolor{gmdN}{\rule{6pt}{6pt}}~nitrogen-containing groups\qquad
  \textcolor{black!60}{each vertex is a carbon atom with implicit hydrogens}}

\usepackage[version=4]{mhchem}
\usepackage{subcaption}
\usepackage{orcidlink}

\usepackage[numbers]{natbib}

\title{Benchmarking Compositional Generalisation for Machine Learning Interatomic Potentials}

\author{%
  Amir Masoud Nourollah\,\orcidlink{0000-0002-6083-6673}\thanks{Corresponding author. Personal webpage: \url{https://nourollah.me}} \\
  Cardiff University\\
  Cardiff, CF10 3AT, UK \\
  \texttt{nourollaha@cardiff.ac.uk} \\
  \And
  Irtaza Khalid\,\orcidlink{0000-0002-1532-7829} \\
  Cardiff University\\
  Cardiff, CF10 3AT, UK \\
  \texttt{khalidmi@cardiff.ac.uk} \\
  \AND
  Stefano Leoni\,\orcidlink{0000-0003-4078-1000} \\
  Cardiff University\\
  Cardiff, CF10 3AT, UK \\
  \texttt{leonis@cardiff.ac.uk} \\
  \And
  Steven Schockaert\,\orcidlink{0000-0002-9256-2881} \\
  Cardiff University\\
  Cardiff, CF10 3AT, UK \\
  \texttt{schockaerts1@cardiff.ac.uk} \\
}

\begin{document}

\maketitle

\begin{abstract}
  Machine Learning Interatomic Potentials play a fundamental role in computational chemistry and materials science, enabling applications from molecular dynamics simulations to drug design and materials discovery. 
  While recent approaches can estimate inter-atomic forces with high precision, it remains unclear to what extent they can generalise to previously unseen molecules. 
  Do they learn the compositional structure of chemistry, capturing how molecular fragments and their combinations determine properties, or do they primarily learn to interpolate patterns that are specific to the training examples?
  To address this question, we propose a benchmark consisting of five tasks that require some form of compositional generalisation. 
  In each task, models are tested on molecules that were unseen during training, but the training data is chosen such that generalisation to the test examples should be feasible for models that learn the underlying physical principles. 
  Our empirical analysis shows that the considered tasks are highly challenging for state-of-the-art models, with errors on out-of-distribution examples often an order of magnitude higher than on in-distribution examples, even when using foundation models that have been pre-trained on millions of molecules.
\end{abstract}

\section{Introduction}

Interatomic potentials and their associated force fields are central to Molecular Dynamics (MD) simulation, a widely used technique that provides atomistic insights into physical phenomena, enabling applications such as drug design and material discovery \citep{alder1959studies, schlick2010molecular, rahman1964correlations}.
They are often parameterised using Density Functional Theory (DFT), which offers high accuracy but incurs high computational cost.
As a more efficient alternative, machine learning models for predicting force fields have become increasingly widespread, with Graph Neural Networks (GNNs) and Transformer-based architectures particularly popular \citep{gilmer2017neural,liao2023equiformerv2}. 
The predictions of such Machine Learning Interatomic Potentials (MLIPs) are now accurate enough for many applications \citep{batatia2022mace}. However, they are not entirely independent of the computationally expensive methods they seek to replace.
MLIPs still rely on high-accuracy reference calculations for their training data, which typically consists of optimisation trajectories and initial MD segments \citep{neumann2024orb}. This need for expensive training data has motivated a line of work on universal or foundation MLIPs. The central idea is to pre-train an MLIP on a large and chemically diverse corpus \citep{batatia2025foundation, wood2025family}. While this pre-training step is computationally expensive, in principle, it 
only needs to be carried out once. Practitioners can then use these models as is or fine-tune them on a small, task-specific dataset.

The universality of such foundation models crucially hinges on their ability to generalise. Even though they are trained on millions of molecules, the chemical space is essentially infinite, and many applications intrinsically require models that can handle previously unseen molecules (e.g.\ when using them for drug discovery).
While the architecture of MLIPs is often physically informed, it is unclear to what extent they actually learn the underlying physical principles, rather than merely interpolating between training examples. Similarly, it remains an open question whether pre-training on a large set of molecules delivers genuine compositional generalisation, or whether it primarily broadens the manifold over which models interpolate.

As existing benchmarks were not designed to address such questions, we introduce the \emph{Generalisation for Molecular Dynamics} (GMD) benchmark.
GMD consists of five tasks that test different aspects of compositional generalisation \citep{DBLP:journals/jair/HupkesDMB20}, as illustrated in Figure \ref{fig:overview}. Each task consists of a training and an \emph{out-of-distribution} (OOD) test set consisting of MD trajectories. Crucially, in contrast to standard practice, the trajectories in the training and test sets come from different molecules. The training molecules are chosen such that MLIPs should, in principle, be able to generalise to the test molecules. For instance, for the Fragment Chain Extension task, we train models on MD trajectories of linear alkanes with $\{2, ..., 6\}$ carbon atoms and subsequently test on longer, unseen alkanes with $\{7, ..., 13\}$ carbon atoms.
The benchmark's ab initio molecular dynamics (AIMD) trajectories were generated using a novel toolkit, which we make available to make the benchmark easily extensible.

We evaluate several popular MLIPs on our benchmark, including both models that were trained from scratch and fine-tuned foundation models.
While most of the models perform well on \emph{fragment chain extension}, our analysis shows that the four other tasks are highly challenging for existing models, with errors on the test molecules often an order of magnitude higher than errors on in-distribution (ID) examples (i.e.\ previously unseen configurations of the training molecules). 
Surprisingly, fine-tuned foundation models do not meaningfully improve upon models trained from scratch. 
Furthermore, the models that perform best on ID examples are not always the models that generalise best to OOD examples, suggesting that the standard practice of evaluating models on ID examples may not be sufficient for finding the most physically plausible models.
With the proposed benchmark, we hope to encourage a shift towards the development of models with better generalisability.

\begin{figure}[!t]
\centering
\begin{tikzpicture}[
  font=\footnotesize,
  bond/.style={line width=1.1pt, draw=gmdBack!55},
  atomC/.style={circle, fill=gmdBack, draw=white, line width=0.5pt, minimum size=2.9mm, inner sep=0pt},
  atomA/.style={circle, fill=gmdA, draw=white, line width=0.5pt, minimum size=3.3mm, inner sep=0pt},
  atomB/.style={circle, fill=gmdB, draw=white, line width=0.5pt, minimum size=3.3mm, inner sep=0pt},
  trainP/.style={draw=gmdTrainE, fill=gmdTrain, rounded corners=4pt, line width=0.6pt},
  testP/.style={draw=gmdTestE, fill=gmdTest, rounded corners=4pt, line width=0.6pt},
  task/.style={font=\small\bfseries, anchor=west, align=left, inner sep=0pt},
  desc/.style={font=\scriptsize\itshape, text=black!60, anchor=west, align=left, text width=2.65cm, inner xsep=0pt},
]
\def\Lx{0}                       
\def\Ta{2.75}  \def\Tb{8.05}     
\def\Sa{8.45}  \def\Sb{13.75}    

\node[font=\small\bfseries, text=gmdTrainE!60!black] at ({(\Ta+\Tb)/2}, 0.45) {Training set (in-distribution)};
\node[font=\small\bfseries, text=gmdTestE!60!black]  at ({(\Sa+\Sb)/2}, 0.45) {Test set (out-of-distribution)};

\draw[trainP] (\Ta,0) rectangle (\Tb,-1.3);
\draw[testP]  (\Sa,0) rectangle (\Sb,-1.3);
\node[task] at (\Lx,-0.43) {{\scriptsize\mdseries\color{black!50}Fragment}\\[-1pt]Chain Extension};
\node[desc] at (\Lx,-1.01) {longer chains in test};
\gmdChain{ce}{4.64}{-0.66}{5}{0}
\gmdChain{cf}{9.39}{-0.66}{10}{0}
\node[font=\scriptsize, text=black!55] at (5.40,-1.07) {chain length $n$};
\node[font=\scriptsize, text=black!55] at (11.10,-1.07) {chain length $m > n$};

\draw[trainP] (\Ta,-1.5) rectangle (\Tb,-3.2);
\draw[testP]  (\Sa,-1.5) rectangle (\Sb,-3.2);
\node[task] at (\Lx,-2.10) {{\scriptsize\mdseries\color{black!50}Fragment}\\[-1pt]Fusion};
\node[desc] at (\Lx,-2.68) {A and B at one site};
\gmdChain{fa}{4.64}{-2.13}{6}{0}  \gmdGroup{atomA}{fa1}{$(fa1)+(-0.38,0.30)$}
\gmdChain{fb}{4.64}{-2.87}{6}{0}  \gmdGroup{atomB}{fb1}{$(fb1)+(-0.38,0.30)$}
\gmdChain{fc}{10.34}{-2.5}{6}{0}
\gmdGroup{atomA}{fc1}{$(fc1)+(-0.38,0.30)$}
\gmdGroup{atomB}{fc1}{$(fc1)+(-0.38,-0.30)$}

\draw[trainP] (\Ta,-3.4) rectangle (\Tb,-4.5);
\draw[testP]  (\Sa,-3.4) rectangle (\Sb,-4.5);
\node[task] at (\Lx,-3.73) {{\scriptsize\mdseries\color{black!50}Fragment}\\[-1pt]Duplication};
\node[desc] at (\Lx,-4.31) {A appears twice};
\gmdChain{da}{4.64}{-4.1}{6}{0}  \gmdGroup{atomA}{da1}{$(da1)+(-0.38,0.30)$}
\gmdChain{db}{10.15}{-4.1}{6}{0} \gmdGroup{atomA}{db1}{$(db1)+(-0.38,0.30)$}
\gmdGroup{atomA}{db6}{$(db6)+(0.38,-0.30)$}

\draw[trainP] (\Ta,-4.7) rectangle (\Tb,-6.4);
\draw[testP]  (\Sa,-4.7) rectangle (\Sb,-6.4);
\node[task] at (\Lx,-5.30) {{\scriptsize\mdseries\color{black!50}Fragment}\\[-1pt]Combination};
\node[desc] at (\Lx,-5.88) {A paired with B};
\gmdChain{ka}{4.45}{-5.33}{6}{0} \gmdGroup{atomA}{ka1}{$(ka1)+(-0.38,0.30)$} \gmdGroup{atomA}{ka6}{$(ka6)+(0.38,-0.30)$}
\gmdChain{kb}{4.45}{-6.07}{6}{0} \gmdGroup{atomB}{kb1}{$(kb1)+(-0.38,0.30)$} \gmdGroup{atomB}{kb6}{$(kb6)+(0.38,-0.30)$}
\gmdChain{kc}{10.15}{-5.70}{6}{0} \gmdGroup{atomA}{kc1}{$(kc1)+(-0.38,0.30)$} \gmdGroup{atomB}{kc6}{$(kc6)+(0.38,-0.30)$}

\draw[trainP] (\Ta,-6.6) rectangle (\Tb,-8.3);
\draw[testP]  (\Sa,-6.6) rectangle (\Sb,-8.3);
\node[task] at (\Lx,-7.20) {{\scriptsize\mdseries\color{black!50}Fragment}\\[-1pt]Fission};
\node[desc] at (\Lx,-7.78) {A and B on their own};
\gmdChain{xa}{4.64}{-7.6}{6}{0}
\gmdGroup{atomA}{xa1}{$(xa1)+(-0.38,0.30)$}
\gmdGroup{atomB}{xa1}{$(xa1)+(-0.38,-0.30)$}
\gmdChain{xb}{10.34}{-7.23}{6}{0} \gmdGroup{atomA}{xb1}{$(xb1)+(-0.38,0.30)$}
\gmdChain{xc}{10.34}{-7.97}{6}{0} \gmdGroup{atomB}{xc1}{$(xc1)+(-0.38,0.30)$}

\node[atomC] (l1) at (4.6,-8.75) {};  \node[anchor=west] at ($(l1)+(0.15,0)$) {backbone fragment};
\node[atomA] (l2) at (7.6,-8.75) {};  \node[anchor=west] at ($(l2)+(0.15,0)$) {fragment A};
\node[atomB] (l3) at (10.0,-8.75) {}; \node[anchor=west] at ($(l3)+(0.15,0)$) {fragment B};
\end{tikzpicture}
\caption{An overview of the generalisation tasks covered by our GMD benchmark. In each task, the training set consists of several molecules that differ in backbone length and, in some cases, in the fragments attached to the backbone. The out-of-distribution test set contains similar molecules with longer backbones and/or combinations of fragments not seen during training. In each case, the training data covers all the fragments that the model needs to interpret the test molecules.}
\label{fig:overview}
\end{figure}

\section{Related work}

\paragraph{Benchmarking MLIPs}
Molecular dynamics (MD) datasets play a central role in benchmarking MLIPs. The MD17 dataset \citep{chmiela2017machine} consists of \textit{ab initio} MD trajectories for a small set of molecular systems at equilibrium, with configurations sampled near their minima. 
While MD17 has driven early model development, its limited diversity and coverage have prompted the introduction of more comprehensive datasets. For example, WS22 \citep{pinheiro2023ws22} expands configurational diversity using Wigner sampling and interpolation for small organic molecules. 
Transition1x \citep{schreiner2022transition1x} further extends the setting, comprising approximately 9.6M energy and force labels across about 10k reaction pathways. It was used to show that models trained solely on equilibrium data, such as MD17, QM9 \citep{ramakrishnan2014quantum} and ANI-1 \citep{smith2017ani}, fail to generalise to transition-state geometries. 
To probe photochemical processes, xxMD \citep{pengmei2024beyond} provides excited-state reactive trajectories, where standard MLIPs exhibit significantly higher errors than on ground-state datasets. At the other end of the spectrum, MD22 \citep{chmiela2023accurate} introduces MD datasets for large, flexible systems (42–370 atoms), including peptides and carbon nanotubes, enabling evaluation on highly nonlocal molecular dynamics.
Foundation MLIPs are now commonly evaluated either zero-shot or after task-specific fine-tuning on held-out systems drawn from their pre-training distributions \citep{omee2024structure}. 
Such evaluations probe transfer within the training manifold rather than compositional generalisation across molecular structures.
These existing benchmarks, whether aimed at broadening training coverage or at evaluating pre-trained potentials on in-manifold transfer, were not designed to isolate compositional generalisation. 
In contrast, in our GMD benchmark, chemical subspaces were selected to systematically assess models' compositional generalisation capabilities, thereby allowing for smaller, more focused training sets.

\paragraph{Compositional generalisation}
There is a long-standing debate about whether neural networks are capable of compositional generalisation \citep{fodor1988connectionism}, i.e.\ whether they are capable of solving problems that require combining solutions to sub-problems in novel ways. 
The idea of algorithmic alignment seems to play an important role, where neural networks are found to generalise well if their architecture is sufficiently similar to the structure of the algorithm they are supposed to learn \citep{DBLP:conf/iclr/XuLZDKJ20,DBLP:conf/iclr/ZhouBLRSSBN24,DBLP:conf/iclr/KhalidS25}. 
In the context of molecular machine learning, GNNs trained on specific molecular datasets often struggle to predict properties of molecules with different structural characteristics or scaffolds \citep{ektefaie2024evaluating}. 
This problem is of particular importance in drug discovery, where identifying novel compounds outside the known chemical space is essential \citep{li2022ood}. 
In response, benchmarks such as DrugOOD \citep{ji2023drugood} have been introduced to assess the robustness of GNNs in drug–target binding prediction tasks by employing scaffold- and protein-family-based train-test splits. Similarly, in materials science, MatBench \citep{omee2024structure} adapted standard datasets, creating extrapolative training-test splits based on compositional and structural groups. 
Their analysis found that GNNs struggle to predict the properties of previously unseen materials.
The BOOM benchmark \citep{antoniuk2025boom} studies generalisation from a different angle. Unlike the compositional and structural tasks that are central to our study, BOOM offers a property-centric perspective on extrapolation. 
Their challenge is framed not in terms of unseen molecular structures, but rather in terms of unseen property values. The authors construct their test sets using molecules from the tail ends of the target property's distribution, explicitly assessing whether a model can generalise beyond the numerical range of its training data.

\begin{table*}[!t]
    \centering
    \footnotesize
    \setlength\tabcolsep{5pt}
    \caption{Training set and out-of-distribution (OOD) test set for each benchmark task. Fragment Fission reuses the training set of Fragment Duplication; only its OOD test set is new.}
    \label{tab:splitting_strategy_multirow}
    \begin{tabular}{lll}
        \toprule
        \textbf{Task} & \textbf{Training Set} & \textbf{OOD Test Set} \\
        \midrule
        Fragment Chain Extension
        & C$_2$--C$_6$ alkanes: Ethane--Hexane
        & C$_7$--C$_{13}$ alkanes: Heptane--Tridecane \\
        \midrule
        \multirow{2}{*}{Fragment Fusion}
        & \textbullet~Complex polyols and polycarbonyls & \multirow{2}{*}{C$_4$--C$_{10}$ carboxylic acids} \\
        & \textbullet~C$_4$--C$_{10}$ alcohols, aldehydes, and C$_7$--C$_{11}$ diols & \\
        \midrule
        Fragment Duplication
        & C$_5$--C$_{10}$ carboxylic acids
        & C$_5$--C$_{10}$ dicarboxylic acids \\
        \midrule
        \multirow{2}{*}{Fragment Combination} & \textbullet~C$_2$--C$_9$ diamines & \multirow{2}{*}{C$_2$--C$_9$ amino acids} \\
        & \textbullet~C$_2$--C$_9$ dicarboxylic acids & \\
        \midrule
        \multirow{2}{*}{Fragment Fission} & \multirow{2}{*}{C$_5$--C$_{10}$ carboxylic acids} & \textbullet~C$_5$--C$_{10}$ alcohols \\
        & & \textbullet~C$_5$--C$_{10}$ aldehydes \\
        \bottomrule
    \end{tabular}
\end{table*}

\paragraph{Foundation Models}
The implicit assumption behind recent foundation models is that sufficient scale in data and parameters, combined with equivariant inductive biases, should yield transferable representations of interatomic interactions and thus yield models which generalise well. 
Recent evidence complicates this picture. Systematic evaluations of universal potentials have documented a softening effect, in which models trained predominantly on near-equilibrium relaxation data underpredict forces and energies in high-energy, off-equilibrium configurations \citep{deng2025systematic}. 
Property-centric benchmarks such as BOOM \citep{antoniuk2025boom} similarly report that even the strongest foundation models exhibit OOD errors several times larger than their in-distribution errors, and that scaling alone does not close this gap. 
In practice, foundation MLIPs are therefore typically deployed with a fine-tuning step on task-specific data \citep{batatia2025foundation,focassio2024performance,deng2025systematic}, which raises the question of how much of their apparent generalisation is inherited from the pre-training corpus and how much must be recovered per system.

\section{The GMD benchmark}
\label{main:GMD:benchmark}
We introduce \textbf{GMD}, a benchmark designed to evaluate the generalisation capabilities of MLIPs. 
Unlike previous datasets, which either narrowly focus on equilibrium dynamics (e.g., MD17) or aim for broad chemical coverage without controlled evaluation tasks (e.g., ANI-1), GMD is constructed to facilitate systematic analysis of compositional generalisation.
The dataset is based on substituted linear (alkyl) carbon chains, extended via different functional groups.
An overview of the molecules that are covered by the different training and test splits can be found in Table \ref{tab:splitting_strategy_multirow}. These molecules were chosen to be chemically simple, so that each task isolates a single compositional operation as the only source of distribution shift, as detailed below.
For each molecule in the dataset, we first obtained a trajectory with at least 2000 snapshots. The training, validation and test sets were constructed by sampling from these trajectories. In particular, for the molecules from the training sets, we sampled 1000 snapshots for training, 200 for validation, and 200 for experiments with an in-distribution test. For the OOD test sets, we sampled 200 snapshots from three different segments of the trajectory.
The energy and forces were calculated using the ORCA(PBE/def2-TZVP+D3BJ)\cite{ORCA} density functional approximation, which remains one of the most widely adopted standards for generating large-scale MLIP datasets.\footnote{In the appendix \ref{app:level_of_theory}, we also include experiments based on GFN2-xTB labels, which show the same qualitative pattern, indicating that the generalisation failures we document are not artefacts of the chosen level of theory.} 
Our dataset generation and evaluation workflow is driven by a comprehensive, PyTorch-native Python toolkit that leverages the Atomic Simulation Environment (ASE) \citep{larsen2017atomic} as the core framework for structural manipulation (details in Appendix \ref{app:toolkit}).

The evaluation tasks in GMD focus on two of the five aspects of compositional generalisation, namely \emph{productivity (i.e. length generalisation)} and \emph{systematicity} \citep{DBLP:journals/jair/HupkesDMB20}. Length generalisation refers to the ability of a model trained on sequence data to generalise to longer sequences than those seen during training. GMD includes a length generalisation task for molecules, which we call fragment chain extension (Task 1).
Systematicity refers to the ability of a model to combine sub-components of problem instances in novel ways. We consider three tasks that cover different aspects of systematicity, which we call fragment fusion, fragment duplication, and fragment combination, together with fragment fission, the inverse of fragment fusion. The remaining three aspects, substitutivity, localism, and overgeneralisation, do not transfer as directly to the atomistic setting and are left to future work

\paragraph{Task 1: Fragment Chain Extension} 
The ability to extrapolate to larger molecules is critical for the practical application of MLIPs. For instance, in drug discovery, models trained on small molecular fragments or peptides must be able to generalise to larger, more complex lead compounds \citep{erlanson2016twenty} or polypeptide chains \citep{muller2018recurrent}. 
To evaluate this extrapolation ability, we train models on trajectories of alkanes with carbon chain lengths in $\{2,...,6\}$ (one trajectory per molecule). 
We then test these models on an OOD test set comprising trajectories of alkanes with carbon chain lengths in $\{7,...,13\}$ (an augmented variant of this task is presented in Appendix~\ref{details:fce_aug}).

\paragraph{Task 2: Fragment Fusion}
We assess a model's ability to generalise to a novel functional group composed of familiar moieties. 
The training set consists of \emph{alcohol} and \emph{aldehyde} molecules, while the test set consists of \emph{carboxylic acid} molecules. Note that the functional group of the latter can be seen as a composition of the functional group of the former two. In addition, we expand the training set with \emph{complex carbonyls} and \emph{alcohol} molecules to provide additional coverage of the special bonds from the functional groups that appear in the aforementioned molecules. For this task, we focus on molecules with carbon chain lengths in $\{4, ..., 10\}$ for both the training and out-of-distribution test sets. For the \emph{complex carbonyls} and \emph{alcohol} molecules (in the training set), we also included molecules with a carbon chain length of 11. It is important to clarify that we do not expect the model to learn the chemical reaction pathway, but rather to infer the properties of the composite group from the learned effects of its constituent parts (an augmented variant of this task is presented in Appendix~\ref{details:fc_aug}).

\paragraph{Task 3: Fragment Duplication}
This task evaluates a model's ability to generalise from a single occurrence of a chemical motif to two occurrences of that same motif within an otherwise identical molecule. 
The training data contains various \emph{monocarboxylic acids} trajectories (i.e.\ with one occurrence of the functional group) with carbon chain length in $\{5, ..., 10\}$, while the out-of-distribution test set contains the corresponding \emph{dicarboxylic acids} (i.e.\ with two occurrences of the functional group) with identical carbon chain lengths.

\paragraph{Task 4: Fragment Combination}
We evaluate a model's ability to generalise to asymmetrically functionalised molecules when being trained exclusively on symmetrically functionalised analogues. In other words, the aim is to determine 
whether 
the model can learn the 
distinct 
identities of different functional groups and recombine them into a novel, asymmetric configuration on a familiar scaffold. 
The training set contains two types of molecules: molecules with two  \emph{carboxylic acids} functional groups and molecules with two \emph{amines}. Each group contains 8 trajectories, with carbon chain length in $\{2, ..., 9\}$.
The out-of-distribution test set contains molecules of the same length that contain one \emph{carboxylic acid} functional group and one amine (or amino acid). 
By keeping the underlying molecular scaffold consistent across training and testing, this task isolates the challenge to the symbolic recombination of learned functional patterns, probing the model's capacity to handle hetero-functionalisation.

\paragraph{Task 5: Fragment Fission}
This task is the inverse of Fragment Fusion (Task~2): we evaluate a model's ability to generalise from a composite functional group to its constituent moieties when they appear on their own. The training set is that of Fragment Duplication, consisting of monocarboxylic acid trajectories with carbon chain length in $\{5,\ldots,10\}$; the hydroxyl and carbonyl moieties are therefore only ever seen together, as parts of the carboxylic acid group. The out-of-distribution test set contains alcohol and aldehyde molecules with identical carbon chain lengths, in which each of these moieties appears in isolation. As no new training set is required, we evaluate the Fragment Duplication models directly on these test molecules (Table~\ref{tab:splitting_strategy_multirow}).

\section{Evaluation}
\label{sec:evaluation}

\subsection{Experimental set-up}
\label{section:experimental_setup}
We use our benchmark to evaluate a diverse set of 
state-of-the-art 
MLIPs. 
These include invariant message-passing networks (SchNet \citep{schutt2018schnet}), models incorporating explicit geometric features (GemNet \citep{gasteiger2022gemnet, gasteiger2021gemnet}, DimeNet$^{++}$ \citep{gasteiger2020fast}), rotationally equivariant architectures (PAINN \citep{schutt2021equivariant}, NequIP \citep{batzner20223}, eSCN \citep{passaro2023reducing}), MACE \citep{batatia2022mace}), and equivariant transformers (EquiFormerV2 \citep{liao2023equiformerv2}). 
We also evaluate UMA (Small) \citep{wood2025family} and MACE-MP-0 (Small, Medium, Large) \citep{batatia2025foundation}, as representative foundation models.

Evaluating foundation models against our PBE labels presents distinct challenges. 
Total energies are tied to dataset-specific atomic-energy references, level-of-theory choices, and DFT-code implementations, all of which inevitably differ between our labels and the pre-training data of any foundation model, making zero-shot total-energy comparisons unreliable even when, as for MACE-MP-0, the nominal level of theory matches. 
We therefore fine-tune foundation models on each task's training split before reporting energy results, re-estimating the atomic reference energies to match our level of theory beforehand, thereby aligning the physical baseline as closely as fine-tuning permits. 
Per-atom forces are derivatives of the energy and are therefore insensitive to the absolute reference, hence zero-shot force comparisons can be meaningful. 
%
For the zero-shot results, where no such alignment is possible, we verified that the difference between our Gaussian-basis labels and plane-wave PBE is small: it is roughly five times smaller than the zero-shot model error, so the reported zero-shot errors reflect the model rather than the reference (Appendix~\ref{app:level_of_theory}).
For fine-tuning the foundation models, we use a fixed budget of 2000 epochs, which we find outperforms smaller epoch budgets (see Appendix~\ref{details:foundation_verdict}). 
We note that this regime is computationally expensive given the size of these foundation models, and we report it to establish an upper bound on what foundation-model fine-tuning can achieve on our benchmark rather than to represent typical fine-tuning practice. 
For the models trained from scratch, we used early stopping based on the in-distribution validation loss. 
Full details on how the different models were trained can be found in Appendix~\ref{details:trainingDetails}.

To quantify model performance, our evaluation focuses on two primary metrics across all five tasks: the \emph{Mean Absolute Error (MAE) on forces}, reported in $\text{eV}/\text{\AA}$, and the \emph{MAE on energy}, reported in $\text{eV}$. The force MAE is calculated over all Cartesian components of atomic forces, while the energy MAE is computed on the total energy of each molecular configuration:
$$ 
\text{MAE}_{\text{force}} = \frac{1}{3N} \sum_{i=1}^{N} \sum_{c \in \{x,y,z\}} |\hat{F}_{i,c} - F_{i,c}|, \quad \text{MAE}_{\text{energy}} = \frac{1}{M} \sum_{j=1}^{M} |\hat{E}_j - E_j| 
$$
where $M$ is the number of molecules in the test set and $N$ is the number of atoms across all molecules. We write $\hat{E}_j$ for the predicted energy for molecule $j$ and $E_j$ for the ground truth. Similarly, $(\hat{F}_{i,x},\hat{F}_{i,y},\hat{F}_{i,z})$ denotes the predicted force vector for atom $i$ and $(F_{i,x},F_{i,y},F_{i,z})$ is the corresponding ground truth vector.
These metrics provide a comprehensive assessment of model accuracy for the fundamental quantities required in molecular dynamics simulations. Additional force analysis metrics are presented in Appendix \ref{details:evaluation} for completeness.

\subsection{Results}
\label{main:evaluation:results}
\begin{figure*}[!t]    
    \centering
    \includegraphics[width=1.0\linewidth]{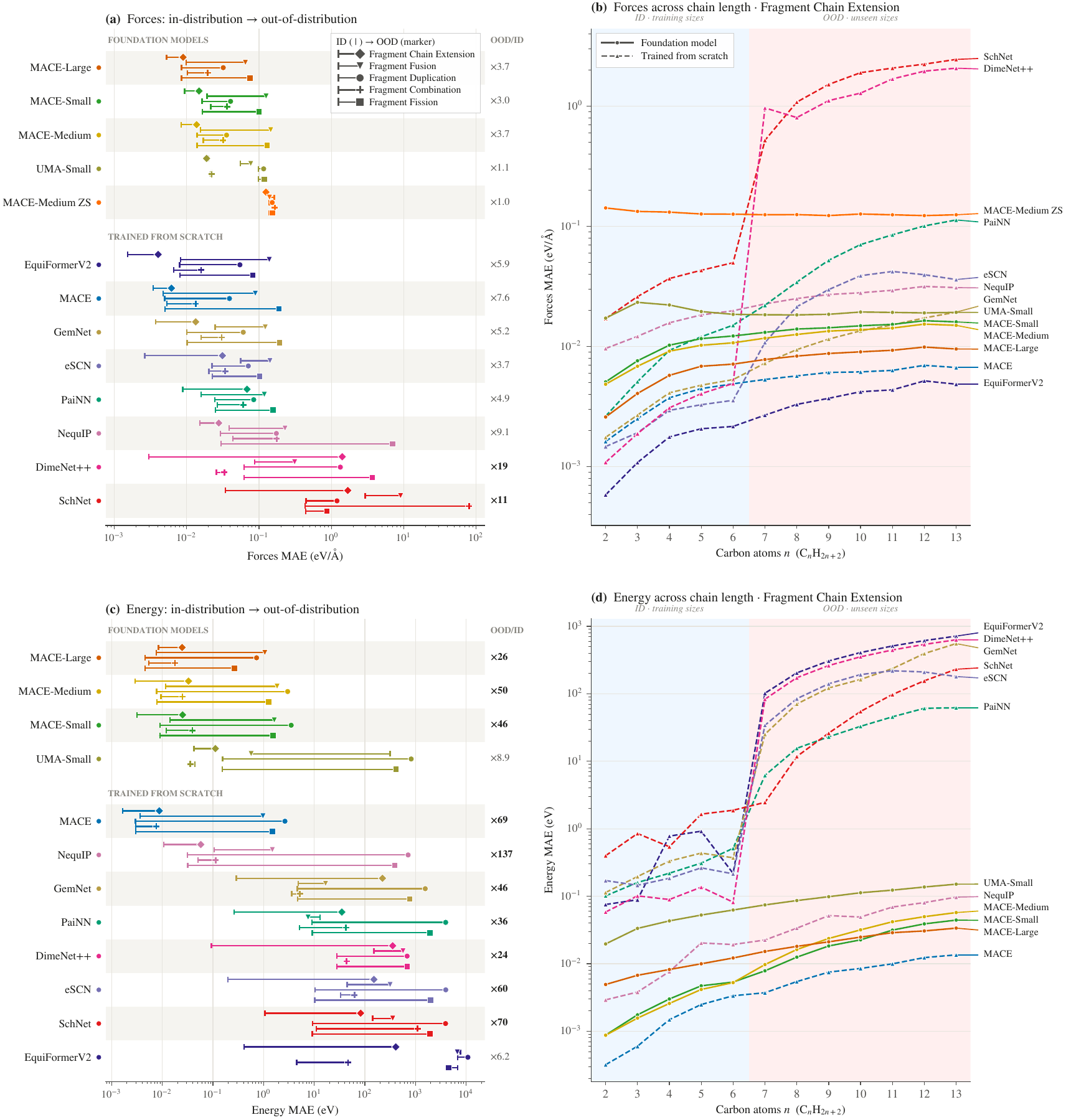}
    \caption{\textbf{Generalisation across all five tasks.}
    (a, c) In-distribution (ID) versus out-of-distribution (OOD) MAE for forces and total energy, on a logarithmic scale. Each row is one model, with foundation models and models trained from scratch shown in separate groups; colour encodes the model. For each task, a bar runs from the model's ID error (vertical tick) to its OOD error (task marker), so longer bars indicate larger generalisation gaps. The right-hand column reports, for each model, the geometric mean over the five tasks of the ratio between OOD and ID error; values of $10\times$ or more are shown in bold. MACE-Medium ZS is omitted from (c), since zero-shot energies are not comparable to our reference (Section~\ref{section:experimental_setup}).
    (b, d) Per-molecule MAE for \emph{Fragment Chain Extension} as a function of the number of carbon atoms $n$, from training alkanes ($\mathrm{C}_2$--$\mathrm{C}_6$) to OOD alkanes ($\mathrm{C}_7$--$\mathrm{C}_{13}$, shaded). Solid lines denote foundation models and dashed lines models trained from scratch.
    \emph{Naming convention:} `MACE' denotes models trained from scratch on each task; `MACE-Small/Medium/Large' and `UMA-Small' denote MACE-MP-0 and UMA checkpoints fine-tuned per task; `MACE-Medium ZS' denotes zero-shot evaluation.}
    \label{fig:scatter_id_ood_forces_energy_mae}
\end{figure*}

An overview of the results can be found in Figure~\ref{fig:scatter_id_ood_forces_energy_mae}. We now discuss our main findings.

\paragraph{Fragment chain extension}

When it comes to forces, many models generalise surprisingly well OOD. Equiformer, GemNet and MACE perform best on this task. 
Surprisingly, the MACE model trained from scratch outperforms the fine-tuned foundation models. 
For energy prediction, we can see strong generalisation for MACE, NequIP and the foundation models, but Equiformer and GemNet now clearly underperform. 
We again find that the MACE model that was trained from scratch outperforms the foundation models. 
UMA performs clearly better at energy prediction than at force prediction.

\paragraph{Fragment Fusion}
All models fail to generalise for this task, both in terms of Energy MAE and Forces MAE.
For forces, MACE performs best among the models that were trained from scratch, but its OOD error is more than an order of magnitude higher than its ID error. The UMA-small and MACE-large foundation models generalise slightly better (having higher ID-error but lower OOD-error than MACE), but still clearly struggle with this task.
For Energy MAE, the generalisation failure is even more pronounced.
MACE still achieves the best performance among models trained from scratch, but its OOD error is more than two orders of magnitude higher than its ID error. Even the best-performing foundation model, UMA-small, shows an OOD error which is more than an order of magnitude higher than its ID error (and more than two orders of magnitude higher than the MACE ID error).

\paragraph{Fragment duplication}
The results for this task again reveal a consistent generalisation failure across all architectures.
In the case of Forces MAE, similar to Fragment Fusion, we find that
MACE performs best among the from-scratch models, but this model still has an OOD error, which is almost an order of magnitude higher than its ID error. The overall best performance is again obtained by the MACE-large foundation model. While MACE-large clearly outperforms MACE-small and MACE-medium OOD, the ID error across the three models is similar.
For Energy MAE, the OOD failure is the most severe across all tasks. For instance, the OOD error of MACE is about three orders of magnitude worse than its ID error. Even the MACE-large foundation model, which again achieves the best overall OOD results, performs more than two orders of magnitude worse than ID.

\paragraph{Fragment combination}

While all models still exhibit a clear generalisation gap, this gap is notably smaller than what we observed for Fragment Fusion and Fragment Duplication, both for Forces and Energy MAE.
Interestingly, for UMA, OOD performance essentially matches ID performance. However, the best overall OOD results are obtained by the MACE model trained from scratch, for both energy and forces.

\paragraph{Fragment Fission}
Since the models evaluated on this task are those trained for Fragment Duplication, their in-distribution errors are identical to those of that task. 
Out-of-distribution, on the alcohol and aldehyde test molecules, the Forces MAE rises sharply for almost all architectures, typically by about an order of magnitude and by more than two orders of magnitude for NequIP (Figure~\ref{fig:scatter_id_ood_forces_energy_mae}a). The exceptions are UMA-Small, whose OOD error remains comparable to its relatively high ID error, and SchNet, whose in-distribution error is already high.
MACE again achieves the lowest ID error, while MACE-Large achieves the lowest OOD error. For most models, the ranking and size of the errors closely follow those on Fragment Fusion: recovering the hydroxyl and carbonyl environments from a model trained on the carboxyl group is about as hard as composing the carboxyl group from them.

\subsection{Analysis}
\label{main:evaluation:analysis}
\paragraph{Where the failure occurs}

The aggregate Forces MAE conflates two distinct sources of error: the angle between predicted and reference force vectors, and the difference in their magnitudes. To separate them, we compare the magnitude error $\textit{norm}(\mathbf{F}_\text{pred})-\textit{norm}(\mathbf{F}_\text{label})$ with the total error $\textit{norm}(\mathbf{F}_\text{pred}-\mathbf{F}_\text{label})$, where \textit{norm} denotes the Euclidean norm, $\mathbf{F}_\text{pred}$ is the predicted force vector and $\mathbf{F}_\text{label}$ the corresponding ground truth, and measure direction by the cosine similarity between the two vectors. Aggregated over all atoms and snapshots of a Fragment Fusion trajectory (Appendix~\ref{details:per_atom_decomposition}), both sources are small in-distribution, with cosine values close to 1. Out-of-distribution, both grow, but the total error remains largely explained by the magnitude error: the dominant failure is in how strongly the model pushes each atom, not in which direction.

\begin{figure*}[!t]
    \centering
    \includegraphics[width=0.81\linewidth]{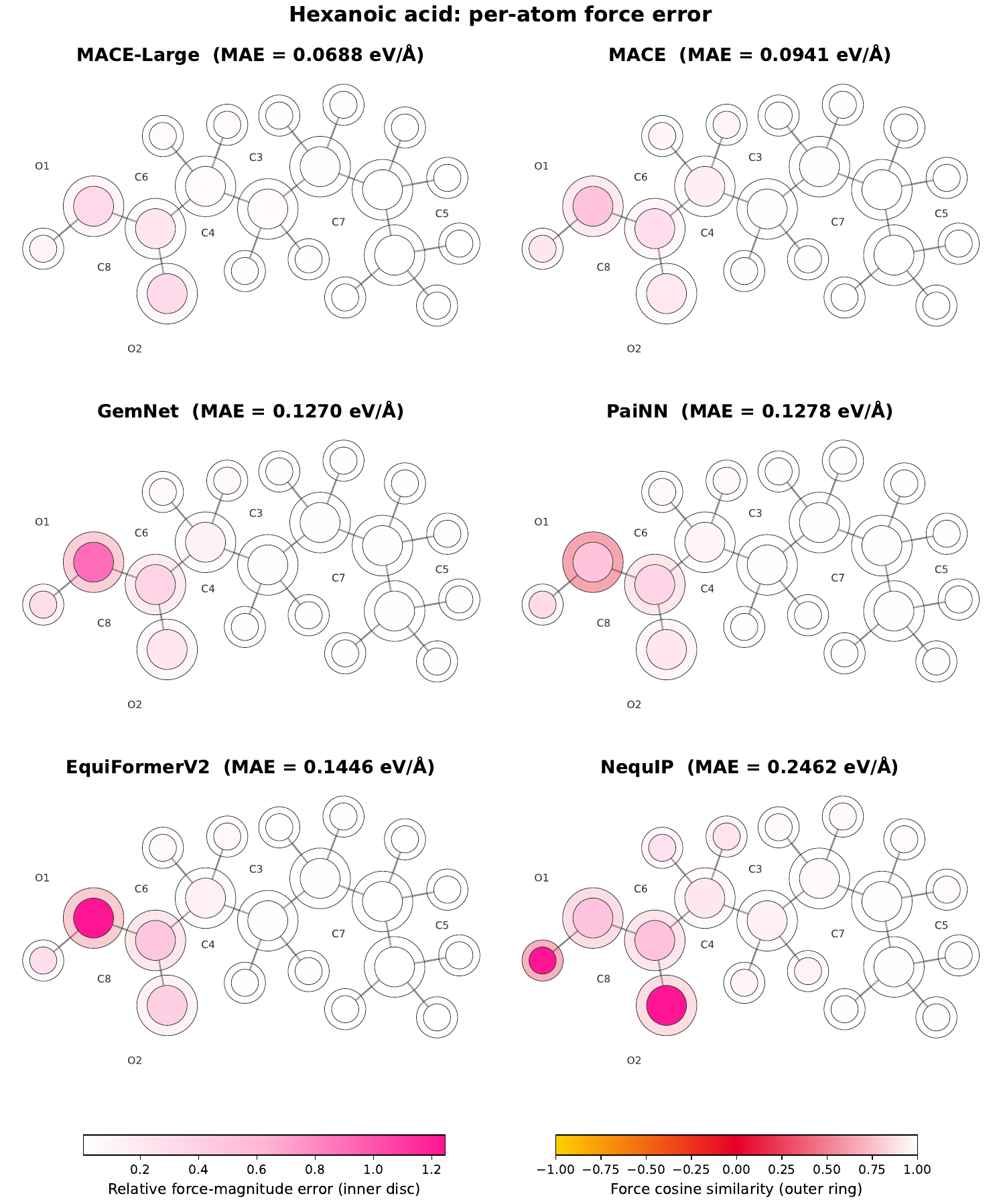}
    \caption{Per-atom force errors of six selected MLIPs on hexanoic acid, \(\text{CH}_3\text{(CH}_2\text{)}_4\text{COOH}\), an out-of-distribution test molecule of the Fragment Fusion task. Each atom's error is averaged over 200 test snapshots; panel titles give the mean force MAE.}
    \label{fig:FF_PAE_OOD}
\end{figure*}

Figure~\ref{fig:FF_PAE_OOD} shows where this error arises, by mapping it onto the molecule for one representative out-of-distribution example from the Fragment Fusion task, for six models. Each atom is drawn as a pair of concentric discs: the inner disc shows the relative force-magnitude error, $|\textit{norm}(\mathbf{F}_\text{pred})-\textit{norm}(\mathbf{F}_\text{label})| / \textit{norm}(\mathbf{F}_\text{label})$, and the outer ring the cosine similarity, both averaged over 200 test snapshots. Three features are visible that the aggregate metrics cannot show. First, the error is concentrated on the carboxyl group, the composite environment that the task introduces, while the carbon backbone is predicted well by all six models. Second, on the most affected atoms the magnitude error is persistent rather than transient, approaching the size of the reference force itself even after averaging. Third, the smaller directional component also sits within the carboxyl group, most clearly on the hydroxyl oxygen and hydrogen. The models differ, however, in how they distribute this error: some concentrate it on a single atom of the group, while NequIP spreads it over the whole group and into neighbouring hydrogens.

\begin{figure*}[!t]
    \centering
    \includegraphics[width=1.0\linewidth]{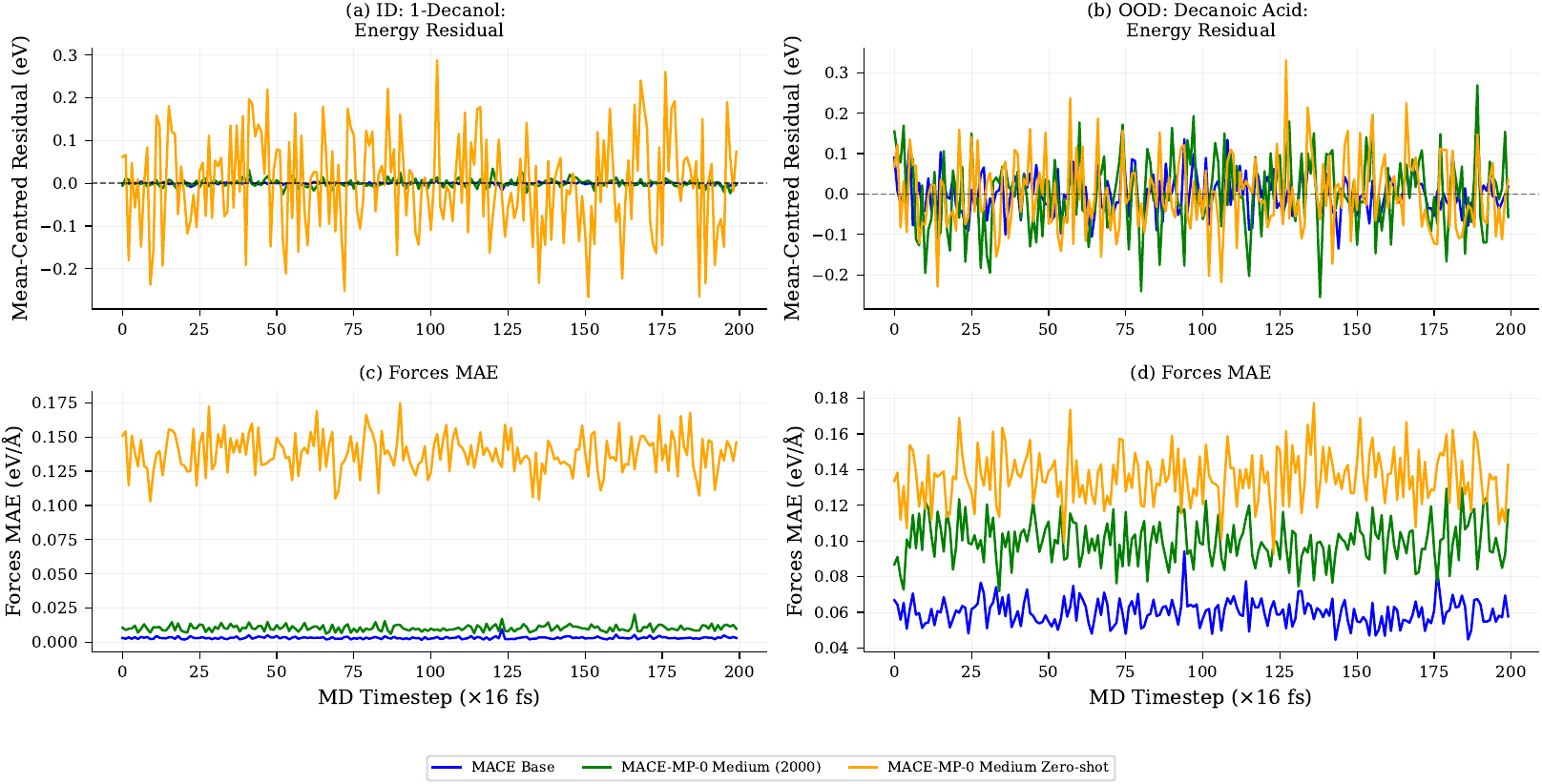}
    \caption{Per-timestep prediction analysis for three representative models on the \emph{Fragment Fusion} task over 200 MD snapshots. 
    Top row: mean-centred energy residual $r_j - \langle r \rangle_t$ in eV, where $r_j = \hat{E}_j - E_j$ is the energy residual at MD timestep $j$ and $\langle r \rangle_t = \frac{1}{T}\sum_{j=1}^{T} r_j$ is its mean across the $T$ snapshots of the trajectory; centring removes the trajectory-level systematic offset to isolate snapshot-to-snapshot fluctuation errors. 
    Bottom row: per-timestep forces MAE averaged over all atoms, computed as $\frac{1}{3N}\sum_{i,c}|\hat{F}_{i,c} - F_{i,c}|$ in eV/Å. 
    The left column shows in-distribution performance on 1-Decanol; the right column shows out-of-distribution performance on Decanoic Acid. 
    }
\label{fig:time_series_mace}
\end{figure*}

\paragraph{A closer look at the foundation-model regime}
The results from our main experiments (Figure~\ref{fig:scatter_id_ood_forces_energy_mae}) revealed that fine-tuned foundation models do not consistently outperform from-scratch baselines. 
To better understand why, Figure~\ref{fig:time_series_mace} shows the per-step force MAE residuals of three MACE variants on a representative Fragment Fusion trajectory. 
The zero-shot foundation model produces residuals of comparable magnitude on the ID and OOD molecules, with no clear separation between the two distributions. 
We interpret this similarity as reflecting the breadth of MACE-MP-0's pre-training corpus rather than evidence of genuine compositional generalisation: the molecules in our Fragment Fusion task are small organic species (decan-1-ol and decanoic acid) that lie well within the chemical scope of typical foundation-model pre-training data, so neither is genuinely out-of-distribution from the foundation model's perspective. 
The ID/OOD distinction in our benchmark is defined relative to the fine-tuning split, not to the pre-training corpus. 
Fine-tuning substantially reduces the residuals on the ID molecule but leaves the OOD residuals largely unchanged, consistent with the foundation model adapting its parameters to the specific reference used for the fine-tuning labels, ORCA(PBE/def2-TZVP+D3BJ), on the molecules it has been shown, without that adaptation transferring to molecules it has not. 
Analogous residual patterns are observed for the other tasks and foundation-model variants in Appendix~\ref{details:time_series}. 
A more thorough characterisation of this effect, including controlled comparisons between full fine-tuning, parameter-efficient adaptation, and readout-only fine-tuning, would be required to localise the failure mechanism within the model architecture, and we leave this as future work.



\paragraph{Further analyses.}
In Appendix~\ref{details}, we present further analyses that sharpen the picture developed above. 
The per-element force decomposition (Appendix~\ref{details:per_atom_decomposition}) reveals that directional errors localise specifically on the heteroatoms defining each task's novel chemical environment (oxygen for Fragment Fusion, Duplication and Fission, the asymmetric oxygen-nitrogen pairing for Fragment Combination), while the same atoms are well-predicted when their local environment matches one seen during training, and this composition differs across architectural families even where aggregate MAE does not (Appendix~\ref{app:failure-composition}).
This task-dependent pattern indicates that the failure is not due to undertraining of any specific element type, but due to a failure in transferring learned local environments to novel chemical contexts. 
Experiments on augmented task variants (Appendix~\ref{details:augmented_settings}) show that additional training coverage does not close the OOD gap for Fragment Fusion. 
Finally, the Fragment Fission results complement those on Fragment Duplication: the carboxyl group transfers as a unit to molecules that contain it, but the models cannot decompose it into the hydroxyl and carbonyl environments it is built from. This is consistent with the composite environment being learned as a single unit rather than built up from its parts, a property related to localism~\cite{DBLP:journals/jair/HupkesDMB20}. We do not, however, regard Fragment Fission as a direct test of localism: the hydroxyl and carbonyl environments in the test molecules differ physically from those inside the carboxyl group, so part of the shift reflects genuinely new local chemistry.

\section{Conclusions}
\label{sec:conclusions}

We introduced GMD, a systematic benchmark for evaluating the compositional generalisation capabilities of MLIPs. 
Our empirical analysis, based on this benchmark, revealed significant limitations in current approaches. 
While most of the models generalise reasonably well on \emph{fragment chain extension}, the remaining four tasks we considered expose substantial gaps between in-distribution and out-of-distribution performance, often spanning one to two orders of magnitude. 
These gaps are particularly pronounced for fragment fusion, fragment fission, and fragment duplication, where energy errors on out-of-distribution test molecules can exceed in-distribution errors by more than three orders of magnitude.
MACE emerged as the most robust performer, both when trained from scratch and when fine-tuning its foundation model variant MACE-MP-0.
Surprisingly, however, we found that the fine-tuned foundation models do not consistently outperform from-scratch models. This suggests that while pre-training broadens the chemical manifold over which models interpolate, it does not, by itself, impart the compositional structure that the GMD tasks require. 
As another important finding, we observed that in-distribution accuracy does not reliably predict out-of-distribution performance.

Our findings highlight a critical gap between the impressive accuracy of MLIPs on standard benchmarks and their ability to extrapolate beyond training distributions. By focusing on controlled compositional tasks, GMD isolates a generalisation challenge distinct from those targeted by existing benchmarks. Despite relying on physically-motivated architectures, current MLIPs do not, on their own, deliver compositional generalisation in the regime we evaluate. We hope the benchmark and accompanying toolkit will support the investigation of this gap.

\paragraph{Limitations}
GMD is scoped to small H/C/N/O molecules in vacuum with ORCA(PBE/def2-TZVP+D3BJ) labels.
In particular, the controlled experimental design comes at the cost of generality: findings on simple alkyl--functional-group fragments may not transfer directly to aromatic, charged, inorganic, or transition-metal chemistry, and the number of molecules was also limited by the cost of trajectory generation and DFT relabelling at this accuracy.
Conversely, failure under this minimal shift is a stronger result than failure on distant molecules would be, since the training data contains every component needed for each test molecule.
Extensions to higher-accuracy functionals, condensed phases, and richer fragment vocabularies (e.g.\ peptides for proteins, monomers for polymers, secondary building units for metal-organic frameworks) are left to future work.
Finally, foundation models were evaluated under full fine-tuning at a fixed budget.

\clearpage
\paragraph{Acknowledgements}
This paper was supported by EPSRC grants EP/W003309/1 and EP/Y028805/1 and Leverhulme Trust grant RPG-2025-041. We also acknowledge the School of Computational and Mathematical Sciences and the School of Physical, Chemical and Environmental Sciences for support.
The authors acknowledge the use of resources provided by the Isambard-AI National AI Research Resource (AIRR). Isambard-AI is operated by the University of Bristol and is funded by the UK Government’s Department for Science, Innovation and Technology (DSIT) via UK Research and Innovation; and the Science and Technology Facilities Council [ST/AIRR/I-A-I/1023]~\citep{mcintosh2024isambard}.
The authors acknowledge the use of resources provided by the Isambard 3 Tier-2 HPC Facility. Isambard 3 is hosted by the University of Bristol and operated by the GW4 Alliance (\url{https://gw4.ac.uk}) and is funded by UK Research and Innovation; and the Engineering and Physical Sciences Research Council [EP/X039137/1].
This work used Isambard-AI project \texttt{AIRR-GW-2026-01-01}. Access to Isambard 3 was provided via ARCCA through Cardiff University's membership of the GW4 Alliance.

{%

\bibliographystyle{abbrvnat}
\bibliography{src/references}
}

\clearpage
\appendix

\section{Additional details and analysis}
\label{details}
\subsection{Experiments with additional evaluation metrics}
\label{details:evaluation}

Force MAE conflates errors in both the direction and magnitude of the predicted force vectors, while total energy MAE is inherently confounded by molecular size. To disentangle these distinct sources of error, we present results for three additional metrics below: the unitless \emph{Cosine Similarity} and the \emph{MAE on Force Magnitude} (in eV/\AA) for forces, alongside the \emph{MAE on per-atom energy}.

\paragraph{Cosine similarity} This metric assesses the similarity between the directions of the predicted and true force vectors. A value of 1 indicates perfect alignment, 0 indicates orthogonality, and -1 indicates that the vectors point in opposite directions. The metric is computed by averaging the cosine of the angle between the predicted and true force vectors over all atoms in a given configuration:

$$ \text{Cosine Similarity} = \frac{1}{N} \sum_{i=1}^{N} \frac{\hat{\mathbf{F}}_i \cdot \mathbf{F}_i}{||\hat{\mathbf{F}}_i||_2 ||\mathbf{F}_i||_2} $$

where $N$ is the total number of atoms, $\hat{\mathbf{F}}_i$ is the predicted force vector for atom $i$, $\mathbf{F}_i$ is the ground truth force vector, $\cdot$ denotes the dot product, and $||\cdot||_2$ is the L2 norm (i.e., the vector magnitude).

\paragraph{MAE on force magnitude} This metric evaluates the accuracy of the predicted force strengths (magnitudes) independently of their direction. It quantifies the average absolute error between the magnitude of the predicted force vectors and the magnitude of the true force vectors:

$$ \text{MAE}_{\text{mag}} = \frac{1}{N} \sum_{i=1}^{N} \left| ||\hat{\mathbf{F}}_i||_2 - ||\mathbf{F}_i||_2 \right| $$

with $\hat{\mathbf{F}}_i$ and $\mathbf{F}_i$ as before. A lower value for $\text{MAE}_{\text{mag}}$ indicates a more accurate prediction of force strengths.

\paragraph{MAE on per-atom energy}
This metric normalises the total energy error by molecular size, disentangling the extensive growth of total energies with the number of atoms from genuine generalisation failure. Since total energies scale approximately linearly with molecular size, absolute energy MAE on longer-chain OOD molecules is confounded by their larger atom count. The per-atom normalised MAE removes this size effect, enabling a direct comparison of energetic prediction quality across molecules of different sizes:

$$ \mathrm{MAE}_{\text{energy, per atom}} = \frac{1}{M} \sum_{j=1}^{M} \frac{\lvert \hat{E}_j - E_j \rvert}{N_j} $$

where $M$ is the number of molecules in the test set, $N_j$ is the number of atoms in molecule $j$, and $\hat{E}_j$ and $E_j$ are the predicted and ground-truth total energies, respectively. A lower value indicates a more accurate per-atom energy prediction, independent of molecular size.

\begin{figure*}[!t]
  \centering
  \includegraphics[width=1.0\linewidth]{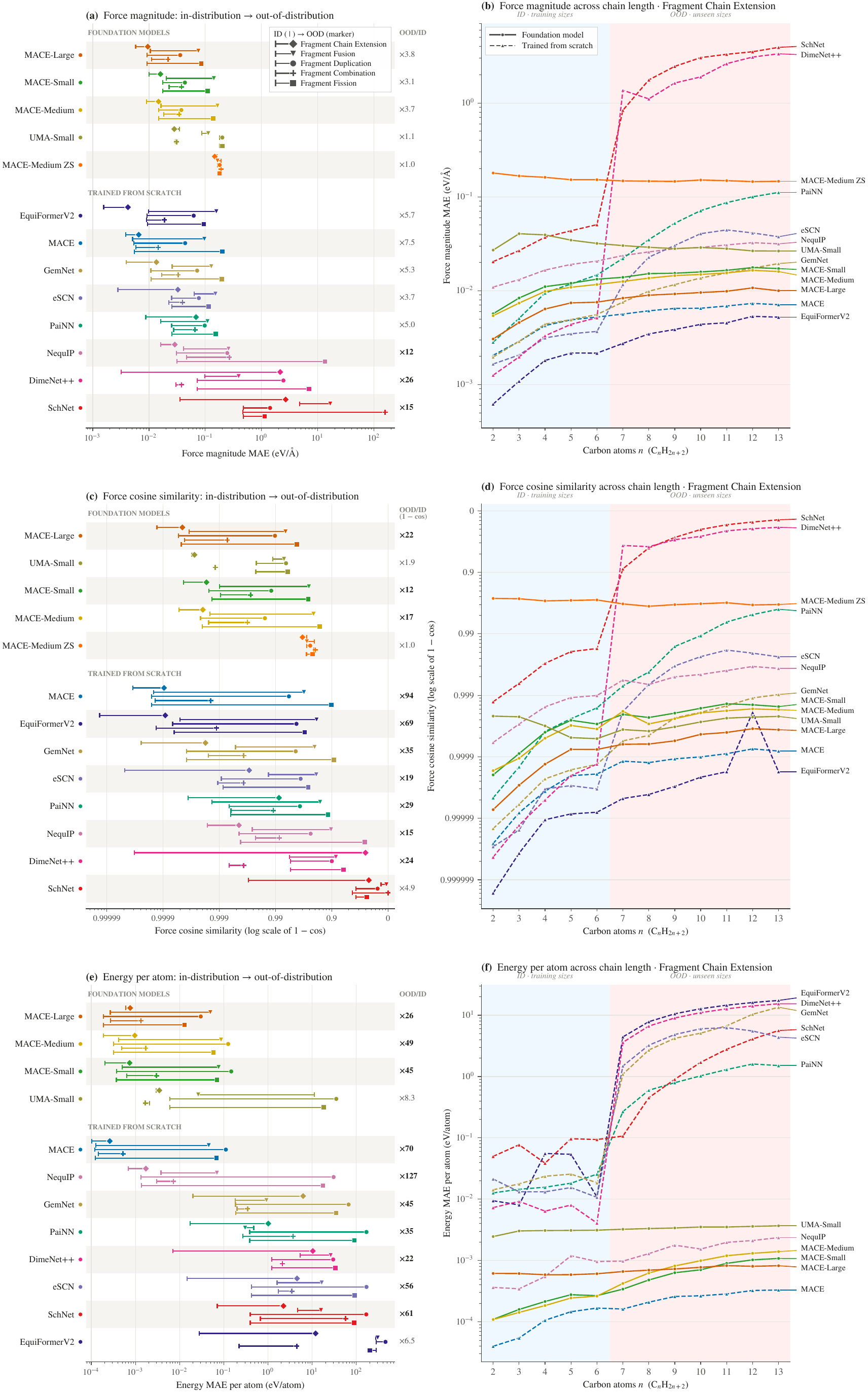}
  \caption{Supplementary force and energy analysis across all five GMD tasks (caption continued on the next page).}
  \label{fig:scatter_grid_off_metrics}
\end{figure*}

\paragraph{Results}
Figure~\ref{fig:scatter_grid_off_metrics} summarises the results across all five evaluation tasks in terms of cosine similarity, force magnitude MAE, and per-atom energy MAE. 
The analysis clearly demonstrates that errors in the predicted magnitude account for the vast majority of the overall force prediction errors. 
While out-of-distribution generalisation is relatively more robust with respect to force direction than magnitude, it is by no means practically sufficient. 
For \emph{Fragment Chain Extension} and \emph{Fragment Combination}, cosine similarities remain high, but even slight deviations from 1.0 are enough to accumulate trajectory errors and destabilise molecular dynamics simulations over time. 
For the more challenging tasks (\emph{Fragment Fusion} and \emph{Fragment Duplication}), the degradation in directional accuracy becomes distinctly visible, further undermining the physical reliability of the predicted forces.

In stark contrast, the force magnitude MAE exhibits massive out-of-distribution gaps that closely mirror the total force MAE failures. 
This confirms a critical mechanistic insight: while current MLIPs reliably capture the structural direction of atomic forces on novel compositions, they fundamentally fail to extrapolate their magnitudes. 
Furthermore, the per-atom energy MAE results (Figures~\ref{fig:scatter_grid_off_metrics}e,f) confirm that the catastrophic energy prediction failures observed in the main text are genuine compositional shortcomings; the orders-of-magnitude performance gap persists even after strictly normalising by atom count, ruling out the extensive scaling of larger out-of-distribution test molecules as the primary confounder.

\clearpage\noindent\textbf{Figure~\ref{fig:scatter_grid_off_metrics} (continued).}
Panels (a), (c) and (e) compare in-distribution (ID) and out-of-distribution (OOD) performance for three further metrics: (a) force magnitude MAE (eV/\AA), (c) force cosine similarity (1 indicates perfect directional agreement), and (e) energy MAE per atom (eV/atom), all on logarithmic axes; in (c), the axis shows $1-\cos$ on a log scale, labelled with the corresponding cosine values.
As in Figure~\ref{fig:scatter_id_ood_forces_energy_mae}, each row is one model, with foundation models and models trained from scratch shown in separate groups, and for each task a bar runs from the model's ID value (vertical tick) to its OOD value (task marker); colour encodes the model and marker shape the task.
The right-hand column reports the geometric mean over the five tasks of the ratio between OOD and ID values, computed on $1-\cos$ in (c) so that larger values always indicate a larger degradation; values of $10\times$ or more are shown in bold. MACE-Medium ZS is omitted from (e), since zero-shot energies are not comparable to our reference.
Panels (b), (d) and (f) show the same three metrics per molecule as a function of the number of carbon atoms $n$ on the Fragment Chain Extension task. Models were trained on linear alkanes with 2--6 carbon atoms (ID, shaded blue) and evaluated on unseen chains with 7--13 carbon atoms (OOD, shaded red); solid lines denote foundation models and dashed lines models trained from scratch.
Naming convention follows Figure~\ref{fig:scatter_id_ood_forces_energy_mae}.


\subsection{Per-trajectory residual analysis}
\label{details:time_series}

Aggregate metrics such as Mean Absolute Error collapse a model's behaviour across an entire trajectory into a single number, hiding the temporal structure of the prediction error. This temporal structure matters for two reasons. First, MLIPs are typically deployed to drive molecular dynamics over thousands to millions of steps; residuals that are biased, drifting, or temporally correlated propagate through the integrator, producing trajectory-level errors that the aggregate MAE does not capture. Two models with identical MAE can yield qualitatively different dynamics if one has zero-mean noise and the other has a systematic offset. Second, the time-series view is diagnostic in a way the aggregate is not: a constant offset, a slow drift, and high-frequency noise around zero all look the same in MAE but indicate different failure modes. Comparing the residuals of a model on an ID molecule with those on an OOD molecule from the same task is, in our setting, the most direct view of where the distribution shift breaks the prediction. Analogously, decomposing the force error into individual atoms reveals whether the failure is concentrated on chemically specific subsets of the molecule rather than being uniformly distributed.

\begin{figure*}[p]
    \centering
    \includegraphics[width=1.0\linewidth]{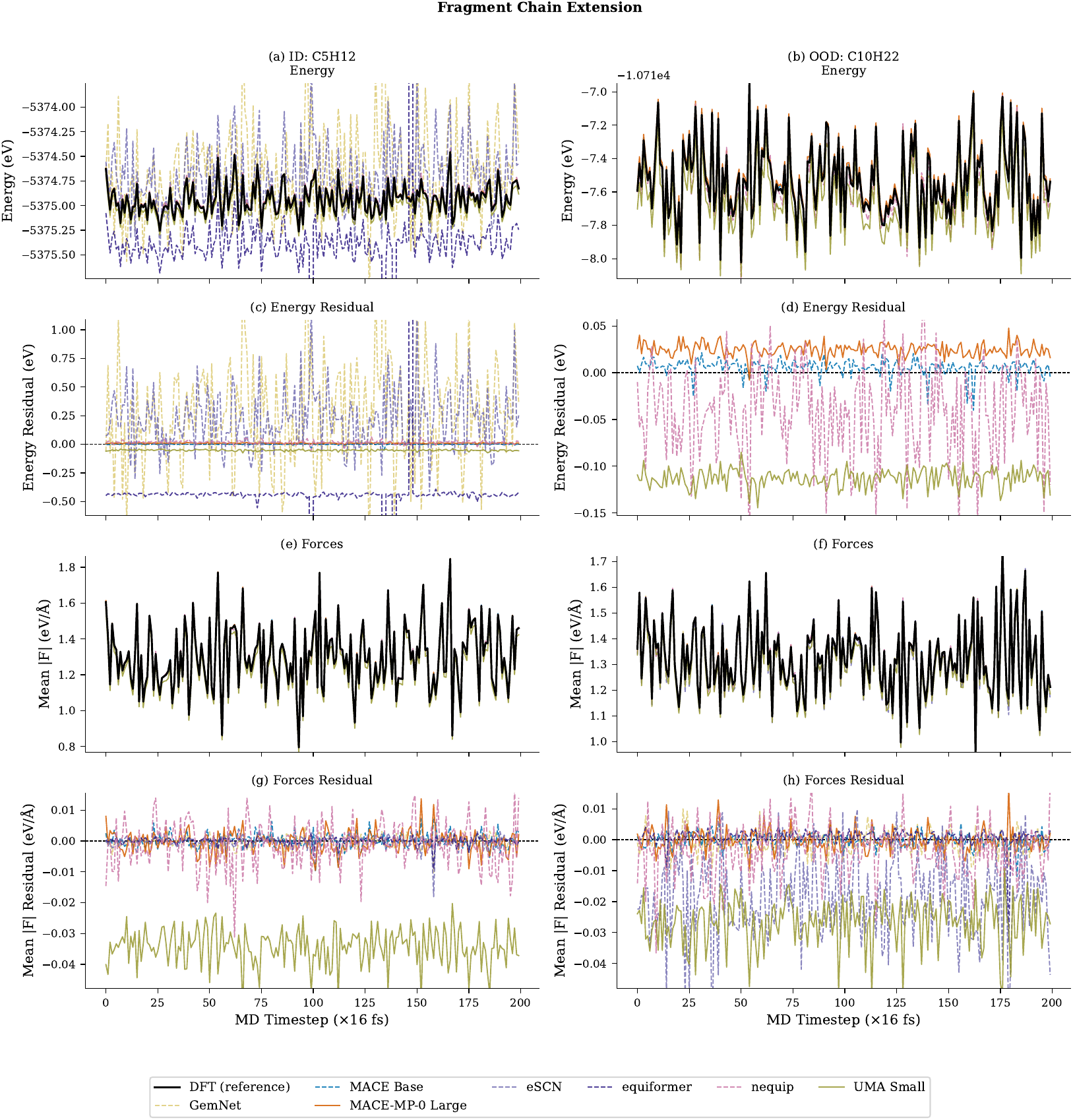}
    \caption{Per-trajectory residual analysis for the \emph{Fragment Chain Extension} task. Top row: predicted total energy (left, ID: pentane; right, OOD: decane) for selected models alongside the DFT reference. Second row: corresponding energy residuals (predicted minus DFT). Third row: per-step forces MAE (averaged over all atoms and Cartesian components) for the same trajectories. Fourth row: per-step forces MAE residuals. All panels span 200 consecutive MD snapshots with 16 fs spacing. Models shown: MACE (trained from scratch), MACE-MP-0 Large, eSCN, EquiFormerV2, NequIP, GemNet, and UMA-Small; remaining evaluated models are omitted from these panels because their absolute energy errors compress the plot's dynamic range to the point where comparison among the displayed models becomes impossible.}
    \label{fig:TS:FCE}
\end{figure*}

\begin{figure*}[p]
    \centering
    \includegraphics[width=1.0\linewidth]{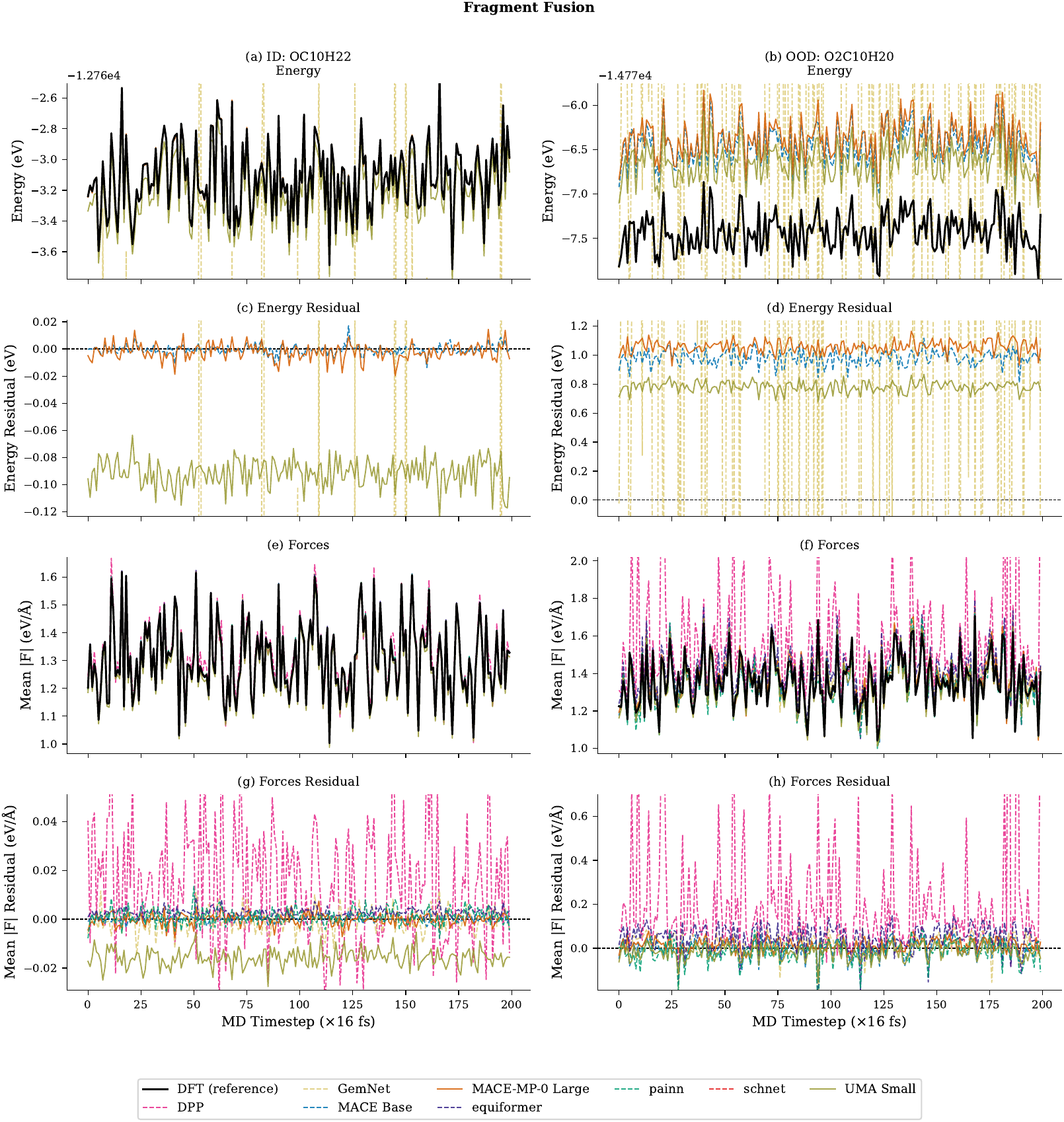}
    \caption{Per-trajectory residual analysis for the \emph{Fragment Fusion} task. Top row: predicted total energy (left, ID: decan-1-ol; right, OOD: decanoic acid) for selected models alongside the DFT reference. Second row: corresponding energy residuals. Third row: per-step forces MAE (averaged over all atoms and Cartesian components). Fourth row: per-step forces MAE residuals. All panels span 200 consecutive MD snapshots at 16~fs spacing. Models shown: MACE, MACE-MP-0 Large, EquiFormerV2, NequIP, PaiNN, DimeNet++, and UMA-Small; remaining evaluated models are omitted as in Figure~\ref{fig:TS:FCE}.}
    \label{fig:TS:FC}
\end{figure*}

\begin{figure*}[p]
    \centering
    \includegraphics[width=1.0\linewidth]{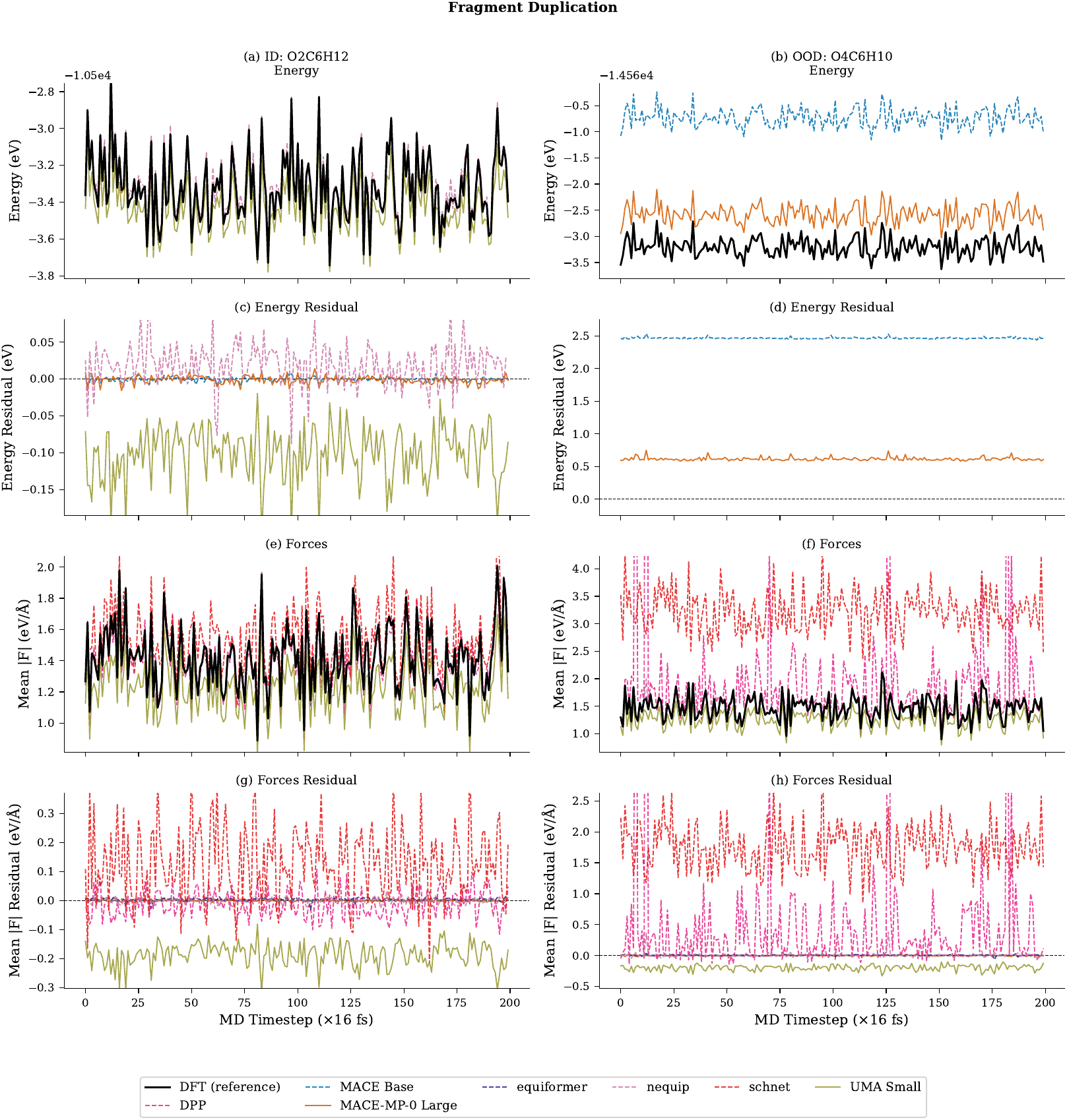}
    \caption{Per-trajectory residual analysis for the \emph{Fragment Duplication} task. Top row: predicted total energy (left, ID: hexanoic acid; right, OOD: hexanedioic acid, for selected models alongside the DFT reference. Second row: corresponding energy residuals. Third row: per-step forces MAE. Fourth row: per-step forces MAE residuals. All panels span 200 consecutive MD snapshots at 16~fs spacing. Models shown: MACE, MACE-MP-0 Large, EquiFormerV2, NequIP, PaiNN, SchNet, DimeNet++, and UMA-Small; remaining evaluated models are omitted as in Figure~\ref{fig:TS:FCE}.}
    \label{fig:TS:FD}
\end{figure*}

\begin{figure*}[p]
    \centering
    \includegraphics[width=1.0\linewidth]{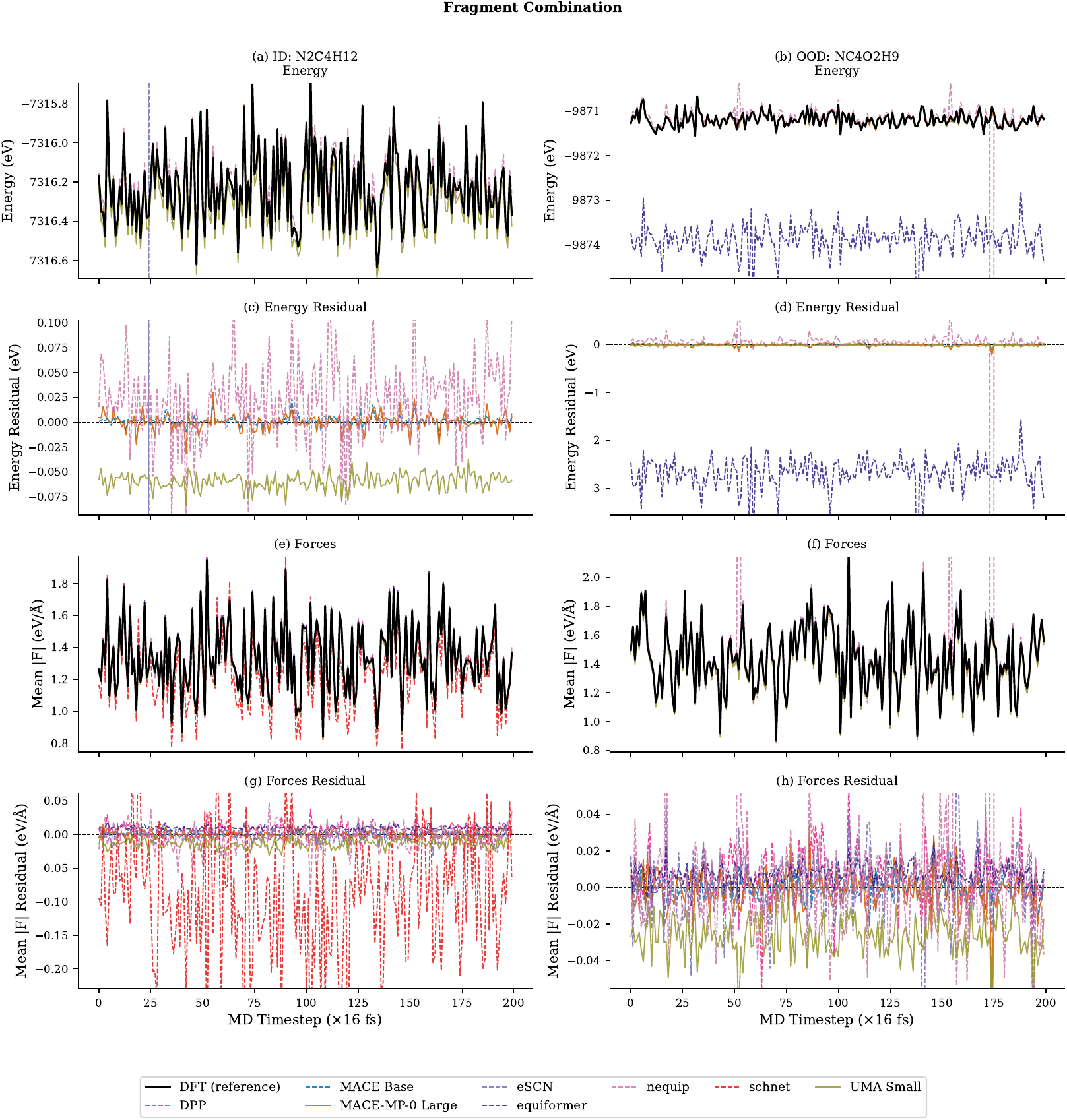}
    \caption{Per-trajectory residual analysis for the \emph{Fragment Combination} task. Top row: predicted total energy (left, ID: a representative diamine; right, OOD: a representative amino acid) for selected models alongside the DFT reference. Second row: corresponding energy residuals. Third row: per-step forces MAE. Fourth row: per-step forces MAE residuals. All panels span 200 consecutive MD snapshots with 16 fs spacing. Models shown: MACE, MACE-MP-0 Large, EquiFormerV2, NequIP, PaiNN, SchNet, DimeNet++, and UMA-Small; remaining evaluated models are omitted as in Figure~\ref{fig:TS:FCE}.}
    \label{fig:TS:FCB}
\end{figure*}

\begin{figure*}[p]
    \centering
    \includegraphics[width=1.0\linewidth]{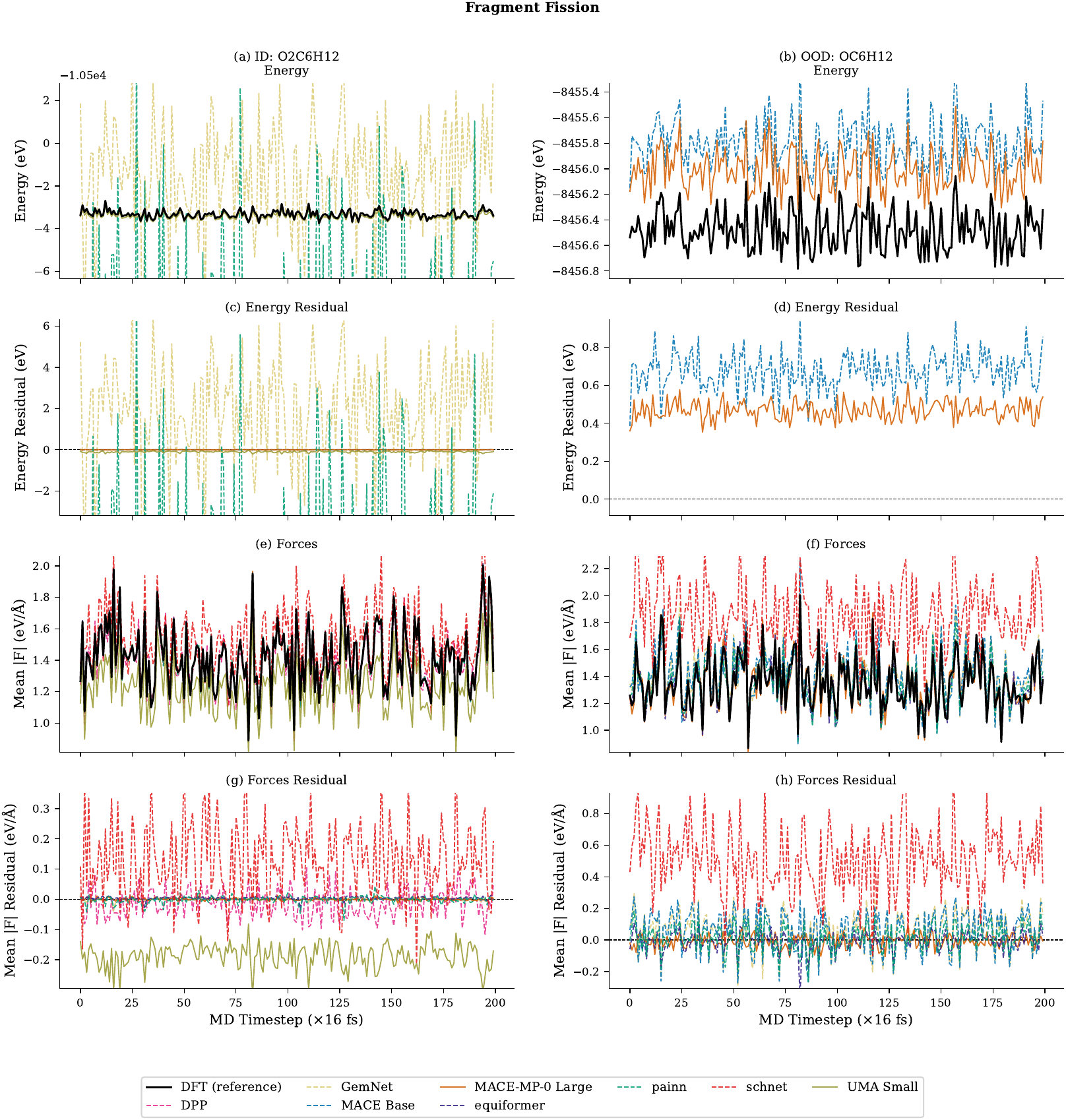}
    \caption{Per-trajectory residual analysis for the \emph{Fragment Fission} task. Left column: in-distribution molecule (hexanoic acid, C$_6$H$_{12}$O$_2$); right column: out-of-distribution molecule (hexanal, C$_6$H$_{12}$O). First row: predicted total energy for selected models alongside the DFT reference. Second row: corresponding energy residuals. Third row: per-step mean force magnitude, $\langle|\mathbf{F}|\rangle$, averaged over all atoms. Fourth row: residual of the mean force magnitude with respect to the DFT reference. All panels span 200 consecutive MD snapshots with 16\,fs spacing. Models shown: MACE, MACE-MP-0 Large, EquiFormerV2, GemNet, PaiNN, SchNet, DimeNet++, and UMA-Small; remaining evaluated models are omitted as in Figure~\ref{fig:TS:FCE}.}
    \label{fig:TS:FF}
\end{figure*}

\begin{figure*}[p]
    \centering
    \includegraphics[width=1.0\linewidth]{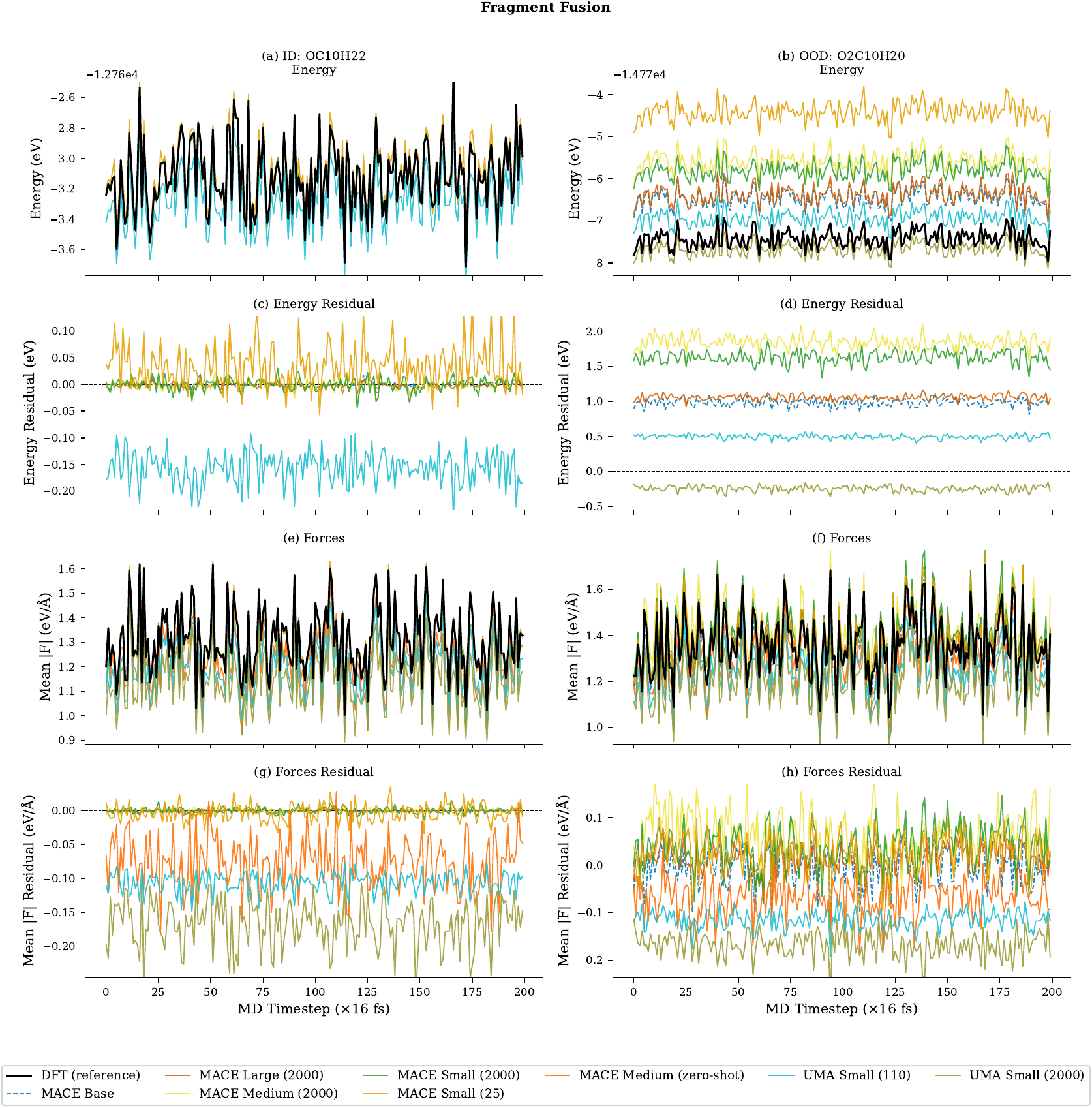}
    \caption{Per-trajectory residual analysis comparing foundation-model variants on a representative trajectory from the \emph{Fragment Fusion} task (ID: decan-1-ol; OOD: decanoic acid). Layout matches Figures~\ref{fig:TS:FCE}--\ref{fig:TS:FF}: top row shows predicted energy alongside DFT, second row shows energy residuals, third row shows per-step forces MAE, fourth row shows forces MAE residuals. Models shown: MACE-MP-0 Small (25, 110, and 2000 epochs of fine-tuning), MACE-MP-0 Medium (2000 epochs), MACE-MP-0 Large (2000 epochs), MACE-MP-0 Medium evaluated zero-shot, and UMA-Small (110 and 2000 epochs of fine-tuning), with the from-scratch MACE included for reference. This view isolates the effect of model size at fixed fine-tuning budget (Small/Medium/Large at 2000 epochs) and the effect of fine-tuning budget at fixed model size (Small at 25, 110, 2000 epochs; UMA-Small at 110, 2000 epochs).}
    \label{fig:TS:FC:Foundations}
\end{figure*}

For each task, Figures~\ref{fig:TS:FCE}--\ref{fig:TS:FF} show one representative ID molecule and one representative OOD molecule. In addition, Figure~\ref{fig:TS:FC:Foundations} specifically represents the performance of foundation models on the \emph{Fragment Fusion} task. For each pair, we plot the predicted total energy and per-atom force magnitude alongside the DFT reference, together with the corresponding residuals, over 200 consecutive MD snapshots. For each task, we select one ID and one OOD molecule and plot the same set of models on both. Models with extremely large absolute errors are omitted from the corresponding panels: a model whose energy predictions are several orders of magnitude off from the DFT reference compresses the plot's dynamic range to the point where comparison among the remaining models becomes impossible, even though its force predictions may still be informative. Concretely, a model may appear in a force panel but not in the corresponding energy panel for the same trajectory if its energy errors are too large to share an axis with the better-performing models. Additionally, a small number of models with severely noisy step-to-step predictions are omitted to keep the residual traces readable. The exact set of models shown in each panel is given in the figure captions.

\paragraph{Observations.}
Several patterns recur across tasks. On ID molecules, MACE trained from scratch produces residuals that are tightly distributed around zero across the full trajectory, while fine-tuned MACE-MP-0 variants show small but persistent offsets. 
On OOD molecules, these offsets grow systematically, and the from-scratch models develop biases of their own that are largely absent in ID. Foundation models evaluated zero-shot exhibit the largest and most persistent offsets, consistent with the calibration argument made in the main text (Section~\ref{main:evaluation:analysis}). 
Force-magnitude residuals follow the same qualitative trends as energy residuals but with smaller relative magnitudes, supporting the observation in the main text that the generalisation failure is concentrated in the extensive readout rather than in the local geometric features. 
We caution that these are observations from individual trajectories rather than a quantitative population analysis; they are intended to illustrate the kind of mechanistic information visible in the time-series view but invisible in aggregate metrics.

\paragraph{Foundation-model variants.}
Figure~\ref{fig:TS:FC:Foundations} additionally shows residuals for the full set of foundation-model variants on a representative trajectory, isolating the effect of model size and fine-tuning budget. The MACE-MP-0 family is shown at three scales (Small, Medium, Large) and at three fine-tuning budgets (25, 110, and 2000 epochs for MACE-MP-0 Small; 2000 epochs for Medium and Large), alongside UMA-Small at 110 and 2000 epochs and MACE-MP-0 Medium evaluated zero-shot. This view isolates two effects that the per-task figures conflate: increasing model size with a fixed fine-tuning budget, and increasing the fine-tuning budget with a fixed model size.

\begin{figure*}[t]
    \centering
    \includegraphics[width=1.0\linewidth]{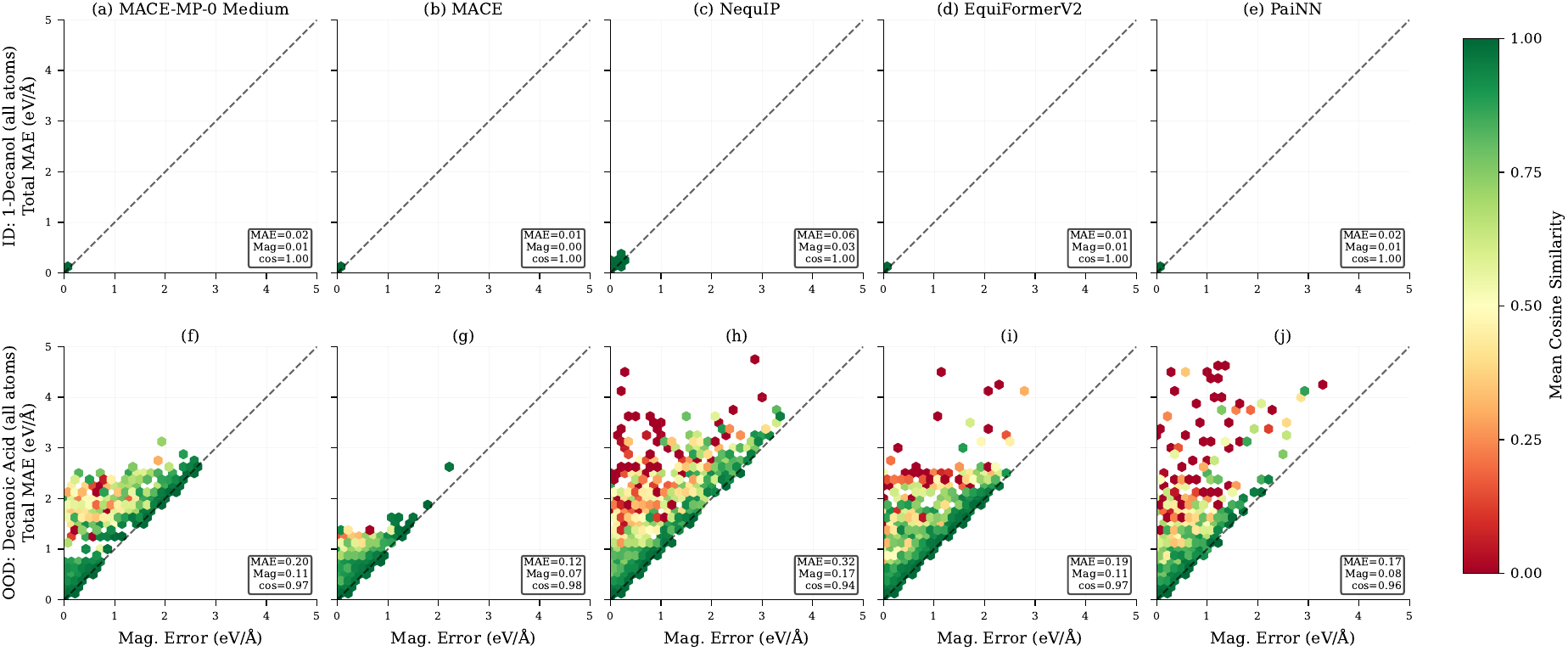}
    \caption{Force error decomposition for selected evaluated models on the \emph{Fragment Fusion} task, comparing in-distribution (ID, top row: 1-Decanol) and out-of-distribution (OOD, bottom row: Decanoic Acid) performance across 200 MD trajectory snapshots. 
    Each point represents a single atom at a single timestep. 
    The x-axis shows the magnitude error $|\textit{norm}(\mathbf{F}_\text{pred}) - \textit{norm}(\mathbf{F}_\text{label})|$ and the y-axis shows the total force vector error $\textit{norm}(\mathbf{F}_\text{pred} - \mathbf{F}_\text{label})$, both in eV/Å. 
    Colour encodes the mean cosine similarity between predicted and reference force vectors per hexbin 
    . 
    The dashed diagonal represents the theoretical lower bound where direction is perfectly predicted, and total error equals magnitude error alone. 
    Points above the diagonal indicate directional error, inflating total MAE beyond the magnitude contribution.}

\label{fig:decompose_all_atoms}
\end{figure*}

\paragraph{Per-atom error decomposition.}

\begin{figure}[p]
    \centering
    \begin{subfigure}{\linewidth}
        \includegraphics[width=\linewidth]{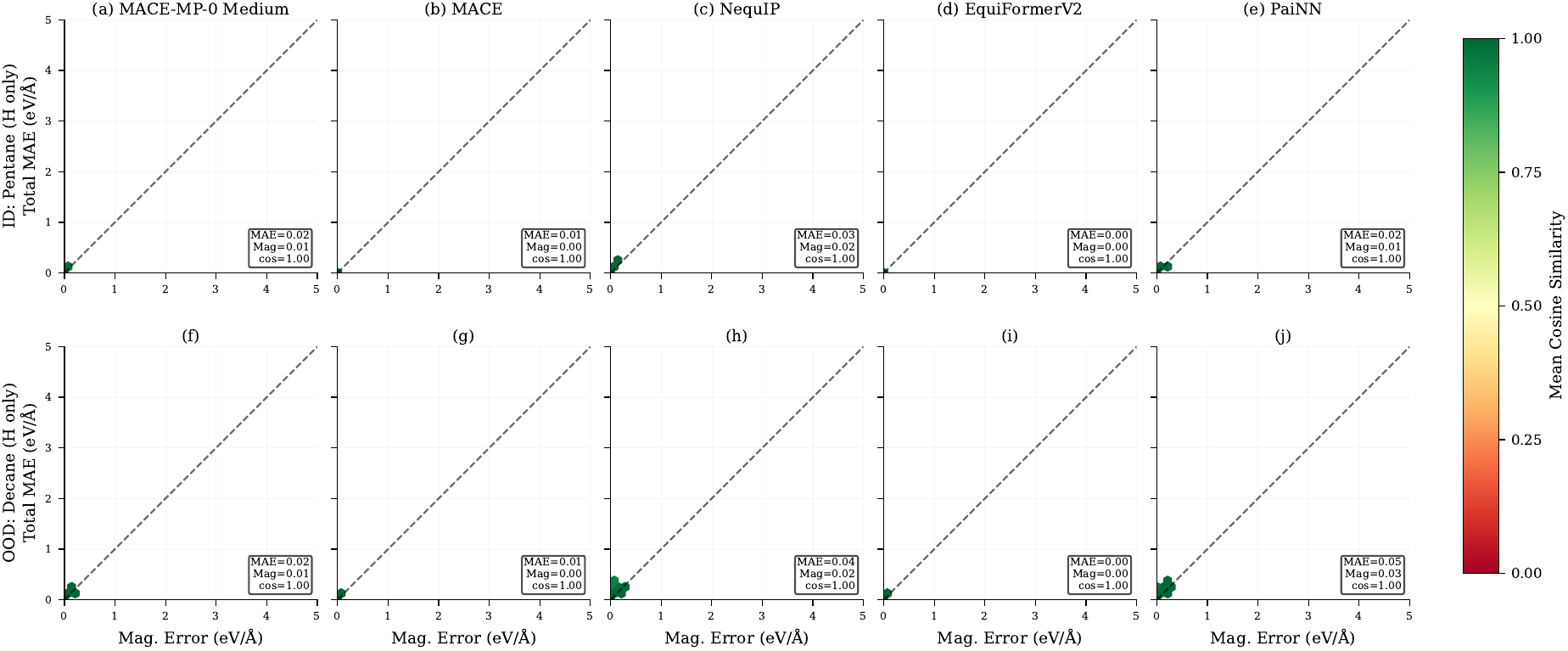}
        \caption{Fragment Chain Extension: Hydrogen}
        \label{fig:FCE:decomp_h}
    \end{subfigure}
    \vspace{0.5cm}
    \begin{subfigure}{\linewidth}
        \includegraphics[width=\linewidth]{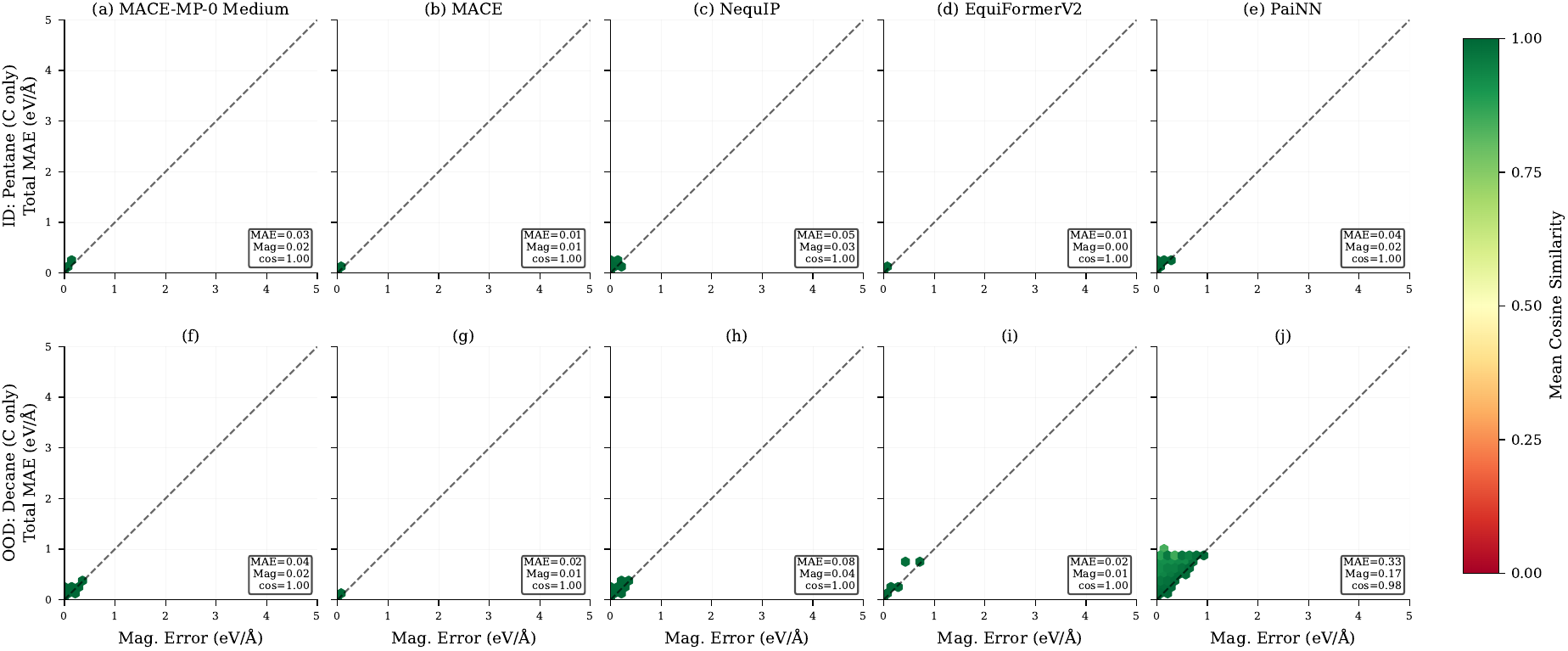}
        \caption{Fragment Chain Extension: Carbon}
        \label{fig:FCE:decomp_c}
    \end{subfigure}
    \caption{Element-specific force error decomposition on the \emph{Fragment Chain Extension} task for hydrogen (top) and carbon (bottom). Each block compares in-distribution (ID: pentane) and out-of-distribution (OOD: decane) predictions over 200 MD snapshots.
    The x-axis shows the magnitude error $|\textit{norm}(\mathbf{F}_\text{pred}) - \textit{norm}(\mathbf{F}_\text{label})|$ and the y-axis shows the total force vector error $\textit{norm}(\mathbf{F}_\text{pred} - \mathbf{F}_\text{label})$, both in eV/\AA.
    Hexbin colour encodes mean cosine similarity (green $= 1.0$, red $= 0.0$). The dashed diagonal represents zero directional error; points above this line highlight where directional mismatch inflates the total MAE. Unlike the other tasks, the points remain concentrated near the origin for all element types, consistent with the comparatively strong out-of-distribution performance on this task.}
    \label{fig:FCE:decomp_all}
\end{figure}

\begin{figure}[p]
    \centering
    \begin{subfigure}{\linewidth}
        \includegraphics[width=\linewidth]{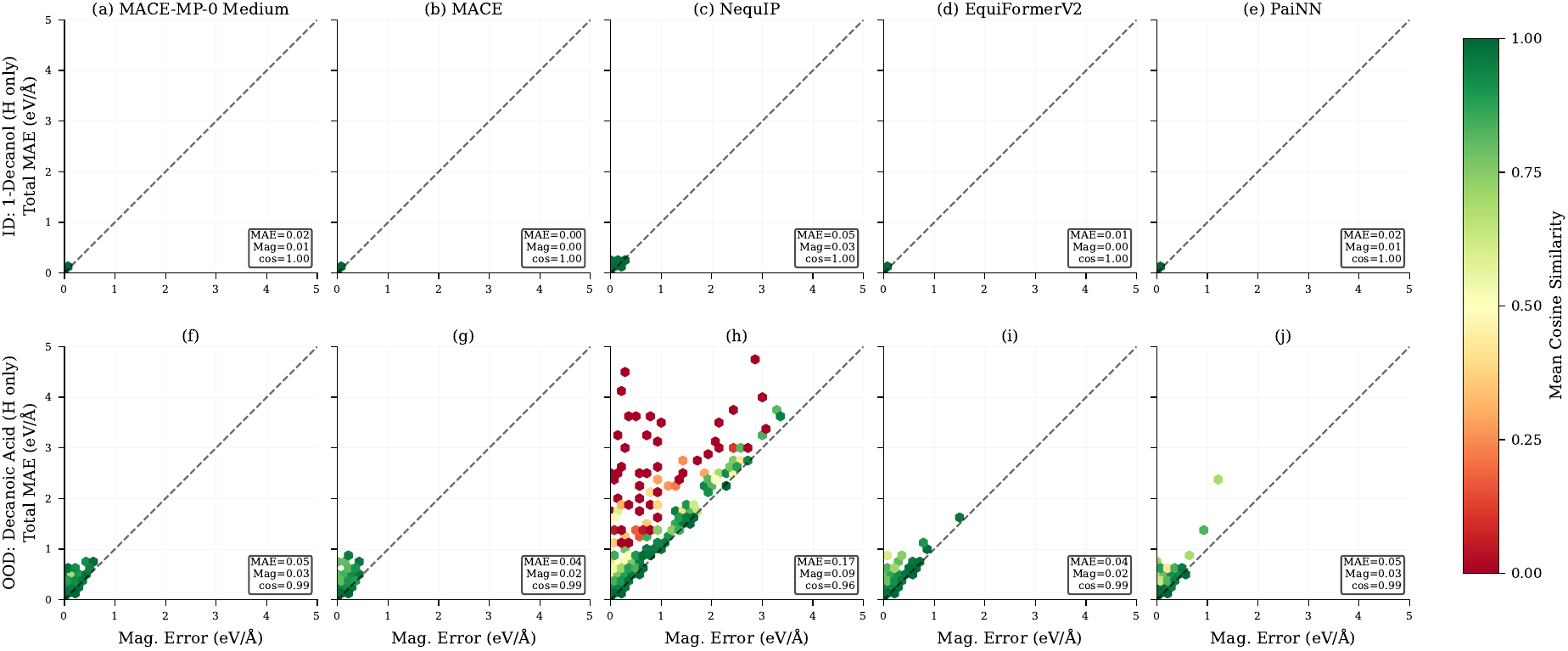}
        \caption{Fragment Fusion: Hydrogen}
        \label{fig:FC:decomp_h}
    \end{subfigure}
    \vspace{0.5cm}
    \begin{subfigure}{\linewidth}
        \includegraphics[width=\linewidth]{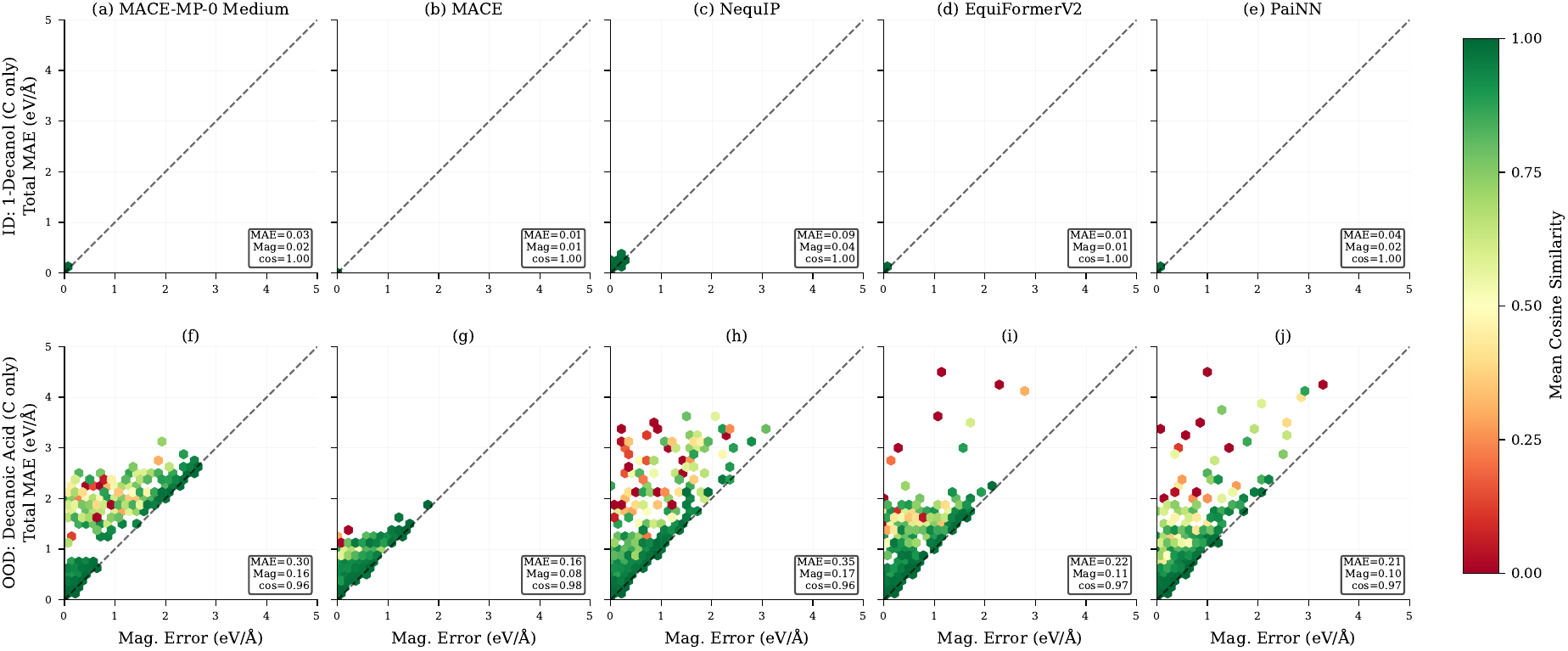}
        \caption{Fragment Fusion: Carbon}
        \label{fig:FC:decomp_c}
    \end{subfigure}
    \vspace{0.5cm}
    \begin{subfigure}{\linewidth}
        \includegraphics[width=\linewidth]{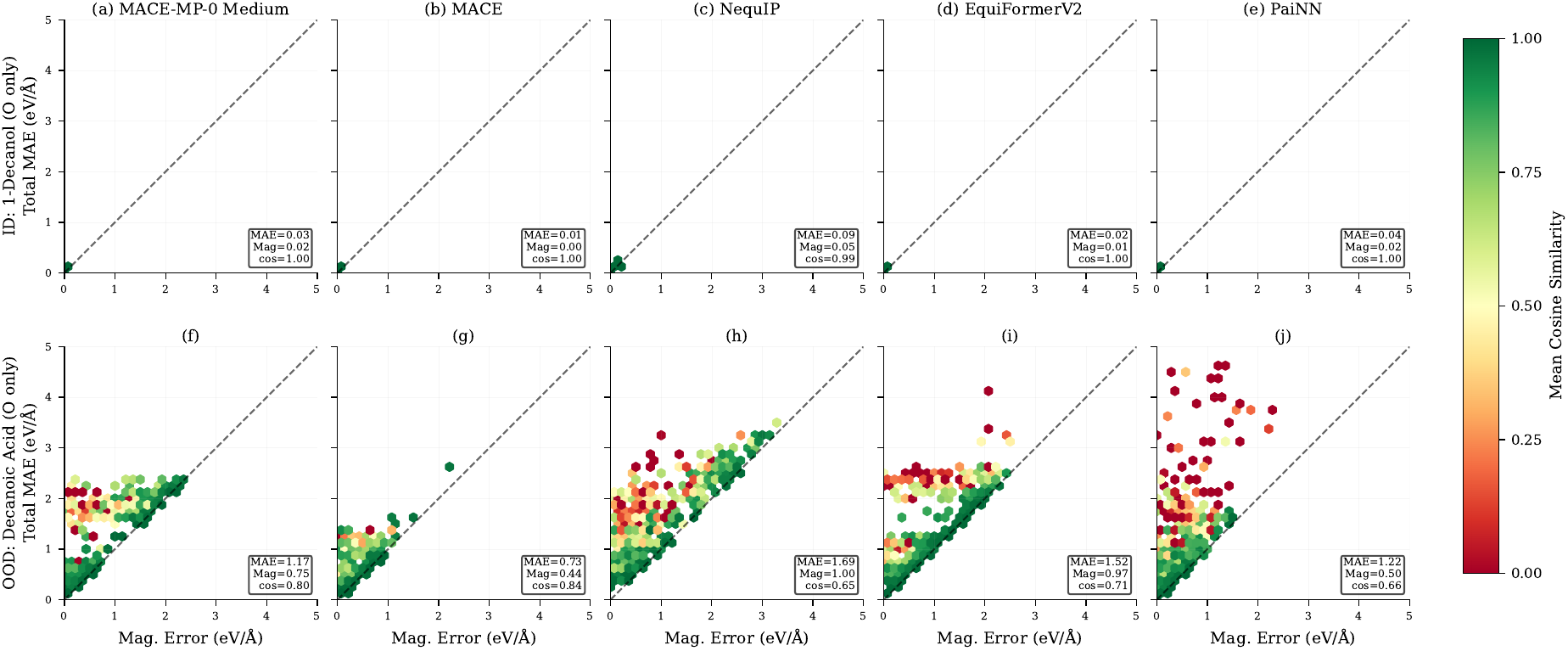}
        \caption{Fragment Fusion: Oxygen}
        \label{fig:FC:decomp_o}
    \end{subfigure}
    \caption{Element-specific force error decomposition on the \emph{Fragment Fusion} task for hydrogen (top), carbon (middle), and oxygen (bottom). Each block compares in-distribution (ID: decan-1-ol) and out-of-distribution (OOD: decanoic acid) predictions over 200 MD snapshots.
    The x-axis shows the magnitude error $|\textit{norm}(\mathbf{F}_\text{pred}) - \textit{norm}(\mathbf{F}_\text{label})|$ and the y-axis shows the total force vector error $\textit{norm}(\mathbf{F}_\text{pred} - \mathbf{F}_\text{label})$, both in eV/\AA.
    Hexbin colour encodes mean cosine similarity (green $= 1.0$, red $= 0.0$). The dashed diagonal represents zero directional error; points above this line highlight where directional mismatch inflates the total MAE. The concentration of points near the origin is itself a result rather than an artefact of the plotting range: the majority of atoms remain accurately predicted out-of-distribution, and the error concentrates on a minority of them.}
    \label{fig:FC:decomp_all}
\end{figure}

\begin{figure}[p]
    \centering
    \begin{subfigure}{\linewidth}
        \includegraphics[width=\linewidth]{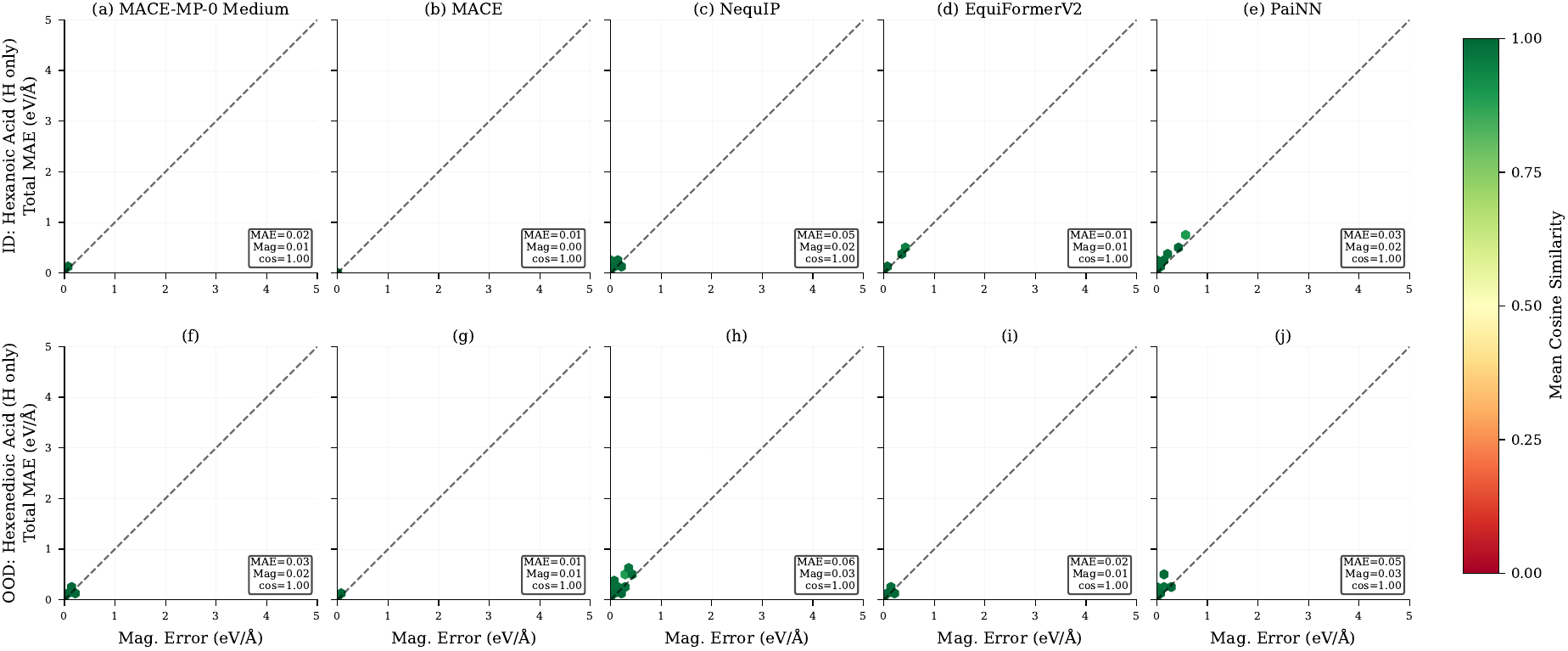}
        \caption{Fragment Duplication: Hydrogen}
        \label{fig:FD:decomp_h}
    \end{subfigure}
    \vspace{0.5cm}
    \begin{subfigure}{\linewidth}
        \includegraphics[width=\linewidth]{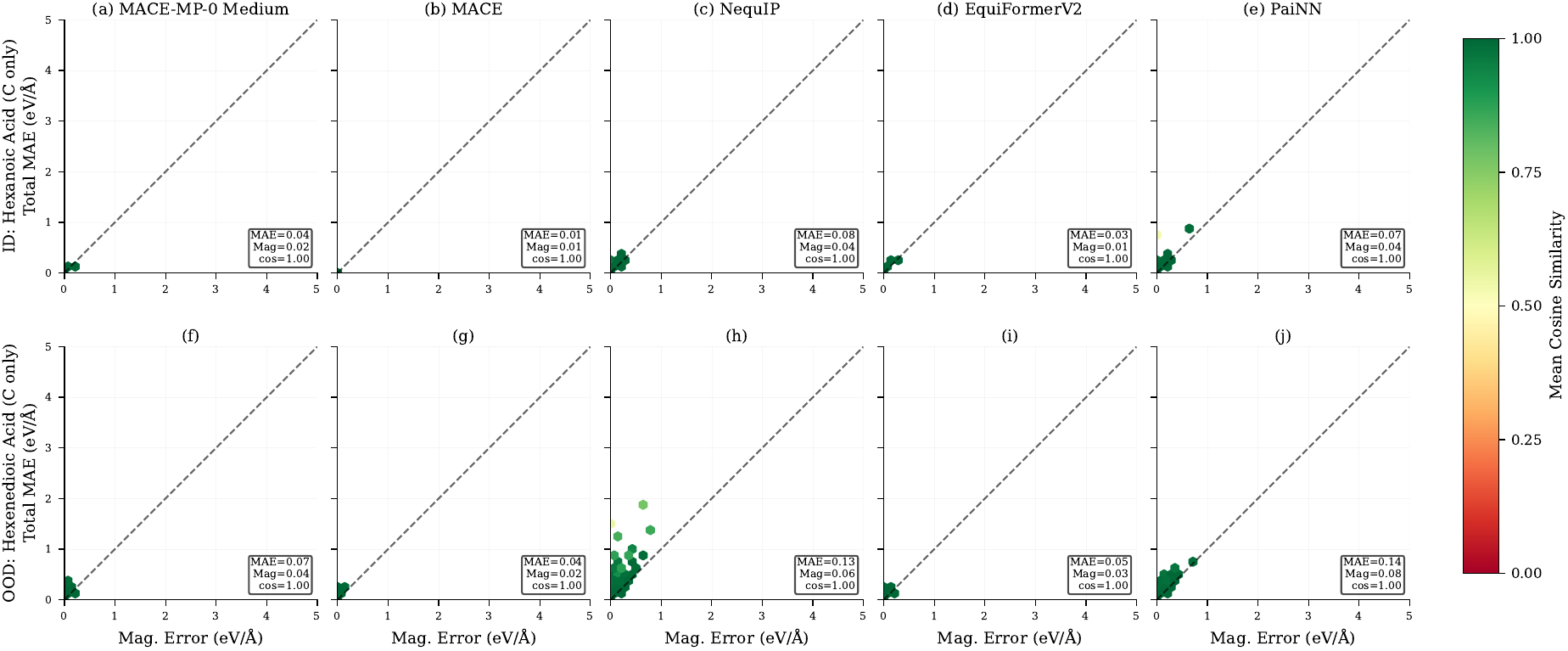}
        \caption{Fragment Duplication: Carbon}
        \label{fig:FD:decomp_c}
    \end{subfigure}
    \vspace{0.5cm}
    \begin{subfigure}{\linewidth}
        \includegraphics[width=\linewidth]{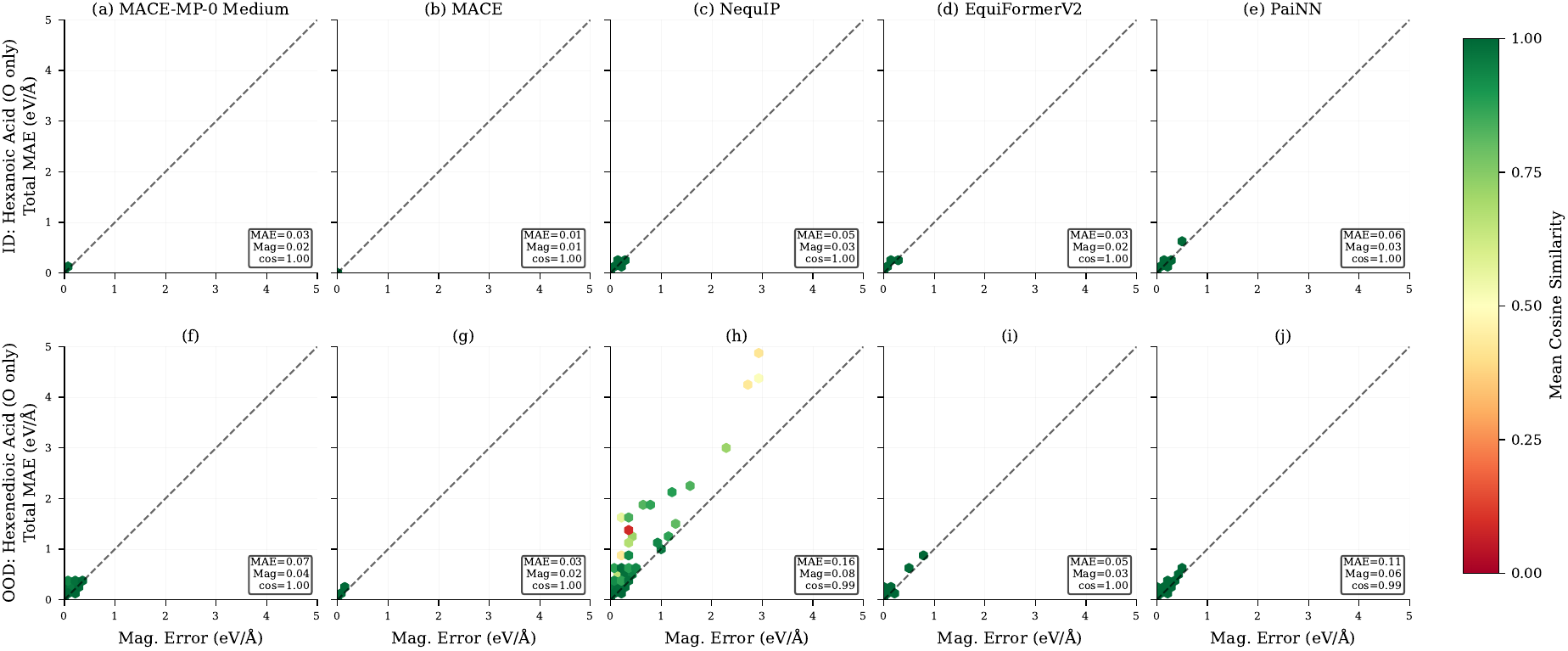}
        \caption{Fragment Duplication: Oxygen}
        \label{fig:FD:decomp_o}
    \end{subfigure}
    \caption{Element-specific force error decomposition on the \emph{Fragment Duplication} task for hydrogen (top), carbon (middle), and oxygen (bottom). Each block compares in-distribution (ID: hexanoic acid) and out-of-distribution (OOD: hexanedioic acid) predictions over 200 MD snapshots.
    The x-axis shows the magnitude error $|\textit{norm}(\mathbf{F}_\text{pred}) - \textit{norm}(\mathbf{F}_\text{label})|$ and the y-axis shows the total force vector error $\textit{norm}(\mathbf{F}_\text{pred} - \mathbf{F}_\text{label})$, both in eV/\AA.
    Hexbin colour encodes mean cosine similarity (green $= 1.0$, red $= 0.0$). The dashed diagonal represents zero directional error; points above this line highlight where directional mismatch inflates the total MAE. As above, the density of points near the origin reflects the finding rather than the plotting range: most atoms remain accurately predicted out-of-distribution, with the error concentrated on a minority of sites.}
    \label{fig:FD:decomp_all}
\end{figure}

\begin{figure}[p]
    \centering
    \begin{subfigure}{\linewidth}
        \includegraphics[width=\linewidth]{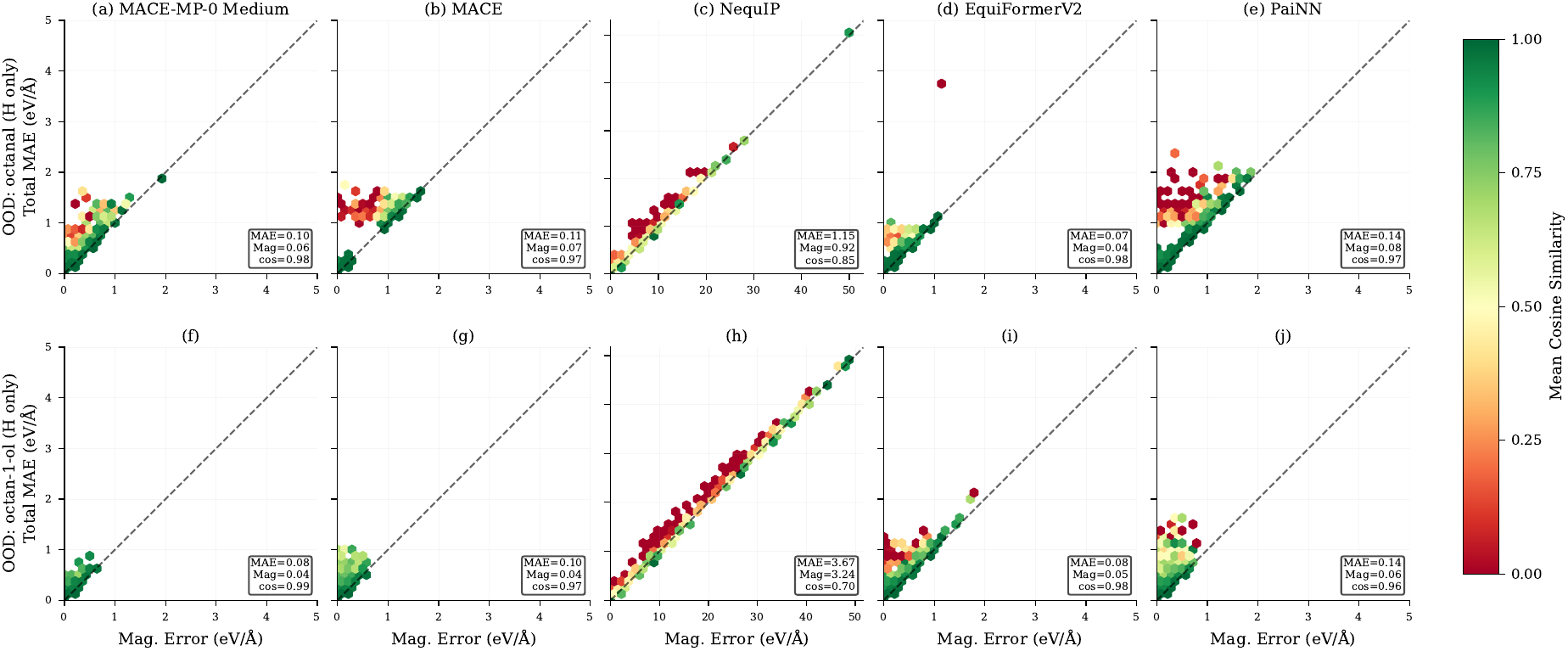}
        \caption{Fragment Fission: Hydrogen}
        \label{fig:FF:decomp_h}
    \end{subfigure}
    \vspace{0.5cm}
    \begin{subfigure}{\linewidth}
        \includegraphics[width=\linewidth]{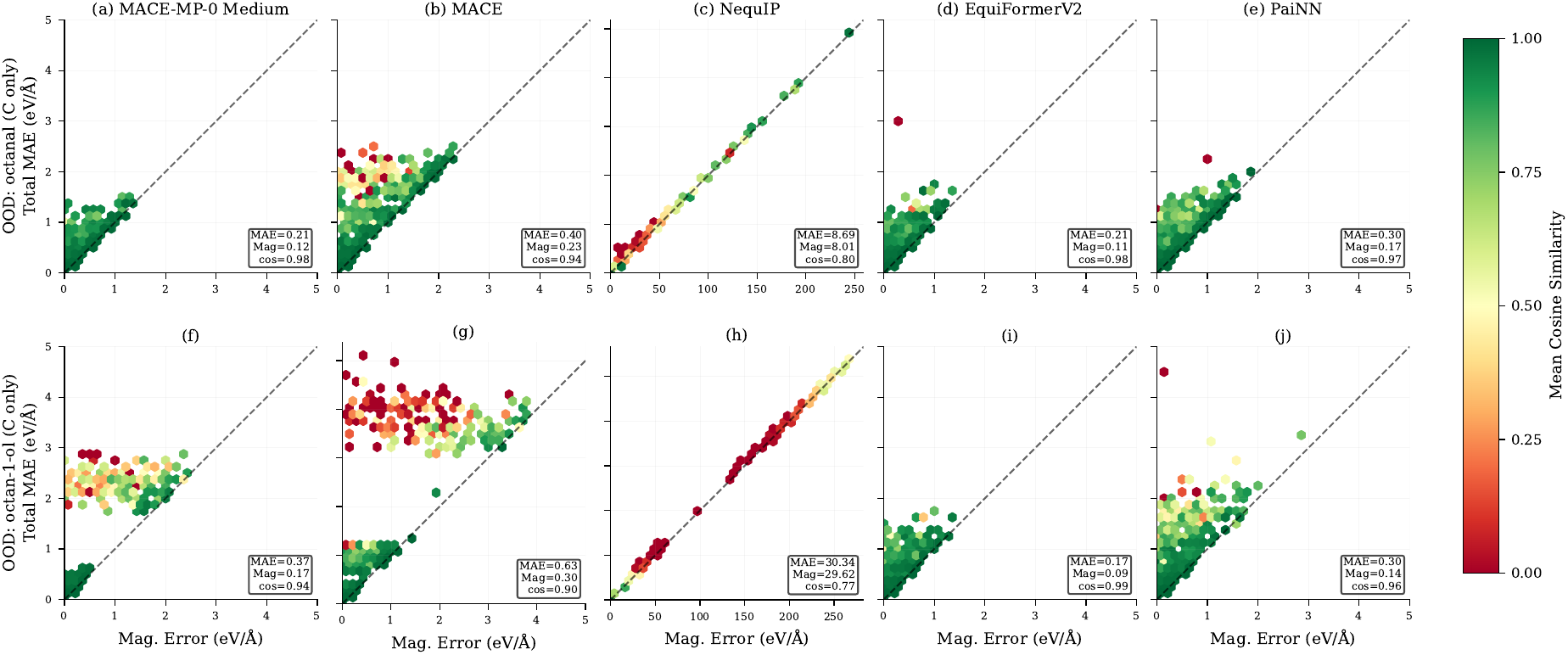}
        \caption{Fragment Fission: Carbon}
        \label{fig:FF:decomp_c}
    \end{subfigure}
    \vspace{0.5cm}
    \begin{subfigure}{\linewidth}
        \includegraphics[width=\linewidth]{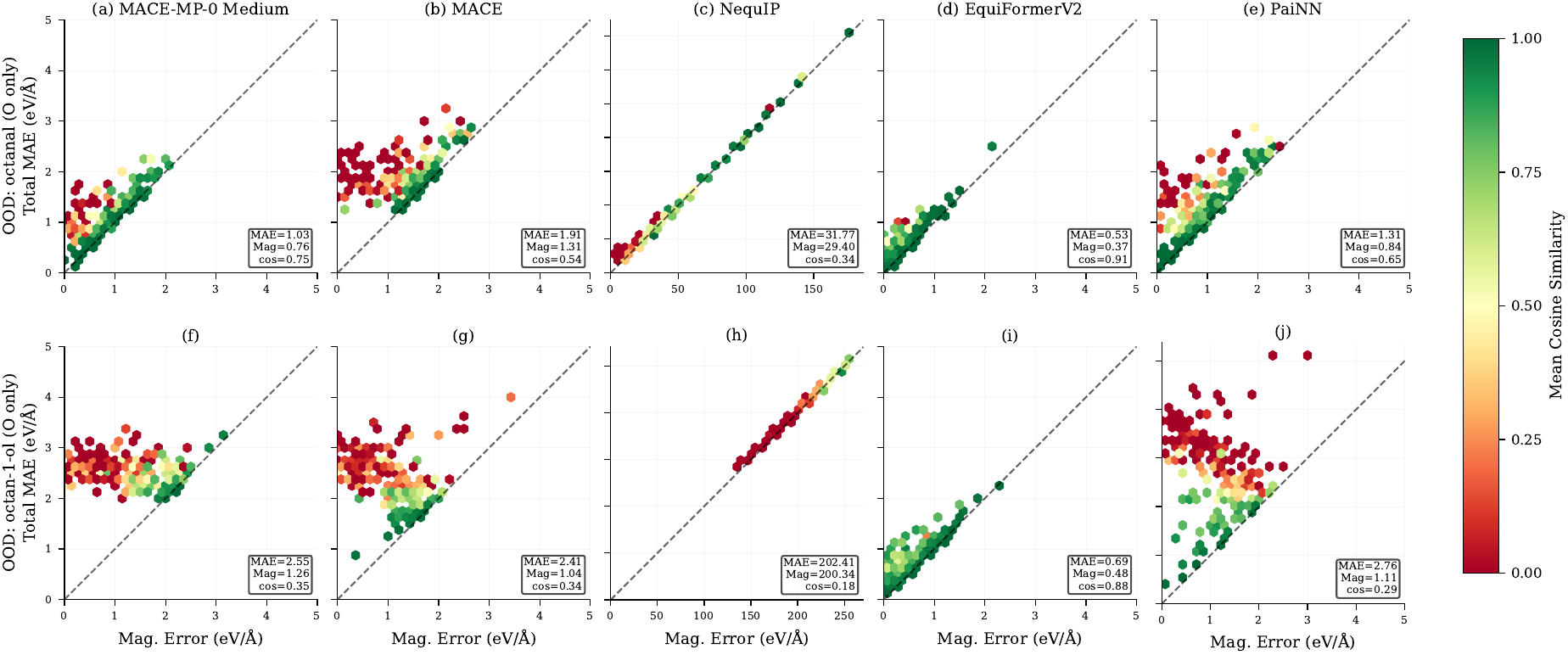}
        \caption{Fragment Fission: Oxygen}
        \label{fig:FF:decomp_o}
    \end{subfigure}
    \caption{Element-specific force error decomposition on the \emph{Fragment Fission} task for hydrogen (top), carbon (middle), and oxygen (bottom). Each block compares two out-of-distribution molecules drawn from different functional-group families: an alcohol (octan-1-ol) and an aldehyde (octanal), evaluated over 200 MD snapshots each.
    The x-axis shows the magnitude error $|\textit{norm}(\mathbf{F}_\text{pred}) - \textit{norm}(\mathbf{F}_\text{label})|$ and the y-axis shows the total force vector error $\textit{norm}(\mathbf{F}_\text{pred} - \mathbf{F}_\text{label})$, both in eV/\AA.
    Hexbin colour encodes mean cosine similarity (green $= 1.0$, red $= 0.0$). The dashed diagonal represents zero directional error; points above this line highlight where directional mismatch inflates the total MAE. Directional errors are concentrated on the oxygen atoms, while hydrogen and carbon remain close to the diagonal for most models.}
    \label{fig:FFI:decomp_all}
\end{figure}

The trajectory-level residuals discussed above average over all atoms in a molecule. 
Figure~\ref{fig:decompose_all_atoms} and Figures~\ref{fig:FCE:decomp_all}--\ref{fig:FFI:decomp_all} additionally decompose the force error onto individual atoms for representative ID/OOD molecule pairs from each of the five tasks, separating the magnitude error $|\textit{norm}(\mathbf{F}_\text{pred}) - \textit{norm}(\mathbf{F}_\text{label})|$ from the directional error captured by cosine similarity. 
This per-atom view reveals whether the generalisation failure is concentrated on specific atoms or distributed uniformly across the molecule.

The error distribution patterns closely follow the nature of the out-of-distribution shift. 
For tasks introducing novel functional group configurations (\emph{Fragment Fusion}, \emph{Fragment Duplication}, and \emph{Fragment Combination}), directional errors localise heavily on the heteroatoms defining the unseen chemical environment.
Specifically, in \emph{Fragment Fusion}, the failure concentrates on the oxygen and carbon atoms of the novel C(=O)OH group, while in \emph{Fragment Duplication}, carbon atoms remain well-predicted and the small residual directional error localises on the oxygen atoms of the duplicated group.
In \emph{Fragment Combination}, the directional accuracy is largely preserved across all elements, with \emph{NequIP} as the notable outlier exhibiting elevated errors on hydrogen and oxygen atoms. 
Nitrogen predictions remain accurate across most models despite nitrogen being underrepresented in the training distribution. 
Conversely, in \emph{Fragment Chain Extension}, where the OOD shift is purely an extrapolation of backbone length rather than a novel functional group, the error profile is fundamentally distinct: directional accuracy is preserved across all elements, while the magnitude error localises more prominently on the backbone carbon atoms.

This task-dependent pattern demonstrates that the generalisation failure is neither uniform nor a simple inability to model specific element types. 
For example, while oxygen atoms are well-predicted in alcohol and aldehyde training molecules, they become primary directional error sources when forced into the completely unseen \ce{C(=O)OH} configuration of a carboxylic acid in \emph{Fragment Fusion}. 
However, when that exact same \ce{C(=O)OH} local environment is present in the training set (\emph{Fragment Duplication}), the oxygen directional error is notably mitigated, even though the overall extensive prediction still fails. 
Furthermore, the resilience of nitrogen predictions in \emph{Fragment Combination}, even when introduced into a novel asymmetric pairing, suggests that certain local elemental representations transfer more robustly than others. 
Ultimately, directional errors systematically localise on the precise features that make the OOD environment chemically or structurally novel. 
When the local environment is familiar, the local directional vectors survive, and the failure is relegated to the global readout, an observation that perfectly aligns with the magnitude/extensive-readout framing developed in Section~\ref{main:evaluation:analysis}. 
We caution that this is a qualitative illustration drawn from individual trajectories rather than a quantitative population analysis; we include it because it links the aggregate generalisation failure to a specific, chemically interpretable subset of the atomic environments.
\label{details:per_atom_decomposition}

\begin{figure}[p]
    \centering
    \begin{subfigure}{\linewidth}
        \includegraphics[width=\linewidth]{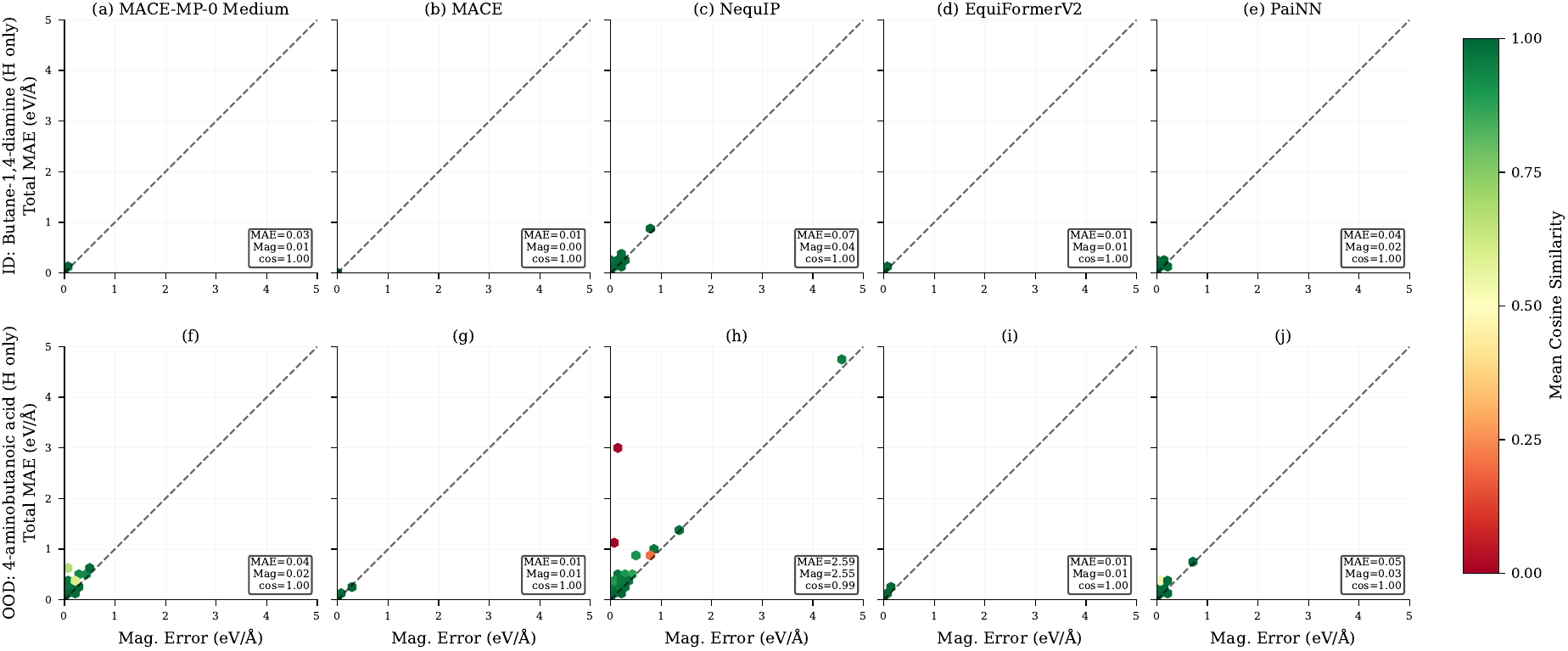}
        \caption{Fragment Combination: Hydrogen}
        \label{fig:FCB:decomp_h}
    \end{subfigure}
    \vspace{0.5cm}
    \begin{subfigure}{\linewidth}
        \includegraphics[width=\linewidth]{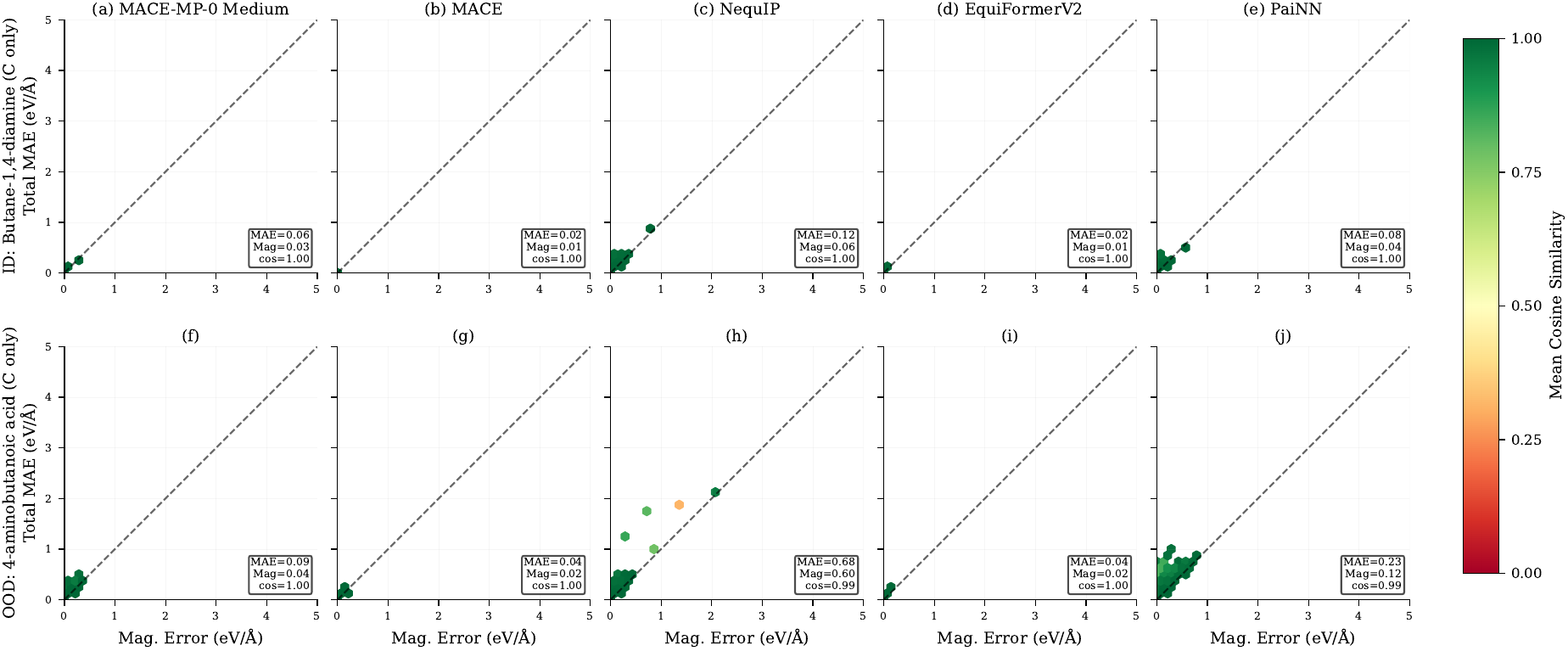}
        \caption{Fragment Combination: Carbon}
        \label{fig:FCB:decomp_c}
    \end{subfigure}
    \caption{Element-specific force error decomposition on the \emph{Fragment Combination} task for hydrogen (first), carbon (second), oxygen (third), and nitrogen (fourth). Each block compares in-distribution (ID: butane-1,4-diamine) and out-of-distribution (OOD: 4-aminobutanoic acid) predictions over 200 MD snapshots.
    The x-axis shows the magnitude error $|\textit{norm}(\mathbf{F}_\text{pred}) - \textit{norm}(\mathbf{F}_\text{label})|$ and the y-axis shows the total force vector error $\textit{norm}(\mathbf{F}_\text{pred} - \mathbf{F}_\text{label})$, both in eV/\AA.
    Hexbin colour encodes mean cosine similarity (green $= 1.0$, red $= 0.0$). The dashed diagonal represents zero directional error; points above this line highlight where directional mismatch inflates the total MAE. \emph{(Continued on next page.)}}
    \label{fig:FCB:decomp_all}
\end{figure}

\begin{figure}[p]
    \ContinuedFloat   
    \centering
    \begin{subfigure}{\linewidth}
        \includegraphics[width=\linewidth]{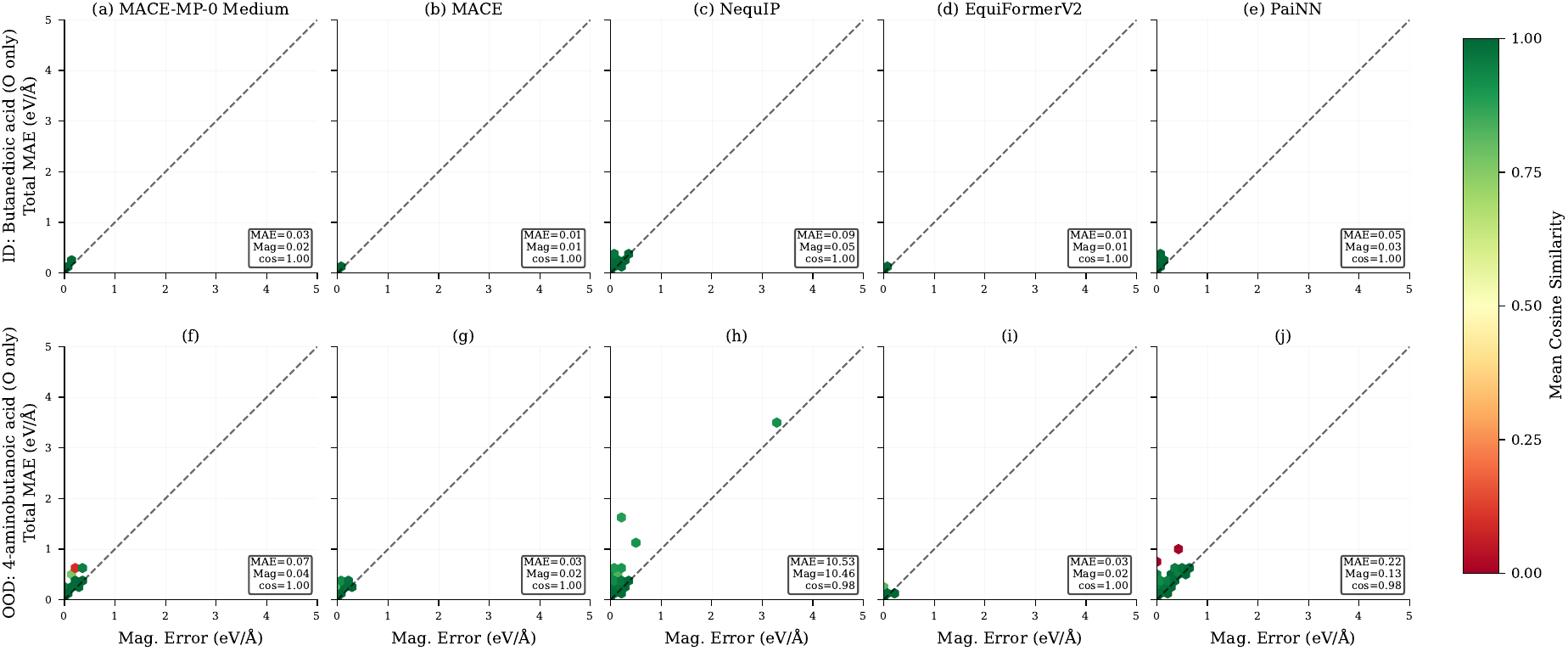}
        \caption{Fragment Combination: Oxygen}
        \label{fig:FCB:decomp_o}
    \end{subfigure}
    \vspace{0.5cm}
    \begin{subfigure}{\linewidth}
        \includegraphics[width=\linewidth]{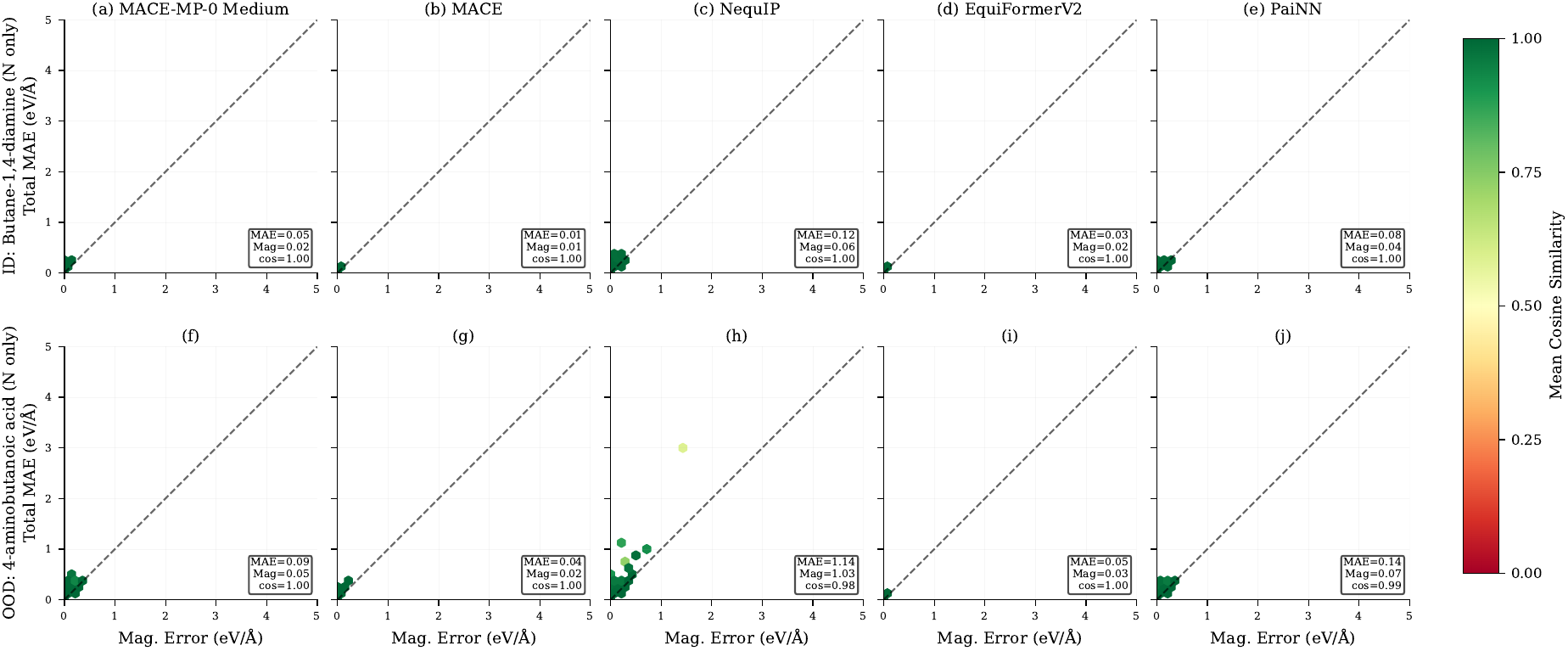}
        \caption{Fragment Combination: Nitrogen}
        \label{fig:FCB:decomp_n}
    \end{subfigure}
    \caption{(Continued.) Element-specific force error decomposition on the \emph{Fragment Combination} task for oxygen (third) and nitrogen (fourth). As above, the density of points near the origin is the result rather than a plotting artefact, with most atoms accurately predicted out-of-distribution.}
    \label{fig:FCB:decomp_all_cont}
\end{figure}

\clearpage

\subsection{Augmented task variants}
\label{details:augmented_settings}
\begin{figure*}[h]
    \centering
    \includegraphics[width=1.0\linewidth]{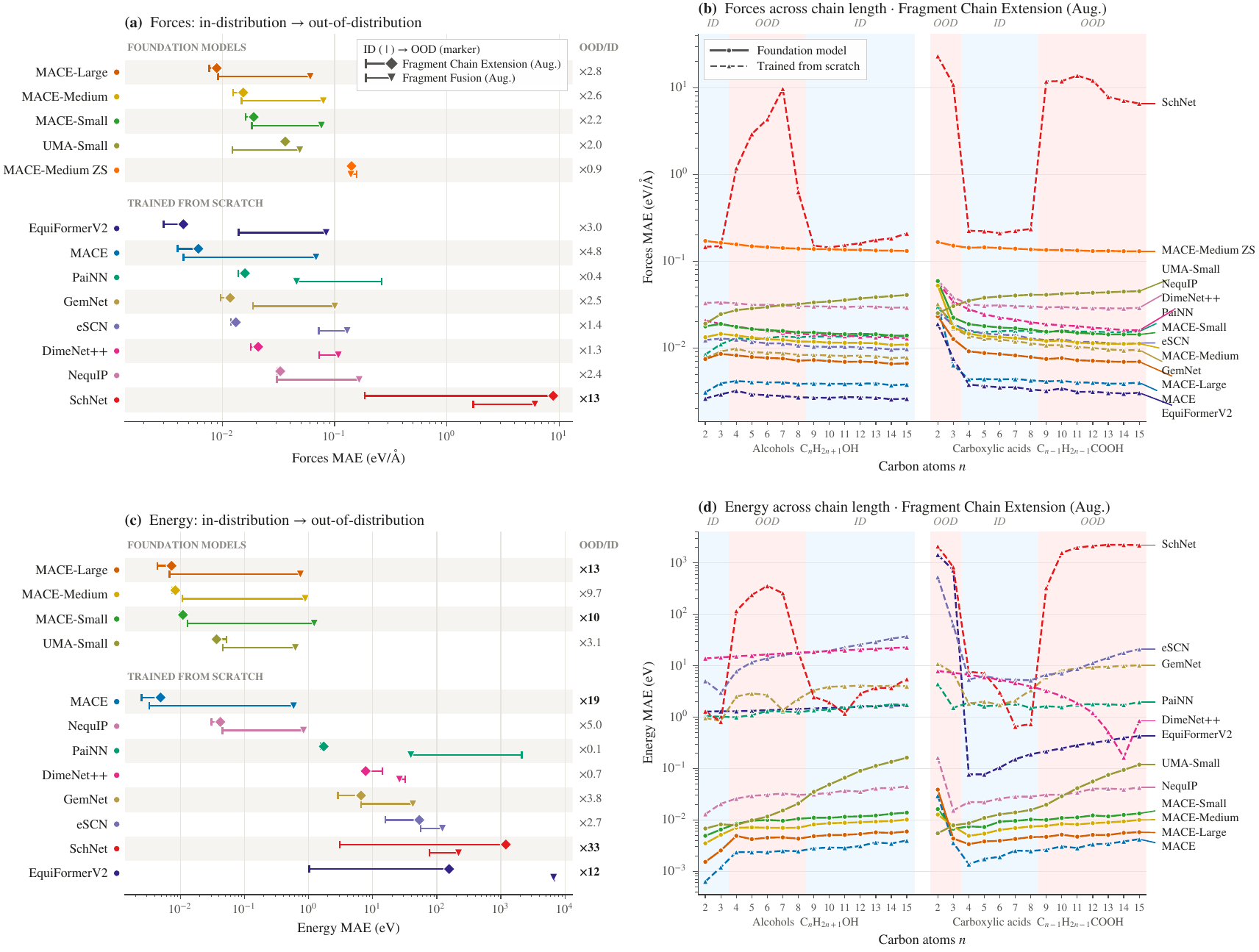}
    \caption{Performance on the two augmented task variants for all evaluated models.
    (a, c) In-distribution (ID) versus out-of-distribution (OOD) MAE for forces and total energy, on a logarithmic scale. As in Figure~\ref{fig:scatter_id_ood_forces_energy_mae}, each row is one model, grouped into foundation models and models trained from scratch, and for each task a bar runs from the model's ID error (vertical tick) to its OOD error (diamond: augmented Fragment Chain Extension; triangle: augmented Fragment Fusion). The right-hand column reports the geometric mean over the two augmented tasks of the ratio between OOD and ID error; values below $1\times$ indicate a lower error out-of-distribution than in-distribution, and values of $10\times$ or more are shown in bold. MACE-Medium ZS is omitted from (c), since zero-shot energies are not comparable to our reference.
    (b, d) Per-molecule MAE on augmented Fragment Chain Extension as a function of the number of carbon atoms $n$, for alcohols (left) and carboxylic acids (right). Shading marks the chain lengths of each family used for training (ID, blue) and held out for testing (OOD, red); solid lines denote foundation models and dashed lines models trained from scratch.
    Naming convention follows Figure~\ref{fig:scatter_id_ood_forces_energy_mae}.}
    \label{fig:aug_overview}
\end{figure*}

For Task 1 (Fragment Chain Extension) and Task 2 (Fragment Fusion), we additionally consider augmented variants in which the training set is enriched in a way that, in principle, should make the corresponding OOD test set easier to generalise to. The motivation is to disentangle two distinct sources of generalisation difficulty: (i) the lack of training coverage on a particular dimension (chain length or functional group), and (ii) the absence of a worked example of how the relevant compositional operation can be carried out. The augmented variants close gap (i) while leaving gap (ii) intact, allowing us to test whether models exploit the additional coverage to generalise more reliably, or whether the compositional shift remains the bottleneck. Figure~\ref{fig:molecules_aug} illustrates the molecules used in both augmented settings, and Figure~\ref{fig:aug_overview} summarises the results.


\subsubsection{Augmented Fragment Chain Extension}
\label{details:fce_aug}

\paragraph{Setup.} In the augmented variant of Fragment Chain Extension, the training set contains alcohols with carbon chain lengths in $\{2,3\}\cup\{9,\dots,15\}$ and carboxylic acids with carbon chain lengths in $\{4,\dots,8\}$.
The out-of-distribution test set contains alcohols with carbon chain lengths in $\{4,\dots,8\}$ and carboxylic acids with carbon chain lengths in $\{2,3\}\cup\{9,\dots,15\}$.
In other words, the model is tested on combinations of (length, functional group) that it has not seen during training, but every individual chain length and every individual functional group does appear somewhere in the training set.
The split is summarised in Table~\ref{tab:fce_aug_split}.

\begin{table}[h]
    \centering
    \footnotesize
    \setlength\tabcolsep{5pt}
    \caption{Training set and OOD test set for the augmented variant of Fragment Chain Extension.}
    \label{tab:fce_aug_split}
    \begin{tabular}{lll}
        \toprule
        \textbf{Task} & \textbf{Training Set} & \textbf{OOD Test Set} \\
        \midrule
        \multirow{2}{*}{Fragment chain extension (augmented)}
            & \textbullet~C2--C3 \& C9--C15 alcohols
            & \textbullet~C4--C8 alcohols \\
            & \textbullet~C4--C8 carboxylic acids
            & \textbullet~C2--C3 \& C9--C15 carboxylic acids \\
        \bottomrule
    \end{tabular}
\end{table}

Compared with the base variant, where the model never sees long carbon chains during training, the augmented variant exposes the model to chains across the full range of test lengths, just on a different functional group.
If the model has learned a transferable representation of how interatomic interactions scale with chain length, independent of the functional group present, the augmented setting should be substantially easier than the base setting.
If, instead, the model couples chain-length representations to functional-group identity, the additional training coverage will not help.


\paragraph{Results.}
The augmented variant is substantially easier than the base variant for every model except SchNet. 
On forces, the OOD errors of MACE, MACE-Large, EquiFormerV2, NequIP, and the fine-tuned MACE-MP-0 variants all show short bars, with OOD/ID factors close to 1, in Figure~\ref{fig:aug_overview}a, in contrast to the base variant where DimeNet++ and SchNet failed by orders of magnitude. 
DimeNet++ now generalises in line with the other architectures, whereas SchNet remains an outlier on both forces and energy. 
The energy results follow the same pattern: most models close the gap that was several orders of magnitude in the base variant, with SchNet again the lone exception. 
Per-chain-length plots (Figure~\ref{fig:aug_overview}b) confirm that, for all non-SchNet models, errors are essentially flat across the full chain-length range on both alcohols and carboxylic acids; SchNet exhibits sharp spikes precisely in the OOD region for both functional groups.

\paragraph{Analysis.}
The augmented variant substantially reduces the OOD error for every architecture except SchNet, indicating that the failures observed for several models in the base variant of Task~1 are attributable to the absence of long-chain training examples rather than to a fundamental inability to generalise on longer the structures.
Once chains of every test length appear during training, even on a different functional group, models all transfer chain-length information across the alcohol/carboxylic-acid boundary effectively. 
The residual gap visible for some of the from-scratch models in Figure~\ref{fig:aug_overview}(a) is small and reflects the cost of transferring chain-length information across functional-group boundaries rather than a categorical generalisation failure.

\clearpage

\subsubsection{Augmented Fragment Fusion}
\label{details:fc_aug}

\paragraph{Setup.} In the augmented variant of Fragment Fusion, the training data is expanded to include amines and amides in addition to the original alcohols, aldehydes, complex carbonyls, and complex alcohols. The OOD test set continues to consist of carboxylic acids. The motivation is that the amide functional group can itself be viewed as a composition of an aldehyde-derived carbonyl and an amine, so the augmented training set contains a worked example of how two simple functional groups can be combined into a more complex composite. The split is summarised in Table~\ref{tab:fc_aug_split}.

\begin{table}[h]
    \centering
    \footnotesize
    \setlength\tabcolsep{5pt}
    \caption{Training set and OOD test set for the augmented variant of Fragment Fusion.}
    \label{tab:fc_aug_split}
    \begin{tabular}{lll}
        \toprule
        \textbf{Task} & \textbf{Training Set} & \textbf{OOD Test Set} \\
        \midrule
        \multirow{3}{*}{Fragment Fusion (augmented)}
            & \textbullet~C7--C11 complex carbonyls and complex alcohols
            & \multirow{3}{*}{C4--C10 carboxylic acids} \\
            & \textbullet~C4--C10 alcohols and aldehydes
            & \\
            & \textbullet~C4--C10 amines and amides
            & \\
        \bottomrule
    \end{tabular}
\end{table}

The base variant of Task~2 asks the model to compose a carboxylic acid (C(=O)OH) from the alcohol (--OH) and aldehyde (C=O) functional groups it has seen during training. 
The augmented variant additionally provides amides (C(=O)N), which themselves arise from a carbonyl-plus-heteroatom composition analogous to the carboxylic acid case. 
If the model can generalise from one demonstrated composition to another, the augmented variant should reduce the OOD error on carboxylic acids; if instead the model treats each functional group as a standalone token without abstracting the underlying composition operation, the additional amide training data will not help.

\paragraph{Results.}
In contrast to the augmented variant of Task~1, the augmented variant of Task~2 does not substantially reduce the OOD error on carboxylic acids relative to the base variant. 
On forces, MACE, MACE-Large, and UMA-Small all show OOD errors in the augmented setting that are comparable to those of the base variant, and across the from-scratch architectures the ID-to-OOD ratio remains large and essentially indistinguishable from what we observed in the base variant of Task~2 (Figure~\ref{fig:aug_overview}(a), triangles). 
The energy picture is similar: MACE's OOD energy MAE on carboxylic acids remains orders of magnitude above its ID error, mirroring the gap of the base variant. 
The fine-tuned foundation models (MACE-MP-0 at all sizes, UMA-Small) show the same qualitative behaviour as in the base variant and do not consistently outperform the from-scratch MACE.

\paragraph{Analysis.}
Despite training data that explicitly demonstrates how a carbonyl can be combined with a heteroatom-bearing fragment to form a composite functional group, models continue to fail on the structurally analogous carboxylic-acid case. 
This suggests that current MLIPs do not abstract a transferable composition operation from amide training examples, but rather memorise the amide as a distinct chemical entity in its own right. 
The compositional structure that would allow the model to reason ``carboxylic acid is to alcohol+aldehyde as amide is to amine+aldehyde'' is not learned. 
Read together with the augmented Fragment Chain Extension results, this gives a relatively clean picture of where current MLIPs do and do not generalise: when the missing dimension of the OOD test set is one of \emph{coverage}, a chain length or local environment that simply did not appear during training, additional training data on that dimension closes the gap, even when supplied through a different functional group; when the missing dimension is \emph{compositional}, a way of combining two known fragments into a novel one, additional worked examples of the composition operation do not transfer.

\clearpage

\subsection{The dataset}
\label{details:dataset}
The released GMD dataset comprises 118 molecules with approximately 2{,}000 labelled configurations per trajectory ($>$296{,}000 in total). 
The splitting strategy is described in Section~\ref{main:GMD:benchmark}. 
Table~\ref{tab:splitting_strategy_multirow} in the main paper, together with Tables~\ref{tab:fce_aug_split} and~\ref{tab:fc_aug_split} in this appendix, specifies what molecules are used for each task and variant. 
Figure~\ref{fig:molecules_tasks} illustrates a representative molecular example for each of the five tasks, and Figure~\ref{fig:dataset_molecules} provides the complete list of molecules in the dataset.

Unlike common-practice MD datasets (e.g.\ MD17~\cite{chmiela2017machine}), whose trajectories are sampled near equilibrium minima, GMD trajectories are generated to capture a wide range of off-equilibrium configurations~\cite{bigi2025flashmd}.
To align with standard benchmarking protocols (which often recommend training on around 1000 snapshots from a single trajectory\footnote{\url{https://pytorch-geometric.readthedocs.io/en/2.6.0/generated/torch_geometric.datasets.MD17.html}}), we sample a 1000-snapshot subset per trajectory for the training set. Drawing this subset from a broader, highly diverse trajectory ensures sufficient representation of high-energy states and robustly exposes the models to the distinct dynamics of multiple molecular systems without requiring excessive training budgets.

\clearpage

\subsection{Training details}
\label{details:trainingDetails}
All computational workloads, including training from-scratch models, fine-tuning foundation models, and hyperparameter optimisation, were executed across a mixed hardware cluster comprising NVIDIA GH200, RTX Ada A6000, RTX A100, RTX 5090, and RTX 4090 GPUs. 
For the foundation models (MACE-MP-0 at small, medium, and large scales, and UMA-Small), we performed full-weight fine-tuning on each task's training set using the default optimisation settings from the respective original repositories (learning rate $1 \times 10^{-4}$, no layer freezing, no LoRA-style adapters). 
We chose full fine-tuning rather than parameter-efficient alternatives to provide an upper bound on achievable out-of-distribution performance; parameter-efficient fine-tuning would be expected to perform no better.
Prior to fine-tuning, the atomic reference energies ($E_0$) of each foundation-model checkpoint were reestimated against our own reference data, so that the per-element energy baseline corresponds to the level of theory used for our labels rather than to that of the model's pre-training corpus. Without this step the fine-tuning objective must first absorb a large constant per-element offset before it can begin fitting the chemistry, and residual energy error would not be attributable to the model. This means that our fine-tuned foundation models are evaluated against exactly the reference data they were fit to reproduce, so that the level-of-theory distinction between our Gaussian-basis labels and the plane-wave data used in pre-training does not enter these comparisons; it remains relevant only for the zero-shot results, which we discuss in Appendix~\ref{app:level_of_theory}. Correct $E_0$ initialisation has since been identified independently as a prerequisite for reliable fine-tuning outcomes in this model class~\cite{tompa2026fine}.
For the models trained from scratch, training set sizes ranged from 5{,}000 configurations (e.g., Task~1, 5 molecules $\times$ 1{,}000 snapshots) up to larger sets depending on the specific generalisation task.

To provide context on the computational overhead of these models, Table~\ref{tab:estimated_training_times} summarises the approximate training times for each architecture on our hardware setup. 
Times are reported as approximate per-task averages; actual durations vary across the four tasks because training-set sizes differ, with the Fragment Fusion task taking the longest due to its expanded training set and the Fragment Chain Extension task taking the shortest.

\begin{table}[h]
    \centering
    \footnotesize
    \caption{Estimated training or fine-tuning times for the evaluated MLIP architectures, averaged across the four GMD tasks. Per-task times vary with training-set size; see body text.}
    \label{tab:estimated_training_times}
    \begin{tabular}{llc}
    \toprule
    \textbf{Model} & \textbf{Estimated Training Time} & \textbf{GPU(s)} \\
    \midrule
    SchNet & 2 days & RTX 4090 \\
    PaiNN & 3 days & RTX 5090 \\
    DimeNet$^{++}$ & 2 days & RTX 4090 \\
    GemNet-OC & 4 days & RTX 5090 \\
    EquiFormerV2 & 6 days & RTX Ada A6000\\
    eSCN & 6 days & RTX Ada A6000 \\
    MACE (From scratch) & 2 days & RTX Ada A6000 \\
    UMA-Small (Fine-tuned) & 6 days & [4, 8]$\times$ GH200 \\
    MACE-MP-0 Small (Fine-tuned) & 7 days & 4$\times$ GH200 \\
    MACE-MP-0 Medium (Fine-tuned) & 8 days & 4$\times$ GH200 \\
    MACE-MP-0 Large (Fine-tuned) & 10 days & 4$\times$ GH200 \\
    \bottomrule
    \end{tabular}
\end{table}

\paragraph{Statistical reporting.} All reported metrics across both from-scratch and foundation models are averaged over three independent validation/test splits drawn from the held-out trajectories. The variance reflected by these splits captures sampling uncertainty in the test set rather than training-process variability across independent random seeds.

\subsection{Foundation models fine-tuning}
\label{details:foundation_verdict}

\begin{figure*}[t]
    \centering
    \includegraphics[width=1.0\linewidth]{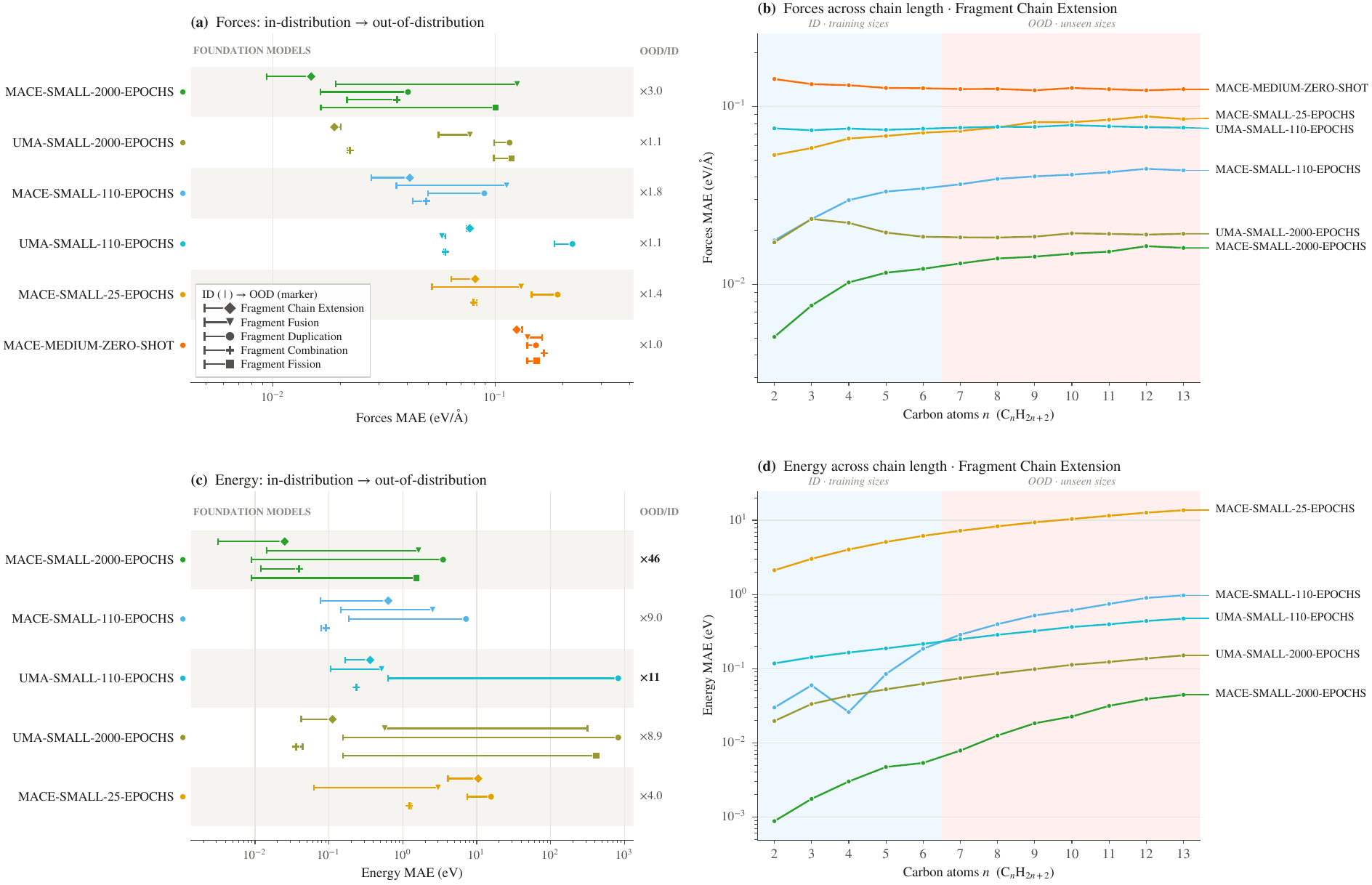}
    \caption{Foundation-model fine-tuning regime across the five GMD tasks: MACE-MP-0 Small at three fine-tuning budgets (25, 110 and 2000 epochs) and UMA-Small at two (110 and 2000 epochs), alongside the zero-shot MACE-MP-0 Medium baseline.
    (a, c) In-distribution (ID) versus out-of-distribution (OOD) MAE for forces and total energy, on a logarithmic scale. As in Figure~\ref{fig:scatter_id_ood_forces_energy_mae}, each row is one model variant, and for each task a bar runs from its ID error (vertical tick) to its OOD error (task marker); the right-hand column reports the geometric mean over the five tasks of the ratio between OOD and ID error, with values of $10\times$ or more in bold. The zero-shot baseline is omitted from (c), since zero-shot energies are not comparable to our reference.
    (b, d) Per-molecule MAE on Fragment Chain Extension as a function of the number of carbon atoms $n$, with training sizes ($n=2$--$6$) shaded blue and unseen sizes ($n=7$--$13$) shaded red.}
    \label{fig:foundation_verdict}
\end{figure*}

The fine-tuning budget chosen for the foundation-model results in the main paper (2000 epochs) is the result of a comparison across smaller budgets. Figure~\ref{fig:foundation_verdict} shows MACE-MP-0 Small at three fine-tuning budgets (25, 110, and 2000 epochs) and UMA-Small at two (110 and 2000 epochs), alongside the zero-shot MACE-MP-0 Medium baseline, on the five GMD tasks. Three observations follow.
 
First, increasing the fine-tuning budget improves both in-distribution and out-of-distribution errors monotonically, on both forces and energy, for both foundation-model families. The 25-epoch MACE-MP-0 Small is closest to the zero-shot baseline; 110 epochs already lowers both errors substantially, and 2000 epochs reaches the lowest errors achieved by these variants. Per-chain-length traces on Fragment Chain Extension (panels b and d) show the same picture: the curves shift downward roughly uniformly across the in-distribution and out-of-distribution chain ranges as the budget grows.
 
Second, additional fine-tuning lowers the absolute error level but does not close the gap between in-distribution and out-of-distribution performance. For MACE-MP-0 Small, the OOD/ID factor even grows with the budget, from $1.4\times$ at 25 epochs to $3.0\times$ at 2000 epochs for forces, and from $4.0\times$ to $46\times$ for energy, because in-distribution errors fall faster than out-of-distribution errors; for UMA-Small it remains essentially unchanged. Longer fine-tuning therefore produces a more accurate model on both distributions, not a more compositionally robust one.
 
Third, even at 2000 epochs, the foundation-model variants do not consistently outperform the from-scratch MACE reported in the main paper, particularly on the compositional tasks (Fragment Fusion, Fragment Duplication). This is the regime in which we report the main-paper foundation-model numbers: a budget large enough that further fine-tuning would be unlikely to change the qualitative conclusions, while still falling short of from-scratch MACE on the tasks that most clearly probe compositional generalisation.
 
We chose 2000 epochs as our reporting point on this basis: it represents an upper bound on what fine-tuning at this scale can reasonably be expected to achieve in our setting, given the additional cost of pushing further is unlikely to alter the conclusions of the benchmark. We caution that this regime is computationally expensive given the size of these foundation models, and is reported to characterise the ceiling of fine-tuning rather than to recommend it as a typical practice.

\subsection{Model hyperparameters}\label{details:hyperparameters}

For the from-scratch architectures, we adopted a two-stage hyperparameter tuning strategy. Initially, models were trained using the curated default hyperparameters provided in the \texttt{fairchem} repository to establish a baseline performance. Subsequently, we conducted Bayesian hyperparameter optimisation using Weights \& Biases (W\&B) sweeps to ensure that each model achieved its best possible performance on the in-distribution validation data within our computational allowance, and to analyse the sensitivity of the models to hyperparameter selection.

We utilised the W\&B count parameter to limit the number of optimisation runs in each setting, allocating 30 runs per hyperparameter for each model. This resulted in 180 runs for EquiFormerV2, 120 runs for DimeNet$^{++}$, 150 runs for GemNet-OC, 90 runs for PaiNN, and 120 runs for SchNet. The search domains and the final optimised parameters for all evaluated architectures—including eSCN—are consolidated in Table~\ref{tab:hyperparams-fragment-chain-extension}, ensuring the reproducibility of our findings. 

Overall, our hyperparameter sweeps indicated that model capacity was not the primary bottleneck for OOD failure; in fact, smaller architectures frequently outperformed larger ones on OOD energy predictions.

\begingroup
\footnotesize
\setlength\tabcolsep{4pt}
\begin{longtable}{l l p{5.4cm} l}
\caption{Hyperparameter search domains and selected optimal values from Bayesian optimisation for the Fragment Chain Extension task.}
\label{tab:hyperparams-fragment-chain-extension} \\
\toprule
\textbf{Model} & \textbf{Parameter} & \textbf{Search domain} & \textbf{Optimal} \\
\midrule
\endfirsthead

\multicolumn{4}{c}{\textbf{Table \ref{tab:hyperparams-fragment-chain-extension} (continued)}} \\
\toprule
\textbf{Model} & \textbf{Parameter} & \textbf{Search domain} & \textbf{Optimal} \\
\midrule
\endhead

\midrule
\multicolumn{4}{r}{\textit{Continued on next page}} \\
\endfoot

\bottomrule
\endlastfoot

SchNet & Hidden ch.     & q-log-unif. [256,2048] (step 128) & 512     \\*
       & Num filters    & q-log-unif. [64,1024] (step 64)   & 256     \\
       & Num interact.  & int-unif. [3,6]                    & 4       \\
       & Num gaussians  & \{50,100,200,300\}                 & 300     \\
       & Cutoff         & unif. [4.0,13.0]                   & 11.51   \\
       & LR init        & log-unif. [5e-5,2e-3]              & 1.24e-3 \\
       & Warmup frac.   & unif. [0.01,0.1]                   & 0.062   \\
       & LR min factor  & unif. [0.001,0.1]                  & 0.0222  \\
\midrule

PaiNN  & Hidden ch.     & q-log-unif. [128,4096] (step 64)  & 1984    \\*
       & Num layers     & int-unif. [3,6]                    & 6       \\
       & Num RBF        & \{64,128,256,512\}                 & 256     \\
       & Cutoff         & unif. [4.0,12.0]                   & 10.38   \\
       & Max neigh.     & \{20,50,100,150\}                  & 100     \\
       & LR init        & log-unif. [1e-5,1e-3]              & 1.27e-4 \\
       & Weight decay   & log-unif. [1e-6,1e-3]              & 1.3e-4  \\
       & Warmup frac.   & unif. [0.01,0.1]                   & 0.081   \\
       & LR min factor  & unif. [0.001,0.1]                  & 0.0867  \\
       & Clip grad norm & \{5,10,20,50\}                     & 10      \\
       & EMA decay      & unif. [0.99,0.9999]                & 0.9905  \\
\midrule

GemNet & Num blocks          & \{4,5,6,7\}               & 4       \\*
       & Emb size atom       & \{128,256\}               & 128     \\
       & Emb size edge       & \{512,1024\}              & 1024    \\
       & Num before skip     & \{1,2,3\}                 & 2       \\
       & Num after skip      & \{1,2,3\}                 & 2       \\
       & Quad interact.      & \{T,F\}                   & T       \\
       & Atom-edge interact. & \{T,F\}                   & T       \\
       & Edge-atom interact. & \{T,F\}                   & T       \\
       & Atom interact.      & \{T,F\}                   & T       \\
       & Emb size trip in    & \{32,64,128\}             & 32      \\
       & Emb size quad in    & \{32,64\}                 & 64      \\
       & Emb size aint in    & \{32,64\}                 & 32      \\
       & Cutoff              & unif. [4.0,13.0]          & 10.41   \\
       & Cutoff qint         & unif. [4.0,13.0]          & 11.13   \\
       & Cutoff aeaint       & unif. [4.0,13.0]          & 12.99   \\
       & Max neigh.          & \{20,30,40\}              & 40      \\
       & Max neigh. qint     & \{4,8,12\}                & 12      \\
       & Max neigh. aeaint   & \{10,20,30\}              & 20      \\
       & Max neigh. aint     & \{500,1000\}              & 1000    \\
       & Num spherical       & \{5,7,9\}                 & 7       \\
       & Num radial          & \{64,128\}                & 128     \\
       & Envelope exp.       & \{4,5,6\}                 & 6       \\
       & LR init             & log-unif. [5e-5,5e-4]     & 2.88e-4 \\
       & Weight decay        & log-unif. [1e-6,1e-3]     & 1.18e-5 \\
       & Warmup frac.        & unif. [0.005,0.05]        & 0.033   \\
       & LR min factor       & unif. [0.001,0.1]         & 0.0112  \\
       & Warmup factor       & \{0.1,0.2,0.3\}           & 0.2     \\
       & Clip grad norm      & \{5,10,20\}               & 20      \\
       & EMA decay           & unif. [0.99,0.9999]       & 0.9959  \\
\midrule

DimeNet$^{++}$ & Hidden ch.      & \{256,384,512,640,768\} & 256     \\*
               & Out emb. ch.    & \{192,256,384\}         & 256     \\
               & Num blocks      & \{3,4,5,6,7,8\}         & 5       \\
               & Num before skip & \{1,2,3\}               & 3       \\
               & Num after skip  & \{1,2,3\}               & 3       \\
               & Num output lay. & \{2,3,4\}               & 4       \\
               & Num radial      & \{6,8,10\}              & 8       \\
               & Num spherical   & \{5,7,9\}               & 9       \\
               & Cutoff          & unif. [4.0,13.0]        & 6.07    \\
               & LR init         & log-unif. [5e-5,1e-3]   & 5.26e-4 \\
               & Warmup frac.    & unif. [0.01,0.1]        & 0.031   \\
               & Warmup factor   & \{0.1,0.2,0.3\}         & 0.3     \\
               & LR min factor   & unif. [0.001,0.05]      & 0.00825 \\
\midrule

EquiFormerV2 & Num layers      & int-unif. [6,10]                   & 9       \\*
             & Sphere ch.      & q-log-unif. [128,512] (step 64)    & 448     \\
             & Attn hidden ch. & \{64,96,128\}                      & 64      \\
             & Num heads       & \{4,8\}                            & 4       \\
             & FFN hidden ch.  & q-log-unif. [384,768] (step 64)    & 704     \\
             & Edge ch.        & q-log-unif. [512,1024] (step 128)  & 640     \\
             & Max neigh.      & \{16,20,24,28,32\}                 & 16      \\
             & Max radius      & unif. [10.0,15.0]                  & 11.10   \\
             & Alpha drop      & \{0.0,0.1\}                        & 0.1     \\
             & Drop path rate  & \{0.0,0.05,0.1\}                   & 0.0     \\
             & Norm type       & \{rms\_norm\_sh, layer\_norm\_sh, layer\_norm\} & layer\_norm\_sh \\
             & Weight init     & \{uniform, normal\}                & normal  \\
             & LR init         & log-unif. [1e-5,4e-4]              & 2.5e-5  \\
             & Weight decay    & log-unif. [1e-6,1e-3]              & 7.3e-4  \\
             & Warmup frac.    & unif. [0.01,0.1]                   & 0.066   \\
             & LR min factor   & unif. [0.001,0.1]                  & 0.0337  \\
             & Clip grad norm  & \{20,50,100\}                      & 20      \\
             & EMA decay       & unif. [0.99,0.9999]                & 0.9953  \\
\midrule

eSCN & Num layers         & int-unif. [10,20]               & 16      \\*
     & Sphere ch.         & q-log-unif. [96,256] (step 32)  & 96      \\
     & Hidden ch.         & q-log-unif. [256,512] (step 64) & 320     \\
     & Max neigh.         & \{16,20,24,28,32\}              & 32      \\
     & Cutoff             & unif. [4.0,13.0]                & 8.63    \\
     & Lmax list          & \{[4],[6],[8]\}                 & [8]     \\
     & Mmax list          & \{[2],[3]\}                     & [3]     \\
     & Num sphere samples & \{128,256,384,512\}             & 256     \\
     & Basis width scalar & unif. [1.0,3.0]                 & 1.23    \\
     & LR init            & log-unif. [1e-4,2e-3]           & 6.97e-4 \\
     & Weight decay       & log-unif. [1e-6,1e-3]           & 9.08e-4 \\
     & Warmup frac.       & unif. [0.005,0.05]              & 0.021   \\
     & LR min factor      & unif. [0.001,0.1]               & 0.0936  \\
     & Warmup factor      & \{0.1,0.2,0.3\}                 & 0.3     \\
     & Clip grad norm     & \{10,20,50\}                    & 20      \\
     & EMA decay          & unif. [0.99,0.9999]             & 0.9901  \\
\end{longtable}
\endgroup

\subsection{Detailed results}
\label{details:numerical_results}

This section provides the comprehensive numerical results corresponding to the visual summaries presented in the main paper and preceding appendices. Rather than fragmenting the data across multiple tables, we have consolidated the exact performance metrics—including Mean Absolute Error (MAE) for energy and forces, alongside forces cosine similarity and magnitude MAE—for all evaluated architectures across every generalisation task.

Specifically, Tables~\ref{tab:numericals_fmae_emae}-\ref{tab:numericals_cosine_fmag} present the complete in-distribution (ID) and out-of-distribution (OOD) metrics for all five core components of the GMD benchmark.
This unified format allows for a direct, granular comparison of generalisation capabilities and failure modes across all models and tasks simultaneously.

\clearpage
\begingroup
\footnotesize
\setlength\tabcolsep{4pt}
\begin{longtable}{l r@{\,$\pm$\,}l r@{\,$\pm$\,}l r@{\,$\pm$\,}l r@{\,$\pm$\,}l}
\caption{\footnotesize Comparison of MLIP models on the five GMD tasks: Fragment Chain Extension, Fragment Fusion, Fragment Duplication, Fragment Combination and Fragment Fission. Metrics reported are Forces Mean Absolute Error (\emph{F MAE}) and Energy Mean Absolute Error (\emph{E MAE}). Results are averaged across all molecules for each task. ID and OOD indicate in-distribution and out-of-distribution splits, respectively. Energy is in eV and forces in eV/\AA. $^{\dagger}$Fragment Fission uses the Fragment Duplication models and in-distribution test set, so its ID values are those of Fragment Duplication. MACE-Small-25, MACE-Small-110 and UMA-Small-110 were not evaluated on this task.}
\label{tab:numericals_fmae_emae} \\
\toprule
 & \multicolumn{4}{c}{\textbf{F MAE}} & \multicolumn{4}{c}{\textbf{E MAE}} \\
\cmidrule(lr){2-5}\cmidrule(lr){6-9}
\textbf{Model} & \multicolumn{2}{c}{ID} & \multicolumn{2}{c}{OOD} & \multicolumn{2}{c}{ID} & \multicolumn{2}{c}{OOD} \\
\midrule
\endfirsthead

\multicolumn{9}{c}{\textbf{Table \ref{tab:numericals_fmae_emae} (continued)}} \\
\toprule
 & \multicolumn{4}{c}{\textbf{F MAE}} & \multicolumn{4}{c}{\textbf{E MAE}} \\
\cmidrule(lr){2-5}\cmidrule(lr){6-9}
\textbf{Model} & \multicolumn{2}{c}{ID} & \multicolumn{2}{c}{OOD} & \multicolumn{2}{c}{ID} & \multicolumn{2}{c}{OOD} \\
\midrule
\endhead

\midrule
\multicolumn{9}{r}{\textit{Continued on next page}} \\
\endfoot

\bottomrule
\endlastfoot

\multicolumn{9}{l}{\textit{Fragment Chain Extension}} \\*
SchNet          & 0.0345 & 0.0009 & 1.6836 & 0.0822 & 1.0622 & 0.0230 & 82.638 & 0.5178 \\
PaiNN           & 0.0088 & 0.0000 & 0.0682 & 0.0005 & 0.2604 & 0.0010 & 35.113 & 0.3108 \\
GemNet          & 0.0037 & 0.0000 & 0.0133 & 0.0002 & 0.2894 & 0.0020 & 223.20 & 1.5317 \\
EquiFormerV2    & 0.0015 & 0.0000 & 0.0040 & 0.0000 & 0.4155 & 0.0044 & 407.78 & 2.7412 \\
DimeNet$^{++}$  & 0.0030 & 0.0004 & 1.4118 & 0.0100 & 0.0933 & 0.0011 & 354.66 & 2.1894 \\
NequIP          & 0.0151 & 0.0000 & 0.0278 & 0.0003 & 0.0108 & 0.0001 & 0.0576 & 0.0005 \\
eSCN            & 0.0026 & 0.0001 & 0.0312 & 0.0005 & 0.1955 & 0.0097 & 151.45 & 2.1809 \\
MACE            & 0.0034 & 0.0000 & 0.0061 & 0.0001 & 0.0016 & 0.0001 & 0.0087 & 0.0001 \\
MACE-Large      & 0.0052 & 0.0000 & 0.0089 & 0.0001 & 0.0084 & 0.0001 & 0.0246 & 0.0001 \\
MACE-Medium ZS  & 0.1319 & 0.0003 & 0.1245 & 0.0010 & 4233.1 & 0.0004 & 4254.0 & 0.0004 \\
MACE-Medium     & 0.0083 & 0.0000 & 0.0137 & 0.0002 & 0.0028 & 0.0001 & 0.0329 & 0.0001 \\
MACE-Small-25   & 0.0633 & 0.0000 & 0.0811 & 0.0010 & 4.0941 & 0.0033 & 10.465 & 0.0009 \\
MACE-Small-110  & 0.0276 & 0.0000 & 0.0411 & 0.0007 & 0.0770 & 0.0029 & 0.6333 & 0.0007 \\
MACE-Small-2000 & 0.0093 & 0.0000 & 0.0148 & 0.0002 & 0.0031 & 0.0001 & 0.0251 & 0.0001 \\
UMA-Small-110   & 0.0743 & 0.0001 & 0.0766 & 0.0010 & 0.1655 & 0.0020 & 0.3614 & 0.0002 \\
UMA-Small-2000  & 0.0201 & 0.0001 & 0.0188 & 0.0002 & 0.0422 & 0.0005 & 0.1117 & 0.0000 \\
\midrule

\multicolumn{9}{l}{\textit{Fragment Fusion}} \\*
SchNet          & 2.9222 & 0.0080 & 9.0294 & 0.0984 & 142.64 & 1.2772 & 355.36 & 1.2042 \\
PaiNN           & 0.0159 & 0.0000 & 0.1183 & 0.0014 & 12.964 & 0.0116 & 7.5822 & 0.0331 \\
GemNet          & 0.0247 & 0.0000 & 0.1218 & 0.0004 & 4.8028 & 0.0439 & 16.759 & 0.0806 \\
EquiFormerV2    & 0.0082 & 0.0000 & 0.1385 & 0.0006 & 7719.1 & 0.0009 & 6721.4 & 0.0582 \\
DimeNet$^{++}$  & 0.0873 & 0.0002 & 0.3106 & 0.0020 & 151.12 & 0.0012 & 567.24 & 0.0141 \\
NequIP          & 0.0388 & 0.0001 & 0.2286 & 0.0017 & 0.1063 & 0.0009 & 1.4876 & 0.0043 \\
eSCN            & 0.0557 & 0.0002 & 0.1415 & 0.0004 & 44.785 & 0.7286 & 315.44 & 1.8915 \\
MACE            & 0.0047 & 0.0000 & 0.0889 & 0.0001 & 0.0036 & 0.0000 & 0.9734 & 0.0005 \\
MACE-Large      & 0.0098 & 0.0000 & 0.0648 & 0.0002 & 0.0076 & 0.0000 & 1.0584 & 0.0007 \\
MACE-Medium ZS  & 0.1622 & 0.0001 & 0.1394 & 0.0003 & 13506. & 0.1616 & 114438 & 0.3899 \\
MACE-Medium     & 0.0155 & 0.0000 & 0.1448 & 0.0002 & 0.0116 & 0.0002 & 1.8469 & 0.0021 \\
MACE-Small-25   & 0.0518 & 0.0000 & 0.1305 & 0.0001 & 0.0621 & 0.0001 & 2.9924 & 0.0020 \\
MACE-Small-110  & 0.0359 & 0.0000 & 0.1122 & 0.0004 & 0.1454 & 0.0003 & 2.5275 & 0.0014 \\
MACE-Small-2000 & 0.0191 & 0.0000 & 0.1252 & 0.0008 & 0.0142 & 0.0001 & 1.6275 & 0.0023 \\
UMA-Small-110   & 0.0595 & 0.0002 & 0.0574 & 0.0002 & 0.1050 & 0.0004 & 0.5155 & 0.0007 \\
UMA-Small-2000  & 0.0150 & 0.0000 & 0.0659 & 0.0001 & 0.0547 & 0.0001 & 0.7841 & 0.0010 \\
\midrule

\multicolumn{9}{l}{\textit{Fragment Duplication}} \\*
SchNet          & 0.4452 & 0.0025 & 1.1944 & 0.0051 & 9.2223 & 0.0806 & 3928.8 & 0.3185 \\
PaiNN           & 0.0246 & 0.0004 & 0.0843 & 0.0018 & 9.1052 & 0.1183 & 3947.7 & 0.3828 \\
GemNet          & 0.0101 & 0.0001 & 0.0605 & 0.0020 & 4.6360 & 0.0396 & 1560.6 & 0.1826 \\
EquiFormerV2    & 0.0079 & 0.0001 & 0.0544 & 0.0016 & 6780.8 & 0.0832 & 10840. & 0.8656 \\
DimeNet$^{++}$  & 0.0619 & 0.0001 & 1.3287 & 0.0654 & 27.959 & 0.0011 & 684.33 & 2.4980 \\
NequIP          & 0.0294 & 0.0004 & 0.1735 & 0.0055 & 0.0311 & 0.0010 & 715.34 & 0.1114 \\
eSCN            & 0.0224 & 0.0002 & 0.0711 & 0.0019 & 10.382 & 0.1266 & 3963.0 & 0.4553 \\
MACE            & 0.0049 & 0.0000 & 0.0390 & 0.0013 & 0.0029 & 0.0001 & 2.6409 & 0.0073 \\
MACE-Large      & 0.0084 & 0.0000 & 0.0320 & 0.0009 & 0.0045 & 0.0001 & 0.7266 & 0.0050 \\
MACE-Medium ZS  & 0.1388 & 0.0003 & 0.1521 & 0.0003 & 11969. & 1.4008 & 16023. & 0.3940 \\
MACE-Medium     & 0.0139 & 0.0000 & 0.0356 & 0.0007 & 0.0077 & 0.0002 & 2.9888 & 0.0117 \\
MACE-Small-25   & 0.1454 & 0.0006 & 0.1904 & 0.0007 & 7.4707 & 0.0075 & 15.638 & 0.0430 \\
MACE-Small-110  & 0.0498 & 0.0002 & 0.0891 & 0.0016 & 0.1846 & 0.0026 & 7.1629 & 0.0414 \\
MACE-Small-2000 & 0.0163 & 0.0001 & 0.0404 & 0.0009 & 0.0089 & 0.0000 & 3.5101 & 0.0141 \\
UMA-Small-110   & 0.1842 & 0.0005 & 0.2218 & 0.0003 & 0.6322 & 0.0004 & 821.99 & 0.0093 \\
UMA-Small-2000  & 0.0986 & 0.0005 & 0.1157 & 0.0005 & 0.1544 & 0.0004 & 822.51 & 0.0051 \\
\midrule

\multicolumn{9}{l}{\textit{Fragment Combination}} \\*
SchNet          & 0.4325 & 0.0024 & 79.734 & 0.4871 & 11.087 & 0.0274 & 1098.8 & 4.6876 \\
PaiNN           & 0.0265 & 0.0002 & 0.0605 & 0.0003 & 5.0876 & 0.0789 & 42.216 & 0.2081 \\
GemNet          & 0.0158 & 0.0000 & 0.0304 & 0.0001 & 3.6094 & 0.0443 & 5.2326 & 0.0217 \\
EquiFormerV2    & 0.0066 & 0.0000 & 0.0158 & 0.0000 & 4.5004 & 0.0050 & 46.744 & 0.0165 \\
DimeNet$^{++}$  & 0.0257 & 0.0000 & 0.0331 & 0.0001 & 43.532 & 0.0013 & 43.504 & 0.0011 \\
NequIP          & 0.0434 & 0.0001 & 0.1756 & 0.1571 & 0.0502 & 0.0007 & 0.1128 & 0.0653 \\
eSCN            & 0.0203 & 0.0001 & 0.0340 & 0.0006 & 33.211 & 0.0284 & 62.500 & 0.2066 \\
MACE            & 0.0053 & 0.0000 & 0.0133 & 0.0002 & 0.0029 & 0.0001 & 0.0076 & 0.0002 \\
MACE-Large      & 0.0102 & 0.0000 & 0.0196 & 0.0003 & 0.0053 & 0.0000 & 0.0177 & 0.0000 \\
MACE-Medium ZS  & 0.1630 & 0.0004 & 0.1653 & 0.0002 & 11358. & 0.4242 & 11358. & 0.4235 \\
MACE-Medium     & 0.0170 & 0.0000 & 0.0319 & 0.0001 & 0.0093 & 0.0002 & 0.0250 & 0.0000 \\
MACE-Small-25   & 0.0821 & 0.0000 & 0.0797 & 0.0003 & 1.2959 & 0.0016 & 1.2285 & 0.0005 \\
MACE-Small-110  & 0.0424 & 0.0000 & 0.0489 & 0.0002 & 0.0780 & 0.0010 & 0.0911 & 0.0005 \\
MACE-Small-2000 & 0.0215 & 0.0000 & 0.0360 & 0.0001 & 0.0121 & 0.0002 & 0.0394 & 0.0003 \\
UMA-Small-110   & 0.0587 & 0.0002 & 0.0596 & 0.0001 & 0.2371 & 0.0003 & 0.2338 & 0.0001 \\
UMA-Small-2000  & 0.0215 & 0.0001 & 0.0222 & 0.0000 & 0.0439 & 0.0001 & 0.0358 & 0.0002 \\
\midrule

\multicolumn{9}{l}{\textit{Fragment Fission}$^{\dagger}$} \\*
SchNet          & 0.4452 & 0.0025 & 0.8676 & 0.1682 & 9.2223 & 0.0806 & 1935.6 & 1.6820 \\
PaiNN           & 0.0246 & 0.0004 & 0.1557 & 0.0010 & 9.1052 & 0.1183 & 1944.0 & 0.2337 \\
GemNet          & 0.0101 & 0.0001 & 0.1924 & 0.0007 & 4.6360 & 0.0396 & 767.60 & 0.4542 \\
EquiFormerV2    & 0.0079 & 0.0001 & 0.0817 & 0.0006 & 6780.8 & 0.0832 & 4542.0 & 0.9204 \\
DimeNet$^{++}$  & 0.0619 & 0.0001 & 3.6793 & 0.0194 & 27.959 & 0.0011 & 690.66 & 1.1757 \\
NequIP          & 0.0294 & 0.0004 & 7.0260 & 0.0412 & 0.0311 & 0.0010 & 391.68 & 0.0454 \\
eSCN            & 0.0224 & 0.0002 & 0.1016 & 0.0009 & 10.382 & 0.1266 & 1972.4 & 1.1836 \\
MACE            & 0.0049 & 0.0000 & 0.1886 & 0.0004 & 0.0029 & 0.0001 & 1.4918 & 0.0020 \\
MACE-Large      & 0.0084 & 0.0000 & 0.0748 & 0.0004 & 0.0045 & 0.0001 & 0.2635 & 0.0015 \\
MACE-Medium ZS  & 0.1388 & 0.0003 & 0.1529 & 0.0005 & 11969. & 1.4008 & 9416.3 & 0.3020 \\
MACE-Medium     & 0.0139 & 0.0000 & 0.1293 & 0.0003 & 0.0077 & 0.0002 & 1.2501 & 0.0035 \\
MACE-Small-2000 & 0.0163 & 0.0001 & 0.1000 & 0.0005 & 0.0089 & 0.0000 & 1.5256 & 0.0042 \\
UMA-Small-2000  & 0.0986 & 0.0005 & 0.1182 & 0.0002 & 0.1544 & 0.0004 & 411.82 & 0.0016 \\
\end{longtable}
\endgroup

\begingroup
\footnotesize
\setlength\tabcolsep{4pt}
\begin{longtable}{l r@{\,$\pm$\,}l r@{\,$\pm$\,}l r@{\,$\pm$\,}l r@{\,$\pm$\,}l}
\caption{\footnotesize Comparison of MLIP models on the five GMD tasks: Fragment Chain Extension, Fragment Fusion, Fragment Duplication, Fragment Combination and Fragment Fission. Metrics reported are Forces Cosine Similarity (\emph{F Cosine}) and Forces Magnitude Mean Absolute Error (\emph{F Magnitude}). Results are averaged across all molecules for each task. ID and OOD indicate in-distribution and out-of-distribution splits, respectively. Forces in eV/\AA. $^{\dagger}$Fragment Fission uses the Fragment Duplication models and in-distribution test set, so its ID values are those of Fragment Duplication. MACE-Small-25, MACE-Small-110 and UMA-Small-110 were not evaluated on this task.}
\label{tab:numericals_cosine_fmag} \\
\toprule
 & \multicolumn{4}{c}{\textbf{F Cosine}} & \multicolumn{4}{c}{\textbf{F Magnitude}} \\
\cmidrule(lr){2-5}\cmidrule(lr){6-9}
\textbf{Model} & \multicolumn{2}{c}{ID} & \multicolumn{2}{c}{OOD} & \multicolumn{2}{c}{ID} & \multicolumn{2}{c}{OOD} \\
\midrule
\endfirsthead

\multicolumn{9}{c}{\textbf{Table \ref{tab:numericals_cosine_fmag} (continued)}} \\
\toprule
 & \multicolumn{4}{c}{\textbf{F Cosine}} & \multicolumn{4}{c}{\textbf{F Magnitude}} \\
\cmidrule(lr){2-5}\cmidrule(lr){6-9}
\textbf{Model} & \multicolumn{2}{c}{ID} & \multicolumn{2}{c}{OOD} & \multicolumn{2}{c}{ID} & \multicolumn{2}{c}{OOD} \\
\midrule
\endhead

\midrule
\multicolumn{9}{r}{\textit{Continued on next page}} \\
\endfoot

\bottomrule
\endlastfoot

\multicolumn{9}{l}{\textit{Fragment Chain Extension}} \\*
SchNet          & 0.9967 & 0.0001 & 0.5439 & 0.0055 & 0.0357 & 0.0027 & 2.6929 & 0.0904 \\
PaiNN           & 0.9997 & 0.0000 & 0.9885 & 0.0005 & 0.0087 & 0.0001 & 0.0685 & 0.0035 \\
GemNet          & 1.0000 & 0.0000 & 0.9994 & 0.0001 & 0.0039 & 0.0000 & 0.0136 & 0.0012 \\
EquiFormerV2    & 1.0000 & 0.0000 & 0.9999 & 0.0000 & 0.0015 & 0.0000 & 0.0042 & 0.0001 \\
DimeNet$^{++}$  & 1.0000 & 0.0000 & 0.6029 & 0.0071 & 0.0032 & 0.0000 & 2.1529 & 0.0217 \\
NequIP          & 0.9994 & 0.0000 & 0.9978 & 0.0002 & 0.0161 & 0.0002 & 0.0288 & 0.0003 \\
eSCN            & 1.0000 & 0.0000 & 0.9966 & 0.0005 & 0.0028 & 0.0000 & 0.0326 & 0.0005 \\
MACE            & 1.0000 & 0.0000 & 0.9999 & 0.0000 & 0.0038 & 0.0000 & 0.0065 & 0.0001 \\
MACE-Large      & 0.9999 & 0.0000 & 0.9998 & 0.0000 & 0.0057 & 0.0000 & 0.0095 & 0.0001 \\
MACE-Medium ZS  & 0.9641 & 0.0006 & 0.9699 & 0.0004 & 0.1626 & 0.0003 & 0.1475 & 0.0003 \\
MACE-Medium     & 0.9998 & 0.0000 & 0.9995 & 0.0000 & 0.0090 & 0.0001 & 0.0148 & 0.0002 \\
MACE-Small-25   & 0.9930 & 0.0005 & 0.9855 & 0.0010 & 0.0813 & 0.0010 & 0.1011 & 0.0024 \\
MACE-Small-110  & 0.9984 & 0.0005 & 0.9964 & 0.0008 & 0.0310 & 0.0009 & 0.0457 & 0.0023 \\
MACE-Small-2000 & 0.9998 & 0.0000 & 0.9994 & 0.0000 & 0.0101 & 0.0001 & 0.0159 & 0.0006 \\
UMA-Small-110   & 0.9993 & 0.0001 & 0.9990 & 0.0001 & 0.1437 & 0.0020 & 0.1476 & 0.0686 \\
UMA-Small-2000  & 0.9997 & 0.0000 & 0.9996 & 0.0000 & 0.0347 & 0.0005 & 0.0282 & 0.0033 \\
\midrule

\multicolumn{9}{l}{\textit{Fragment Fusion}} \\*
SchNet          & 0.2495 & 0.0017 & 0.0645 & 0.0021 & 4.7377 & 0.0213 & 16.715 & 0.1484 \\
PaiNN           & 0.9993 & 0.0000 & 0.9380 & 0.0014 & 0.0161 & 0.0001 & 0.1096 & 0.0012 \\
GemNet          & 0.9981 & 0.0000 & 0.9506 & 0.0008 & 0.0258 & 0.0000 & 0.1281 & 0.0012 \\
EquiFormerV2    & 0.9998 & 0.0000 & 0.9463 & 0.0009 & 0.0097 & 0.0000 & 0.1573 & 0.0015 \\
DimeNet$^{++}$  & 0.9824 & 0.0001 & 0.8816 & 0.0031 & 0.0984 & 0.0003 & 0.3925 & 0.0034 \\
NequIP          & 0.9962 & 0.0000 & 0.9030 & 0.0008 & 0.0412 & 0.0002 & 0.2599 & 0.0032 \\
eSCN            & 0.9925 & 0.0001 & 0.9470 & 0.0012 & 0.0638 & 0.0002 & 0.1519 & 0.0007 \\
MACE            & 0.9999 & 0.0000 & 0.9680 & 0.0009 & 0.0050 & 0.0000 & 0.0965 & 0.0010 \\
MACE-Large      & 0.9997 & 0.0000 & 0.9849 & 0.0005 & 0.0106 & 0.0000 & 0.0754 & 0.0007 \\
MACE-Medium ZS  & 0.9509 & 0.0002 & 0.9639 & 0.0004 & 0.1911 & 0.0004 & 0.1656 & 0.0002 \\
MACE-Medium     & 0.9993 & 0.0000 & 0.9526 & 0.0004 & 0.0163 & 0.0000 & 0.1654 & 0.0016 \\
MACE-Small-25   & 0.9936 & 0.0001 & 0.9571 & 0.0009 & 0.0559 & 0.0002 & 0.1375 & 0.0023 \\
MACE-Small-110  & 0.9969 & 0.0000 & 0.9666 & 0.0008 & 0.0389 & 0.0001 & 0.1186 & 0.0014 \\
MACE-Small-2000 & 0.9990 & 0.0000 & 0.9608 & 0.0008 & 0.0203 & 0.0000 & 0.1423 & 0.0014 \\
UMA-Small-110   & 0.9963 & 0.0001 & 0.9948 & 0.0002 & 0.1026 & 0.0004 & 0.0888 & 0.0007 \\
UMA-Small-2000  & 0.9996 & 0.0000 & 0.9815 & 0.0007 & 0.0210 & 0.0000 & 0.0799 & 0.0003 \\
\midrule

\multicolumn{9}{l}{\textit{Fragment Duplication}} \\*
SchNet          & 0.7312 & 0.0019 & 0.3475 & 0.0018 & 0.4739 & 0.0039 & 1.4212 & 0.0123 \\
PaiNN           & 0.9985 & 0.0001 & 0.9729 & 0.0019 & 0.0252 & 0.0005 & 0.0978 & 0.0019 \\
GemNet          & 0.9997 & 0.0000 & 0.9772 & 0.0018 & 0.0108 & 0.0001 & 0.0717 & 0.0018 \\
EquiFormerV2    & 0.9999 & 0.0000 & 0.9765 & 0.0016 & 0.0092 & 0.0002 & 0.0623 & 0.0010 \\
DimeNet$^{++}$  & 0.9816 & 0.0001 & 0.8999 & 0.0029 & 0.0715 & 0.0002 & 2.4530 & 0.1186 \\
NequIP          & 0.9978 & 0.0002 & 0.9579 & 0.0021 & 0.0309 & 0.0003 & 0.2466 & 0.0080 \\
eSCN            & 0.9989 & 0.0001 & 0.9721 & 0.0020 & 0.0252 & 0.0002 & 0.0770 & 0.0011 \\
MACE            & 0.9999 & 0.0000 & 0.9827 & 0.0011 & 0.0053 & 0.0000 & 0.0439 & 0.0013 \\
MACE-Large      & 0.9998 & 0.0000 & 0.9902 & 0.0005 & 0.0092 & 0.0001 & 0.0359 & 0.0011 \\
MACE-Medium ZS  & 0.9642 & 0.0004 & 0.9587 & 0.0005 & 0.1649 & 0.0005 & 0.1805 & 0.0005 \\
MACE-Medium     & 0.9995 & 0.0000 & 0.9935 & 0.0006 & 0.0149 & 0.0002 & 0.0376 & 0.0006 \\
MACE-Small-25   & 0.9653 & 0.0003 & 0.9413 & 0.0007 & 0.1878 & 0.0003 & 0.2436 & 0.0004 \\
MACE-Small-110  & 0.9944 & 0.0001 & 0.9761 & 0.0010 & 0.0556 & 0.0004 & 0.0966 & 0.0019 \\
MACE-Small-2000 & 0.9993 & 0.0001 & 0.9916 & 0.0005 & 0.0175 & 0.0001 & 0.0435 & 0.0009 \\
UMA-Small-110   & 0.9844 & 0.0002 & 0.9680 & 0.0004 & 0.3439 & 0.0011 & 0.4054 & 0.0002 \\
UMA-Small-2000  & 0.9955 & 0.0001 & 0.9846 & 0.0006 & 0.1799 & 0.0009 & 0.2009 & 0.0010 \\
\midrule

\multicolumn{9}{l}{\textit{Fragment Combination}} \\*
SchNet          & 0.7670 & 0.0017 & 0.0037 & 0.0060 & 0.4697 & 0.0040 & 157.89 & 1.0425 \\
PaiNN           & 0.9984 & 0.0000 & 0.9909 & 0.0001 & 0.0275 & 0.0003 & 0.0656 & 0.0003 \\
GemNet          & 0.9994 & 0.0000 & 0.9973 & 0.0001 & 0.0169 & 0.0000 & 0.0330 & 0.0001 \\
EquiFormerV2    & 0.9999 & 0.0000 & 0.9991 & 0.0000 & 0.0089 & 0.0000 & 0.0188 & 0.0000 \\
DimeNet$^{++}$  & 0.9985 & 0.0001 & 0.9973 & 0.0001 & 0.0303 & 0.0000 & 0.0377 & 0.0001 \\
NequIP          & 0.9956 & 0.0000 & 0.9883 & 0.0012 & 0.0465 & 0.0001 & 0.2685 & 0.2800 \\
eSCN            & 0.9991 & 0.0000 & 0.9973 & 0.0002 & 0.0227 & 0.0001 & 0.0393 & 0.0007 \\
MACE            & 0.9999 & 0.0000 & 0.9993 & 0.0000 & 0.0058 & 0.0001 & 0.0146 & 0.0002 \\
MACE-Large      & 0.9998 & 0.0000 & 0.9986 & 0.0001 & 0.0111 & 0.0001 & 0.0218 & 0.0001 \\
MACE-Medium ZS  & 0.9515 & 0.0002 & 0.9491 & 0.0005 & 0.1875 & 0.0010 & 0.1904 & 0.0011 \\
MACE-Medium     & 0.9994 & 0.0000 & 0.9968 & 0.0001 & 0.0182 & 0.0001 & 0.0344 & 0.0002 \\
MACE-Small-25   & 0.9869 & 0.0002 & 0.9867 & 0.0008 & 0.0949 & 0.0000 & 0.0881 & 0.0002 \\
MACE-Small-110  & 0.9960 & 0.0001 & 0.9941 & 0.0005 & 0.0460 & 0.0003 & 0.0531 & 0.0001 \\
MACE-Small-2000 & 0.9989 & 0.0000 & 0.9964 & 0.0001 & 0.0228 & 0.0001 & 0.0379 & 0.0001 \\
UMA-Small-110   & 0.9976 & 0.0001 & 0.9977 & 0.0000 & 0.0998 & 0.0004 & 0.1013 & 0.0000 \\
UMA-Small-2000  & 0.9992 & 0.0000 & 0.9992 & 0.0000 & 0.0297 & 0.0001 & 0.0309 & 0.0001 \\
\midrule

\multicolumn{9}{l}{\textit{Fragment Fission}$^{\dagger}$} \\*
SchNet          & 0.7312 & 0.0019 & 0.5792 & 0.0039 & 0.4739 & 0.0039 & 1.1370 & 0.3234 \\
PaiNN           & 0.9985 & 0.0001 & 0.9145 & 0.0010 & 0.0252 & 0.0005 & 0.1542 & 0.0017 \\
GemNet          & 0.9997 & 0.0000 & 0.8911 & 0.0007 & 0.0108 & 0.0001 & 0.1975 & 0.0011 \\
EquiFormerV2    & 0.9999 & 0.0000 & 0.9671 & 0.0006 & 0.0092 & 0.0002 & 0.0940 & 0.0007 \\
DimeNet$^{++}$  & 0.9816 & 0.0001 & 0.8388 & 0.0033 & 0.0715 & 0.0002 & 7.0080 & 0.0360 \\
NequIP          & 0.9978 & 0.0002 & 0.6124 & 0.0011 & 0.0309 & 0.0003 & 13.410 & 0.0729 \\
eSCN            & 0.9989 & 0.0001 & 0.9618 & 0.0010 & 0.0252 & 0.0002 & 0.1150 & 0.0008 \\
MACE            & 0.9999 & 0.0000 & 0.9021 & 0.0005 & 0.0053 & 0.0000 & 0.2002 & 0.0007 \\
MACE-Large      & 0.9998 & 0.0000 & 0.9761 & 0.0003 & 0.0092 & 0.0001 & 0.0850 & 0.0005 \\
MACE-Medium ZS  & 0.9642 & 0.0004 & 0.9547 & 0.0005 & 0.1649 & 0.0005 & 0.1793 & 0.0005 \\
MACE-Medium     & 0.9995 & 0.0000 & 0.9390 & 0.0003 & 0.0149 & 0.0002 & 0.1381 & 0.0007 \\
MACE-Small-2000 & 0.9993 & 0.0001 & 0.9618 & 0.0004 & 0.0175 & 0.0001 & 0.1106 & 0.0008 \\
UMA-Small-2000  & 0.9955 & 0.0001 & 0.9835 & 0.0003 & 0.1799 & 0.0009 & 0.2004 & 0.0012 \\
\end{longtable}
\endgroup

\subsection{Failure composition across architectural families}
\label{app:failure-composition}
The aggregate metrics reported in Appendix~\ref{details:numerical_results}, and the ID-to-OOD profiles in Figure~\ref{fig:aug_overview}, invite a natural but mistaken reading: that architectures which achieve similar Forces MAE on a given task are failing in similar ways. Mean Absolute Error is a scalar summary of a vector-valued prediction, and it is insensitive to how the error is distributed over the two independent degrees of freedom a force prediction has, its magnitude and its direction. Two models with similar Forces MAE may therefore exhibit quite different pathologies: one may systematically underpredict force magnitudes while pointing in almost exactly the right direction, and another may predict magnitudes well but rotate the force vector on a small subset of atoms.

Magnitude and direction are not the only axes along which this summary collapses. A model's error is also distributed unevenly over the atoms of a molecule, and that distribution is itself model-dependent: two models evaluated on the same molecule may concentrate their error on different atoms, or differ in how sharply it is concentrated at all, and still arrive at comparable aggregate error. Because both models are evaluated on identical geometries and identical snapshots, any such divergence cannot be attributed to the difficulty of the configuration. Sample difficulty is a property of the sample, not of the model, so a spatial pattern of error that changes with the architecture localises the failure in the model rather than in the benchmark.

The decomposition is reported for a subset of the evaluated models, chosen to span the distinct inductive biases in our model set rather than to enumerate every checkpoint. Architectures that share a design principle behave similarly under decomposition as well as in aggregate, so including several members of the same family would add panels without adding information. We therefore report one representative per family.

\begin{figure*}[t]
  \centering
  \includegraphics[width=1.0\linewidth]{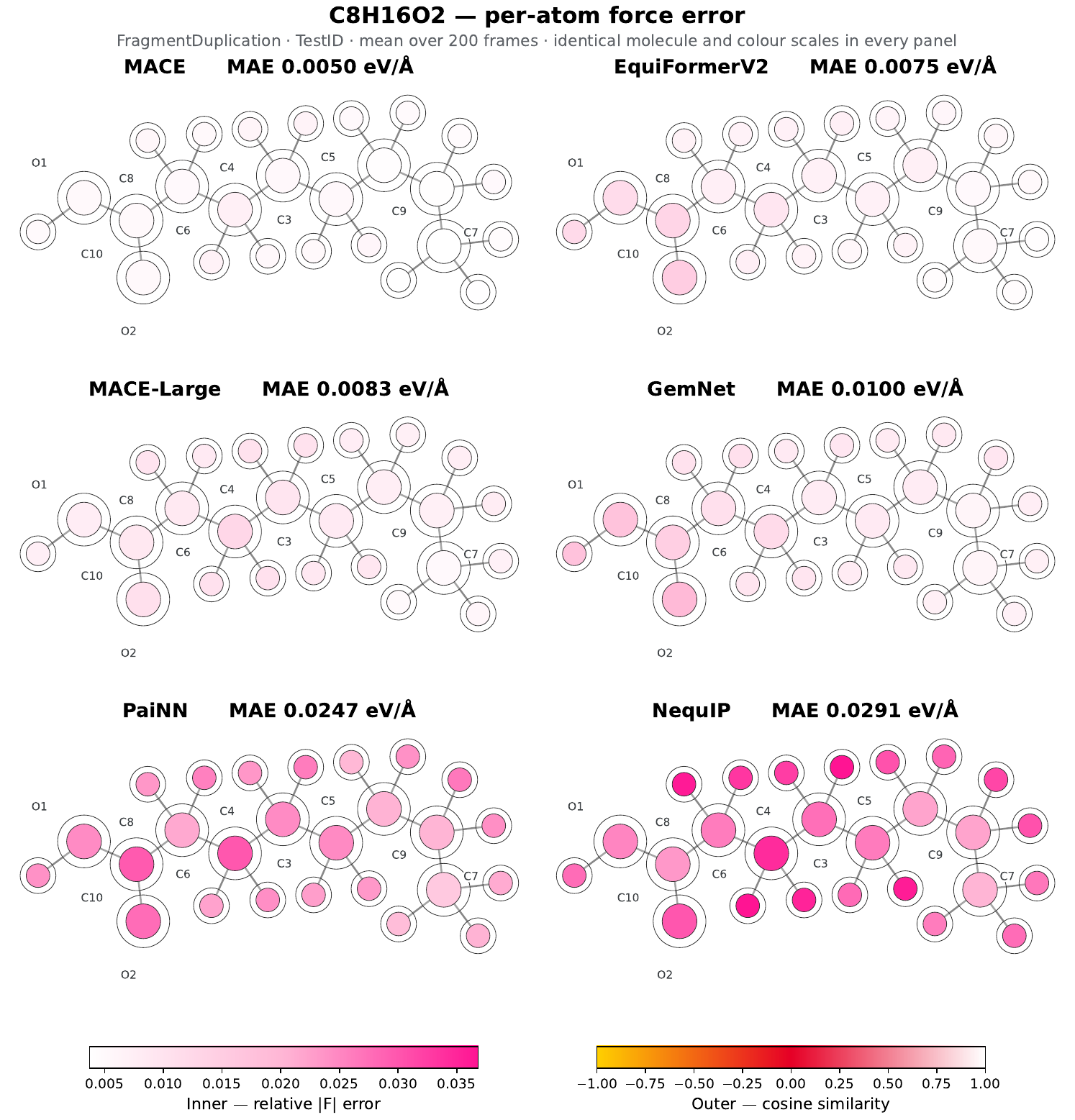}
  \caption{Per-atom force error of six models on octanoic acid
  (C\textsubscript{8}H\textsubscript{16}O\textsubscript{2}), an in-distribution molecule of
  the Fragment Duplication task, averaged over 200 test snapshots. Inner disc: relative
  force-magnitude error; outer ring: force cosine similarity. Geometry and colour scales are
  identical across panels, but differ between figures. All outer rings are white, so force
  direction is predicted correctly and the error lies in magnitude. MACE, EquiFormerV2,
  MACE-Large and GemNet (0.0050--0.0100\,eV/\AA{}) show little error anywhere, while PaiNN
  and NequIP (0.0247 and 0.0291\,eV/\AA{}) show raised error along the whole carbon chain
  and its hydrogens. Atom labels are dataset indices.}
  \label{fig:FD_PAE_ID_ALL}
\end{figure*}

\begin{figure*}[t]
  \centering
  \includegraphics[width=1.0\linewidth]{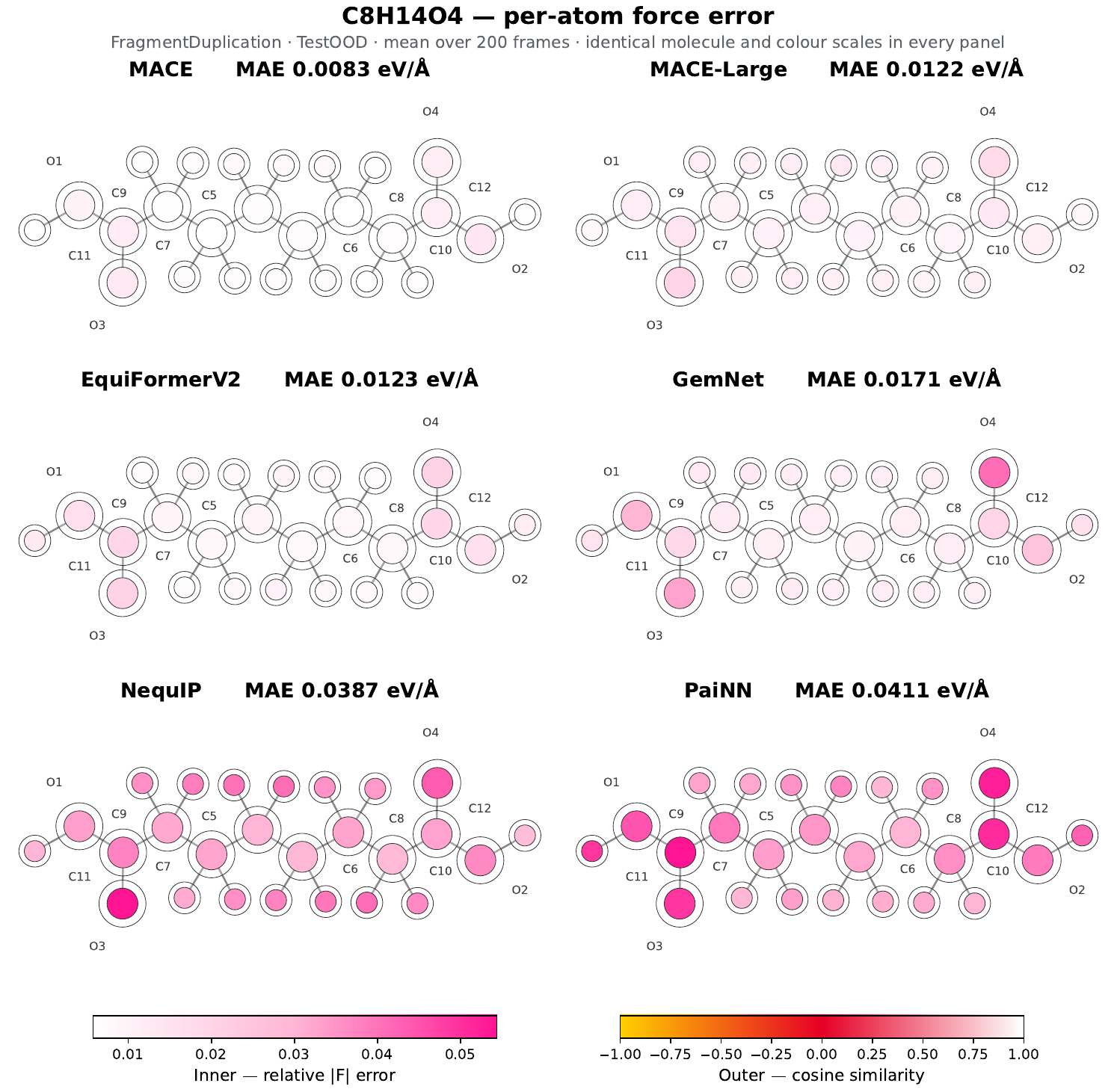}
  \caption{Per-atom force error on octanedioic acid
  (C\textsubscript{8}H\textsubscript{14}O\textsubscript{4}), the out-of-distribution
  molecule of the Fragment Duplication task, averaged over 200 test snapshots. Encoding as
  in Figure~\ref{fig:FD_PAE_ID_ALL}. All four oxygens carry raised error, in both carboxyl
  groups. Away from the oxygens the models differ: MACE, MACE-Large and EquiFormerV2 keep
  the carbon chain nearly error-free, whereas NequIP and PaiNN (0.0387 and
  0.0411\,eV/\AA{}) spread error over most of the molecule. Outer rings remain white, so
  the out-of-distribution error is in magnitude, not direction.}
  \label{fig:FD_PAE_OOD_ALL}
\end{figure*}

\begin{figure*}[t]
  \centering
  \includegraphics[width=1.0\linewidth]{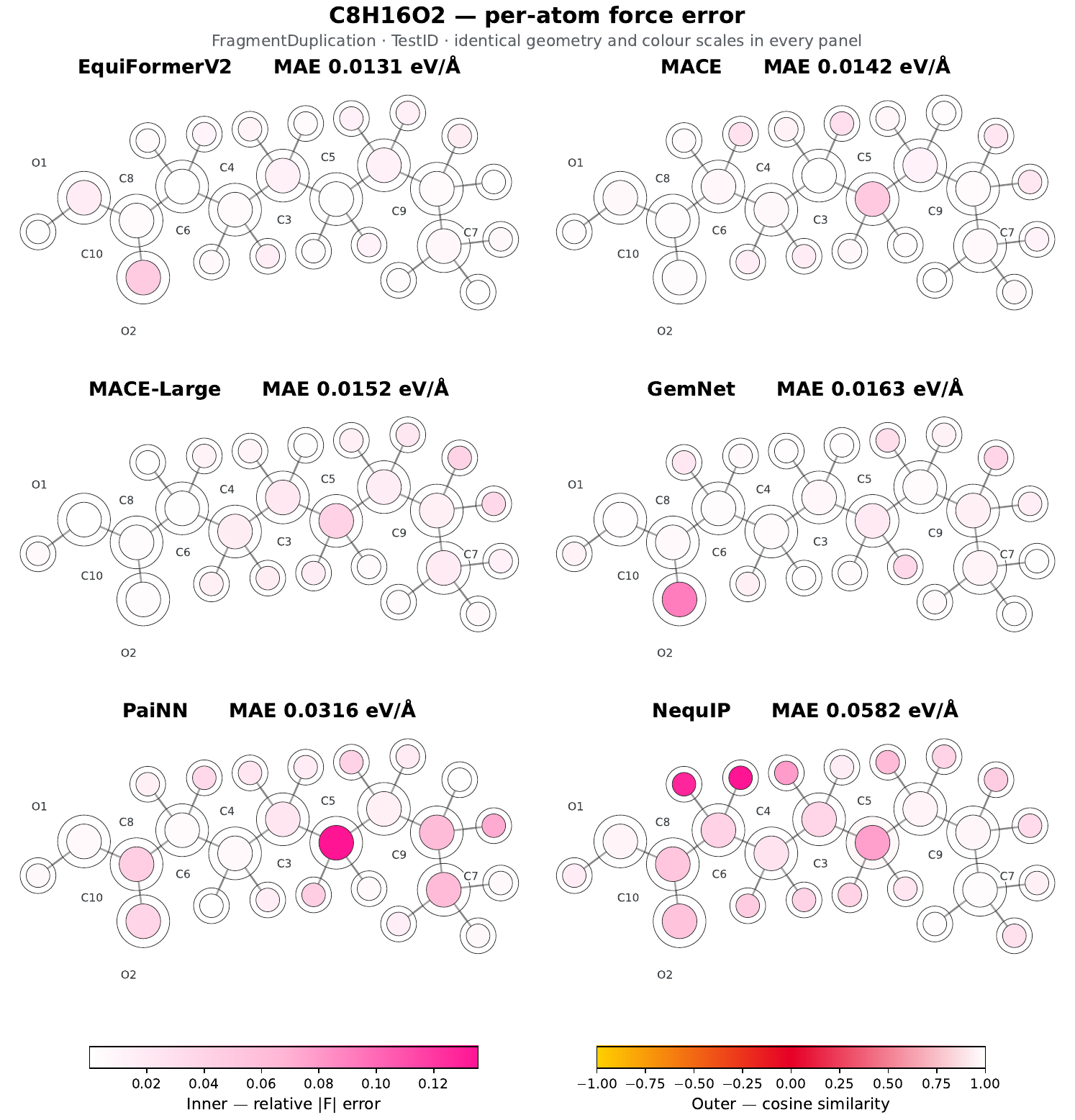}
  \caption{As Figure~\ref{fig:FD_PAE_ID_ALL}, for a single test snapshot (frame 165).
  Without averaging, each model concentrates its error on different atoms: PaiNN on C5,
  NequIP on C5 and two hydrogens next to C8, and GemNet on O2, while MACE and MACE-Large
  spread it more evenly. Since the geometry is the same in all panels, these differences
  come from the models, not from the configuration. Outer rings remain white.}
  \label{fig:FD_PAE_ID_Single}
\end{figure*}

\begin{figure*}[t]
  \centering
  \includegraphics[width=1.0\linewidth]{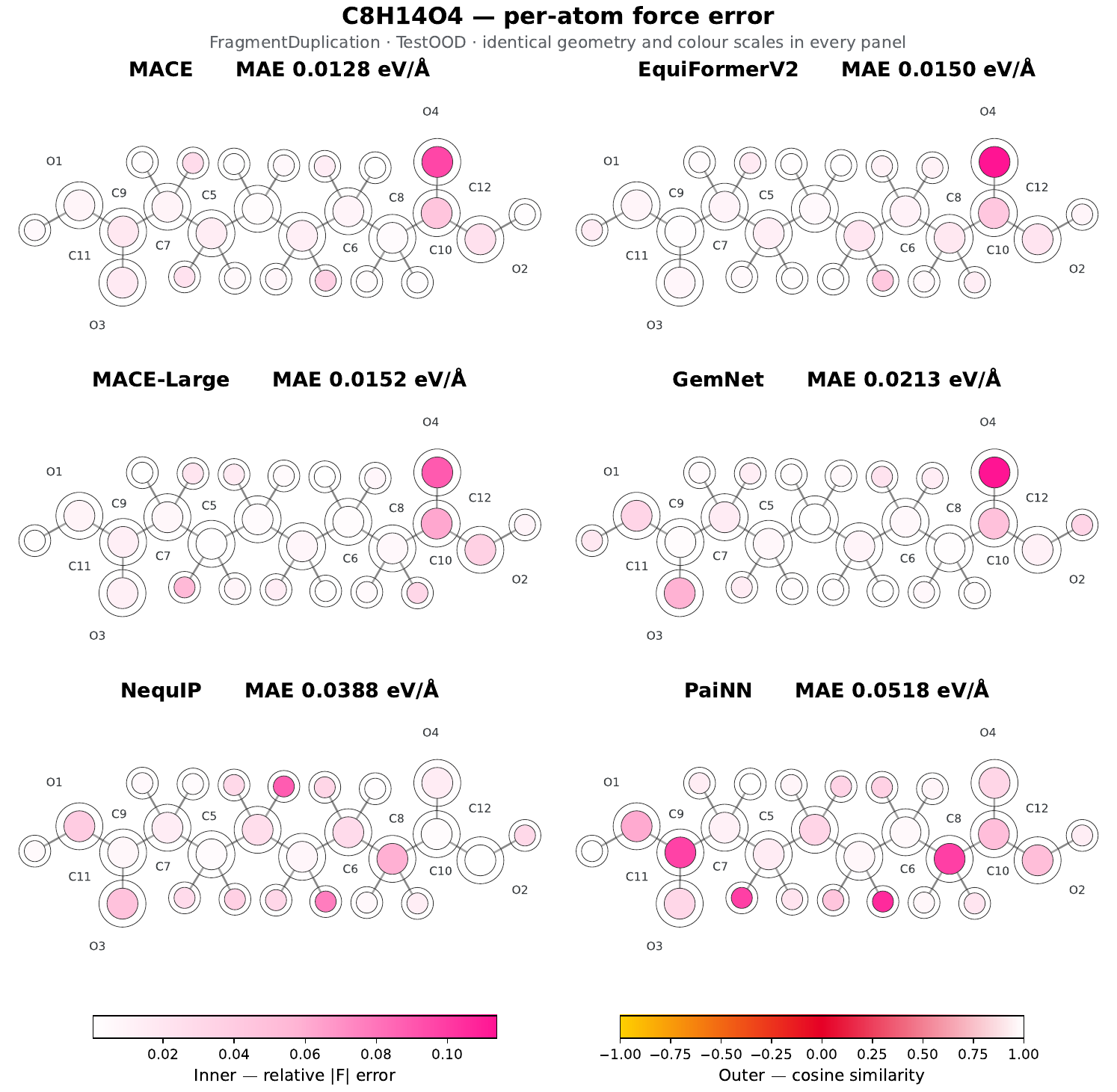}
  \caption{As Figure~\ref{fig:FD_PAE_OOD_ALL}, for a single test snapshot (frame 85).
  MACE, EquiFormerV2, MACE-Large and GemNet (0.0128--0.0213\,eV/\AA{}) have their largest
  error on O4, with C10 next to it also raised, and little error elsewhere. NequIP and
  PaiNN instead spread their error across C6, C9 and many hydrogens. Models with similar
  MAE can therefore have either localised or spread-out error, which MAE alone does not
  show.}
  \label{fig:FD_PAE_OOD_Single}
\end{figure*}

To make this per-atom structure visible, Figures~\ref{fig:FD_PAE_ID_ALL}--\ref{fig:FD_PAE_OOD_Single} map the force error onto the molecular geometry itself for the Fragment Duplication task. Each atom is rendered as a node carrying two quantities: the force cosine similarity, encoded as colour in the outer shell, and the relative force-magnitude error, encoded as colour in the inner shell. The same molecule and the same snapshots are used for every panel, and colour scales are shared across panels, so differences between panels are differences between models and nothing else. We show both trajectory-averaged maps (Figures~\ref{fig:FD_PAE_ID_ALL} and~\ref{fig:FD_PAE_OOD_ALL}) and single-snapshot maps (Figures~\ref{fig:FD_PAE_ID_Single} and~\ref{fig:FD_PAE_OOD_Single}), since averaging over a trajectory smooths precisely the spatial structure that distinguishes the models.

This construction supports a comparison that neither the aggregate metrics nor the element-resolved decompositions (Figures~\ref{fig:FC:decomp_all}, \ref{fig:FD:decomp_all}, \ref{fig:FCB:decomp_all} and~\ref{fig:FCE:decomp_all}) permit. Those figures answer the question of which element types carry error, but it pools all atoms of a given element, so an oxygen that is well predicted and an oxygen that is not appear in the same distribution. Mapping error onto the geometry separates chemically equivalent atoms by position, which matters here because our tasks introduce novel environments at specific sites rather than uniformly across a molecule.

Read this way, the out-of-distribution maps show a shared component and a model-specific one. The shared component concerns the oxygens, and it is not confined to the group the task introduces. Every model carries elevated error across all four oxygen sites, including O1 and O3, which belong to a carboxyl group of exactly the kind seen throughout training and whose local environment is essentially unchanged from the in-distribution molecule. Duplication therefore does not merely add an unlearned site; it degrades the prediction at learned ones. Since the immediate neighbourhood of O1 and O3 lies well within the receptive field of every model considered, this indicates that the representation of a familiar chemical environment is perturbed by the presence of a second, spatially distant instance of the same motif, rather than remaining invariant to it. This is a failure of locality rather than of coverage, and it is consistent with the Fragment Fission results, where composite environments transfer as units but cannot be decomposed into their constituents.

The model-specific component is what happens away from the oxygens, and there the architectures diverge. MACE, MACE-Large and EquiFormerV2 leave the carbon backbone and its hydrogens close to clean, so their error remains concentrated on the functionalised sites. NequIP and PaiNN instead spread error of comparable total magnitude across essentially the whole molecule, including backbone positions that the other models predict well, so that in their maps the oxygens are not distinguished from their surroundings. This is not because they predict the oxygens better in absolute terms, but because their background error is high enough to mask the distinction.

These are structurally different failures rather than different amounts of the same failure. A sparse failure indicates a model that has represented the molecule adequately and mispredicts one unfamiliar environment; a diffuse failure indicates a representation degraded globally, in which the compositional novelty is not the operative difficulty. The distinction is invisible to aggregate MAE, and the four most accurate models on this task sit close enough together by that measure to appear interchangeable. It also bears on the interpretation of the uniformity in Figure~\ref{fig:aug_overview}: architectures converge in scalar error while differing in how that error is organised over the molecule, so the apparent agreement is an artefact of aggregation rather than evidence of a common failure mechanism. For benchmark use this has a practical consequence, in that a model selected on aggregate OOD MAE alone may be selected for having a tolerable amount of the wrong kind of error, distributed over atoms that matter for the dynamics of interest.

A further observation cuts across all panels: the outer shells remain white throughout, in-distribution and out-of-distribution alike, indicating that force direction is recovered essentially exactly even where magnitudes are substantially wrong. This is consistent with the decomposition in Section~\ref{main:evaluation:analysis} and localises the compositional failure in the scalar magnitude channel rather than in the geometric representation of the local environment.

\clearpage

\section{Examining the effects of varying levels of theory on the learned chemical space}
\label{app:level_of_theory}

Initially, we generated data for the proposed generalisation tasks using a semi-empirical method, GFN2-xTB~\cite{bannwarth2019gfn2}, due to its low computational cost and stability in molecular dynamics simulations. During this exploratory phase, several from-scratch models (EquiFormerV2, GemNet-OC, PaiNN, SchNet, and DimeNet$^{++}$) were fully trained using these semi-empirical labels.

\begin{figure*}[h]
    \centering
    \includegraphics[width=1.0\linewidth]{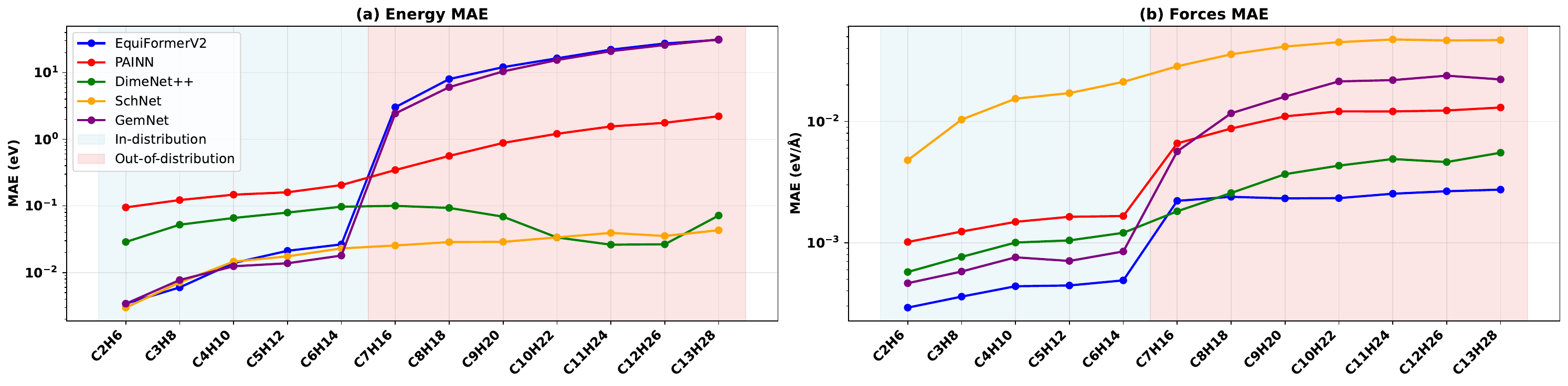}
    \caption{Performance of five from-scratch MLIP architectures on the Fragment Chain Extension task, trained and evaluated using semi-empirical GFN2-xTB labels. Panel (a) displays the Energy MAE (eV) and panel (b) displays the Forces MAE (eV/\AA), both on a logarithmic scale. The background shading denotes the experimental split: models were trained exclusively on short-chain alkanes (C$_2$H$_6$ to C$_6$H$_{14}$, in-distribution, light blue region) and evaluated on longer, previously unseen chains (C$_7$H$_{16}$ to C$_{13}$H$_{28}$, out-of-distribution, light red region). While the models learned this semi-empirical potential energy surface faster and with high in-distribution precision, a distinct out-of-distribution generalisation gap remains visible.}
    \label{fig:xTB_results}
\end{figure*}

We observed that the GFN2-xTB potential energy surface appears to be substantially easier for these models to navigate. The models consistently learned these representations much faster than their PBE counterparts, requiring fewer epochs to converge. Additionally, for some architectures, the final error was lower when trained on the semi-empirical labels. To illustrate this difference in learning dynamics, Figure~\ref{fig:xTB_results} presents the results specifically for the Fragment Chain Extension task.

Furthermore, we conducted full hyperparameter sweeps for these models using the GFN2-xTB labels. The optimal configurations identified during this phase differed notably from those ultimately required for the DFT labels. Because density functional theory provides more rigorous and accurate ground-truth chemical labels, we transitioned the final GMD benchmark to the ORCA(PBE/def2-TZVP+D3BJ) level of theory. This shift in the level of theory fundamentally altered the complexity of the optimisation landscape. Consequently, the hyperparameter sweeps had to be repeated entirely for the PBE labels (as detailed in Appendix~\ref{details:hyperparameters}), and the baseline task difficulty increased markedly compared to the initial GFN2-xTB findings.

\clearpage

\section{Toolkit and dataset generation}
\label{app:toolkit}

Alongside the GMD dataset, we release an open-source toolkit that supports two use cases: regenerating GMD from scratch, and extending it to new molecular families under the same five-task structure. 
The toolkit is available at \url{https://github.com/Nourollah/GMD-26}, both as a standalone repository and as a component of the GOAL project (\url{https://github.com/Nourollah/GOAL}). 
The released dataset is hosted on the Hugging Face Hub at \url{https://huggingface.co/datasets/AmirMasoud/GMD}.
The experiments reported in this paper were produced with separate training scripts; the toolkit is provided to support reproduction of the dataset and extension of the benchmark by the community. Specifically, the GMD toolkit covers data generation and task-split construction; model training and evaluation in this paper were carried out using the original repositories of each architecture family, \texttt{fairchem-core} for SchNet, PAINN, DimeNet++, GemNet, EquiFormerV2, eSCN, and UMA-Small, \texttt{mace-torch} for MACE and MACE-MP-0, and the official NequIP implementation, without modification beyond standard training-loop configuration.

\subsection{Dataset generation pipeline}
\label{app:toolkit_pipeline}

For each molecule in the dataset, the toolkit executes a four-stage generation pipeline implemented through ASE~\citep{larsen2017atomic}: initial 3D structure generation from a SMILES string, off-equilibrium trajectory sampling with FlashMD, snapshot-level relabelling with ORCA(PBE/def2-TZVP+D3BJ), and standardised packaging into ASE-readable extended XYZ format. Each stage is described below in sufficient detail to reproduce the dataset.

\paragraph{Initial structure generation.} The pipeline begins with a molecular specification (a SMILES string), from which RDKit\footnote{\url{https://www.rdkit.org/}} generates an initial 3D conformer. Conformers are subsequently used as the starting geometry for the trajectory-sampling stage.

\paragraph{Trajectory sampling with FlashMD.} We use FlashMD~\citep{bigi2025flashmd} with the \texttt{pet-omatpes} checkpoint, a PET-MAD~\citep{mazitov2025pet} backbone trained against the OMat-PES dataset, to propagate long-stride molecular dynamics. Each molecule is simulated in vacuum for $2 \times 10^5$ steps with a 16~fs integrator timestep using a Langevin thermostat at 300~K and a velocity-relaxation time constant of 100~fs. Initial velocities are drawn from the Maxwell--Boltzmann distribution at 300~K. Snapshots are recorded every 100 steps, yielding approximately 2{,}000 configurations per trajectory. One trajectory is generated per molecule.

\paragraph{Sampling rationale.} While perturbing coordinates with noise is a common alternative for generating off-equilibrium configurations~\citep{feng2023may}, generating dynamic trajectories provides a more physically consistent and transferable basis. We additionally note that direct MD with cheaper methods such as GFN2-xTB produced substantially less diverse configurational coverage even at 100$\times$ longer simulation time, motivating the use of FlashMD. Since our evaluation concerns relative ID versus OOD performance under a fixed sampling protocol, conclusions are invariant to the specific sampler chosen, provided the same sampler is applied uniformly across both distributions.

\paragraph{DFT relabelling with ORCA.} Each stored snapshot is relabelled with ORCA(PBE/def2-TZVP+D3BJ). The full ORCA input keyword line is \texttt{!EnGrad PBE D3BJ def2-TZVP RI TightSCF}. 
The RI approximation (resolution-of-the-identity for Coulomb integrals) is used for computational efficiency, and TightSCF convergence criteria are applied throughout. 
For snapshots where the default SCF procedure failed to converge, we invoked outer-SCF fallback strategies, which yielded successful convergence for all snapshots in the released dataset; no configurations were excluded.
Each single-point calculation produces a total energy and per-atom forces. DFT relabelling was performed on the GRACE CPU.

\paragraph{Packaging.} The toolkit outputs the final dataset containing refined coordinates, per-atom forces, and total energies in both ASE-readable extended XYZ format (\texttt{.extxyz}) and ASE trajectory format (\texttt{.traj}), alongside atomic numbers, positions, and the originating molecule and snapshot indices. Curated train/validation/OOD-test splits are released alongside the raw trajectories, so that reproducing the evaluation protocol does not require re-running upstream generation stages.

\paragraph{Dataset size.} The released GMD dataset comprises 118 molecules, spanning the ten fragment groups detailed in Figure~\ref{fig:dataset_molecules}, and more than 296{,}000 labelled configurations in total. Per-task molecule counts and fragment groups are reported in Table~\ref{tab:splitting_strategy_multirow}; the exact number of snapshots per trajectory is nominally 2{,}000 across all molecules.

The released dataset additionally includes linear alkanes with carbon chain lengths beyond the C$_2$--C$_{13}$ range evaluated in the main paper, which are not used in the experiments reported here. Together with the released fine-tuned model weights, these permit evaluation of chain-length extrapolation over a wider range without retraining, and are provided to support future work on the limits of length generalisation.

\paragraph{Extending the benchmark.} Extending GMD to new chemical families reduces to providing a set of SMILES strings per fragment group and rerunning the pipeline. The five task templates (chain extension, fusion, duplication, combination, fission) are defined over abstract fragment sets and are reusable without modification. The pipeline itself does not assume small molecules; extending the benchmark to larger systems is compute-bound on the DFT relabelling stage rather than architecturally constrained. Substituting the sampler (e.g., replacing FlashMD with long-timescale AIMD) or the relabelling backend (e.g., replacing ORCA(PBE/def2-TZVP+D3BJ) with CP2K or a hybrid functional) requires changing only the corresponding pipeline stage: the benchmark is defined by the task structure described in the main paper, not by the specific sampler or functional used here.

\subsection{Validation of the reference labels}
\label{app:level_of_theory}

Two properties of our reference data bear on the interpretation of the results in the main paper: whether the labels are numerically well converged, and whether the choice of a Gaussian basis set rather than a plane-wave representation materially affects the forces against which models are evaluated. We address each in turn.

\paragraph{Net-force diagnostic.}
For an isolated molecule in vacuum, the vector sum of all atomic forces must vanish by translational invariance. Any departure from zero is therefore numerical rather than physical, and reflects the accumulated effect of finite integration grids, integral screening thresholds, and basis-set incompleteness. \citet{kuryla2025accurate} show that force errors of several to tens of meV/\AA{} can arise within a single DFT method from these settings alone, and that such noise is present in several widely used MLIP training datasets; they identify a residual net force below $1$\,meV/\AA{} per atom as indicative of well-converged forces.

We computed the net force for each of the 164{,}600 configurations that constitute the training, validation, and test splits used in this work. Across all four tasks, the mean net force per atom lies between $5.6 \times 10^{-8}$ and $7.4 \times 10^{-8}$\,meV/\AA{}, roughly seven to eight orders of magnitude below the threshold above, and every configuration in every task falls below it. Our labels are therefore not affected by the numerical noise documented in that work, and the generalisation gaps we report cannot be attributed to inconsistency in the reference forces.

\paragraph{Plane-wave versus Gaussian-basis PBE.}
Our labels were produced with ORCA using a Gaussian basis set, whereas several foundation models, including MACE-MP-0, were pre-trained on plane-wave PBE data. Although the nominal functional is the same, the two representations need not yield identical forces, which raises the question of whether the zero-shot errors we report reflect genuine model error or merely a difference in computational setting.

To quantify this, we recomputed forces and energies with plane-wave PBE on frozen geometries drawn from our production trajectories, holding the molecule and the level of theory fixed and varying only the basis representation. We first performed a plane-wave cutoff convergence sweep on a production geometry to confirm that the chosen cutoff yields converged forces. The comparison itself was then carried out on 20 frozen geometries, with the results in Table~\ref{tab:pw_vs_gaussian}.


\begin{table}[htbp]
\centering
\caption{Force agreement between plane-wave PBE and Gaussian-basis ORCA(PBE/def2-TZVP+D3BJ) on 20 frozen geometries, and its magnitude relative to the zero-shot model error on the same task.}
\label{tab:pw_vs_gaussian}
\begin{tabular}{lrr}
\toprule
Quantity & Value & Ratio \\
\midrule
Force component MAE (level of theory) & 0.0269\,eV/\AA{} & $1\times$ \\
Force component RMSE (level of theory) & 0.0380\,eV/\AA{} & --- \\
Force direction cosine similarity & 0.9993 & --- \\
\midrule
MACE-Medium ZS, ID force MAE & 0.1388\,eV/\AA{} & ${\sim}5.1\times$ \\
MACE-Medium ZS, OOD force MAE & 0.1521\,eV/\AA{} & ${\sim}5.6\times$ \\
\bottomrule
\end{tabular}
\end{table}

Force direction is essentially unaffected by the choice of representation, with a cosine similarity of 0.9993 between the two references. The magnitude discrepancy is approximately five times smaller than the force MAE of the zero-shot MACE-MP-0 Medium model on the same task, so level-of-theory differences account for a minority of the zero-shot error we report, and the remainder reflects model-level generalisation failure rather than a computational artefact.

This concern applies only to the zero-shot results. As described in Appendix~\ref{details:trainingDetails}, the fine-tuned foundation models were fit to our own labels after re-estimation of their atomic reference energies, and are evaluated against the same reference they were fit to reproduce; only one level of theory is involved in those comparisons.

\begin{figure}[p]
\centering
\gmdheader
\gmdrow{Chain Extension}{longer chains in test}
  {\gmdmol{\Malkane{4}}{butane}{C$_4$H$_{10}$}}
  {\gmdmol{\Malkane{9}}{nonane}{C$_9$H$_{20}$}}
\gmdrow{Fusion}{both groups on one carbon}
  {\gmdmol{\Malcohol{5}}{pentan-1-ol}{C$_5$H$_{11}$OH}\hspace{0.6em}\gmdmol{\Maldehyde{5}}{pentanal}{C$_4$H$_9$CHO}}
  {\gmdmol{\Macid{5}}{pentanoic acid}{C$_4$H$_9$COOH}}
\gmdrow{Duplication}{the group appears twice}
  {\gmdmol{\Macid{5}}{pentanoic acid}{C$_4$H$_9$COOH}}
  {\gmdmol{\Mdiacid{5}}{pentanedioic acid}{HOOC(CH$_2$)$_3$COOH}}
\gmdrow{Combination}{different groups at the ends}
  {\begin{tabular}{@{}c@{}}\gmdmol{\Mdiamine{5}}{pentane-1,5-diamine}{H$_2$N(CH$_2$)$_5$NH$_2$}\\[2pt]
   \gmdmol{\Mdiacid{5}}{pentanedioic acid}{HOOC(CH$_2$)$_3$COOH}\end{tabular}}
  {\gmdmol{\Maminoacid{5}}{5-aminopentanoic acid}{H$_2$N(CH$_2$)$_4$COOH}}
\gmdrow{Fission}{each group on its own}
  {\gmdmol{\Macid{5}}{pentanoic acid}{C$_4$H$_9$COOH}}
  {\gmdmol{\Malcohol{5}}{pentan-1-ol}{C$_5$H$_{11}$OH}\hspace{0.6em}\gmdmol{\Maldehyde{5}}{pentanal}{C$_4$H$_9$CHO}}
\vspace{1pt}\gmdlegend
\caption{Representative molecules for each task of the GMD benchmark, drawn as skeletal formulas. For each task, the left panel shows example training molecules and the right panel an example out-of-distribution test molecule. Fragment Chain Extension: butane in training, nonane at test time. Fragment Fusion: pentan-1-ol and pentanal in training, pentanoic acid at test time, which carries both the hydroxyl and the carbonyl group on one carbon. Fragment Duplication: pentanoic acid in training, pentanedioic acid at test time. Fragment Combination: pentane-1,5-diamine and pentanedioic acid in training, 5-aminopentanoic acid at test time. Fragment Fission: pentanoic acid in training, pentan-1-ol and pentanal at test time. Chain lengths are illustrative; Table~\ref{tab:splitting_strategy_multirow} gives the full ranges and Figure~\ref{fig:dataset_molecules} the complete list of molecules.}
\label{fig:molecules_tasks}
\end{figure}

\begin{figure}[p]
\centering
\gmdheader
\gmdrow{Chain Extension (aug.)}{lengths swapped between families}
  {\gmdmol{\Malcohol{3}}{propan-1-ol}{C$_3$H$_7$OH}\hspace{0.8em}\gmdmol{\Macid{5}}{pentanoic acid}{C$_4$H$_9$COOH}}
  {\gmdmol{\Malcohol{5}}{pentan-1-ol}{C$_5$H$_{11}$OH}\hspace{0.8em}\gmdmol{\Macid{3}}{propanoic acid}{C$_2$H$_5$COOH}}
\gmdrow{Fusion (aug.)}{amides as worked examples}
  {\begin{tabular}{@{}cc@{}}\gmdmol{\Malcohol{5}}{pentan-1-ol}{C$_5$H$_{11}$OH} & \gmdmol{\Maldehyde{5}}{pentanal}{C$_4$H$_9$CHO}\\[2pt]
   \gmdmol{\Mamine{5}}{pentan-1-amine}{C$_5$H$_{11}$NH$_2$} & \gmdmol{\Mamide{5}}{pentanamide}{C$_4$H$_9$CONH$_2$}\end{tabular}}
  {\gmdmol{\Macid{5}}{pentanoic acid}{C$_4$H$_9$COOH}}
\vspace{1pt}\gmdlegend
\caption{Representative molecules for the two augmented task variants, drawn as in Figure~\ref{fig:molecules_tasks}. In augmented Fragment Chain Extension, alcohols and carboxylic acids are trained on disjoint ranges of chain length, so every test chain length appears during training, but on the other functional group. In augmented Fragment Fusion, the training set additionally contains amines and amides; the amide group can itself be viewed as a fusion of a carbonyl and an amine group, giving an explicit worked example of fusion, while the test set remains the carboxylic acids.}
\label{fig:molecules_aug}
\end{figure}

\begin{figure}[p]
\centering
\begin{tikzpicture}[
  font=\scriptsize,
  every node/.style={inner xsep=3pt, inner ysep=1.6pt, outer sep=0pt},
  head/.style={draw=#1!60!black, fill=#1, minimum height=4mm,
               rounded corners=1.2pt, line width=0.4pt},
  body/.style={draw=#1!60!black, line width=0.4pt, inner ysep=1.2pt},
]
\renewcommand{\arraystretch}{0.96}
\node[head=hc, text width=5.82cm, anchor=north west] (h1) at (0.000cm,0)
  {\textbf{Alkanes}\hfill 12\,$\cdot$\,C$_{2}$--C$_{13}$};
\node[body=hc, text width=5.82cm, anchor=north west] (b1) at (h1.south west) {%
  \begin{tabular}{@{}>{\raggedright\arraybackslash}p{2.42cm}@{}>{\raggedright\arraybackslash}p{1.18cm}@{}>{\raggedright\arraybackslash\tiny\ttfamily\bfseries}p{2.22cm}@{}}
  Ethane & \ce{C2H6} & CC \\
  Propane & \ce{C3H8} & CCC \\
  Butane & \ce{C4H10} & CCCC \\
  Pentane & \ce{C5H12} & CCCCC \\
  Hexane & \ce{C6H14} & CCCCCC \\
  Heptane & \ce{C7H16} & CCCCCCC \\
  Octane & \ce{C8H18} & CCCCCCCC \\
  Nonane & \ce{C9H20} & CCCCCCCCC \\
  Decane & \ce{C10H22} & CCCCCCCCCC \\
  Undecane & \ce{C11H24} & CCCCCCCCCCC \\
  Dodecane & \ce{C12H26} & CCCCCCCCCCCC \\
  Tridecane & \ce{C13H28} & CCCCCCCCCCCCC \\
  \end{tabular}};
\node[head=ox, text width=5.82cm, anchor=north west] (h2) at ($(b1.south west)+(0,-2.2mm)$)
  {\textbf{Primary alcohols}\hfill 14\,$\cdot$\,C$_{2}$--C$_{15}$};
\node[body=ox, text width=5.82cm, anchor=north west] (b2) at (h2.south west) {%
  \begin{tabular}{@{}>{\raggedright\arraybackslash}p{2.42cm}@{}>{\raggedright\arraybackslash}p{1.18cm}@{}>{\raggedright\arraybackslash\tiny\ttfamily\bfseries}p{2.22cm}@{}}
  Ethanol & \ce{C2H6O} & CCO \\
  Propan-1-ol & \ce{C3H8O} & CCCO \\
  Butan-1-ol & \ce{C4H10O} & CCCCO \\
  Pentan-1-ol & \ce{C5H12O} & CCCCCO \\
  Hexan-1-ol & \ce{C6H14O} & CCCCCCO \\
  Heptan-1-ol & \ce{C7H16O} & CCCCCCCO \\
  Octan-1-ol & \ce{C8H18O} & CCCCCCCCO \\
  Nonan-1-ol & \ce{C9H20O} & CCCCCCCCCO \\
  Decan-1-ol & \ce{C10H22O} & CCCCCCCCCCO \\
  Undecan-1-ol & \ce{C11H24O} & CCCCCCCCCCCO \\
  Dodecan-1-ol & \ce{C12H26O} & CCCCCCCCCCCCO \\
  Tridecan-1-ol & \ce{C13H28O} & CCCCCCCCCCCCCO \\
  Tetradecan-1-ol & \ce{C14H30O} & CCCCCCCCCCCCCCO \\
  Pentadecan-1-ol & \ce{C15H32O} & CCCCCCCCCCCCCCCO \\
  \end{tabular}};
\node[head=ox, text width=5.82cm, anchor=north west] (h3) at ($(b2.south west)+(0,-2.2mm)$)
  {\textbf{Aldehydes}\hfill 14\,$\cdot$\,C$_{2}$--C$_{15}$};
\node[body=ox, text width=5.82cm, anchor=north west] (b3) at (h3.south west) {%
  \begin{tabular}{@{}>{\raggedright\arraybackslash}p{2.42cm}@{}>{\raggedright\arraybackslash}p{1.18cm}@{}>{\raggedright\arraybackslash\tiny\ttfamily\bfseries}p{2.22cm}@{}}
  Ethanal & \ce{C2H4O} & CC=O \\
  Propanal & \ce{C3H6O} & CCC=O \\
  Butanal & \ce{C4H8O} & CCCC=O \\
  Pentanal & \ce{C5H10O} & CCCCC=O \\
  Hexanal & \ce{C6H12O} & CCCCCC=O \\
  Heptanal & \ce{C7H14O} & CCCCCCC=O \\
  Octanal & \ce{C8H16O} & CCCCCCCC=O \\
  Nonanal & \ce{C9H18O} & CCCCCCCCC=O \\
  Decanal & \ce{C10H20O} & CCCCCCCCCC=O \\
  Undecanal & \ce{C11H22O} & CCCCCCCCCCC=O \\
  Dodecanal & \ce{C12H24O} & CCCCCCCCCCCC=O \\
  Tridecanal & \ce{C13H26O} & CCCCCCCCCCCCC=O \\
  Tetradecanal & \ce{C14H28O} & CCCCCCCCCCCCCC=O \\
  Pentadecanal & \ce{C15H30O} & CCCCCCCCCCCCCCC=O \\
  \end{tabular}};
\node[head=ox, text width=5.82cm, anchor=north west] (h4) at ($(b3.south west)+(0,-2.2mm)$)
  {\textbf{Carboxylic acids}\hfill 14\,$\cdot$\,C$_{2}$--C$_{15}$};
\node[body=ox, text width=5.82cm, anchor=north west] (b4) at (h4.south west) {%
  \begin{tabular}{@{}>{\raggedright\arraybackslash}p{2.42cm}@{}>{\raggedright\arraybackslash}p{1.18cm}@{}>{\raggedright\arraybackslash\tiny\ttfamily\bfseries}p{2.22cm}@{}}
  Ethanoic acid & \ce{C2H4O2} & CC(=O)O \\
  Propanoic acid & \ce{C3H6O2} & CCC(=O)O \\
  Butanoic acid & \ce{C4H8O2} & CCCC(=O)O \\
  Pentanoic acid & \ce{C5H10O2} & CCCCC(=O)O \\
  Hexanoic acid & \ce{C6H12O2} & CCCCCC(=O)O \\
  Heptanoic acid & \ce{C7H14O2} & CCCCCCC(=O)O \\
  Octanoic acid & \ce{C8H16O2} & CCCCCCCC(=O)O \\
  Nonanoic acid & \ce{C9H18O2} & CCCCCCCCC(=O)O \\
  Decanoic acid & \ce{C10H20O2} & CCCCCCCCCC(=O)O \\
  Undecanoic acid & \ce{C11H22O2} & CCCCCCCCCCC(=O)O \\
  Dodecanoic acid & \ce{C12H24O2} & CCCCCCCCCCCC(=O)O \\
  Tridecanoic acid & \ce{C13H26O2} & CCCCCCCCCCCCC(=O)O \\
  Tetradecanoic acid & \ce{C14H28O2} & CCCCCCCCCCCCCC(=O)O \\
  Pentadecanoic acid & \ce{C15H30O2} & CCCCCCCCCCCCCCC(=O)O \\
  \end{tabular}};
\node[head=ni, text width=5.82cm, anchor=north west] (h5) at ($(b4.south west)+(0,-2.2mm)$)
  {\textbf{Primary amines}\hfill 6\,$\cdot$\,C$_{2}$, C$_{4}$--C$_{8}$};
\node[body=ni, text width=5.82cm, anchor=north west] (b5) at (h5.south west) {%
  \begin{tabular}{@{}>{\raggedright\arraybackslash}p{2.42cm}@{}>{\raggedright\arraybackslash}p{1.18cm}@{}>{\raggedright\arraybackslash\tiny\ttfamily\bfseries}p{2.22cm}@{}}
  Ethanamine & \ce{C2H7N} & CCN \\
  Butan-1-amine & \ce{C4H11N} & CCCCN \\
  Pentan-1-amine & \ce{C5H13N} & CCCCCN \\
  Hexan-1-amine & \ce{C6H15N} & CCCCCCN \\
  Heptan-1-amine & \ce{C7H17N} & CCCCCCCN \\
  Octan-1-amine & \ce{C8H19N} & CCCCCCCCN \\
  \end{tabular}};
\node[head=no, text width=7.28cm, anchor=north west] (h6) at (6.232cm,0)
  {\textbf{Primary amides}\hfill 5\,$\cdot$\,C$_{4}$--C$_{8}$};
\node[body=no, text width=7.28cm, anchor=north west] (b6) at (h6.south west) {%
  \begin{tabular}{@{}>{\raggedright\arraybackslash}p{3.17cm}@{}>{\raggedright\arraybackslash}p{1.38cm}@{}>{\raggedright\arraybackslash\tiny\ttfamily\bfseries}p{2.73cm}@{}}
  Butanamide & \ce{C4H9NO} & CCCC(N)=O \\
  Pentanamide & \ce{C5H11NO} & CCCCC(N)=O \\
  Hexanamide & \ce{C6H13NO} & CCCCCC(N)=O \\
  Heptanamide & \ce{C7H15NO} & CCCCCCC(N)=O \\
  Octanamide & \ce{C8H17NO} & CCCCCCCC(N)=O \\
  \end{tabular}};
\node[head=ni, text width=7.28cm, anchor=north west] (h7) at ($(b6.south west)+(0,-2.2mm)$)
  {\textbf{Diamines}\hfill 10\,$\cdot$\,C$_{2}$--C$_{11}$};
\node[body=ni, text width=7.28cm, anchor=north west] (b7) at (h7.south west) {%
  \begin{tabular}{@{}>{\raggedright\arraybackslash}p{3.17cm}@{}>{\raggedright\arraybackslash}p{1.38cm}@{}>{\raggedright\arraybackslash\tiny\ttfamily\bfseries}p{2.73cm}@{}}
  Ethane-1,2-diamine & \ce{C2H8N2} & NCCN \\
  Propane-1,3-diamine & \ce{C3H10N2} & NCCCN \\
  Butane-1,4-diamine & \ce{C4H12N2} & NCCCCN \\
  Pentane-1,5-diamine & \ce{C5H14N2} & NCCCCCN \\
  Hexane-1,6-diamine & \ce{C6H16N2} & NCCCCCCN \\
  Heptane-1,7-diamine & \ce{C7H18N2} & NCCCCCCCN \\
  Octane-1,8-diamine & \ce{C8H20N2} & NCCCCCCCCN \\
  Nonane-1,9-diamine & \ce{C9H22N2} & NCCCCCCCCCN \\
  Decane-1,10-diamine & \ce{C10H24N2} & NCCCCCCCCCCN \\
  Undecane-1,11-diamine & \ce{C11H26N2} & NCCCCCCCCCCCN \\
  \end{tabular}};
\node[head=ox, text width=7.28cm, anchor=north west] (h8) at ($(b7.south west)+(0,-2.2mm)$)
  {\textbf{Dicarboxylic acids}\hfill 15\,$\cdot$\,C$_{2}$--C$_{16}$};
\node[body=ox, text width=7.28cm, anchor=north west] (b8) at (h8.south west) {%
  \begin{tabular}{@{}>{\raggedright\arraybackslash}p{3.17cm}@{}>{\raggedright\arraybackslash}p{1.38cm}@{}>{\raggedright\arraybackslash\tiny\ttfamily\bfseries}p{2.73cm}@{}}
  Ethanedioic acid & \ce{C2H2O4} & O=C(O)C(=O)O \\
  Propanedioic acid & \ce{C3H4O4} & O=C(O)CC(=O)O \\
  Butanedioic acid & \ce{C4H6O4} & O=C(O)CCC(=O)O \\
  Pentanedioic acid & \ce{C5H8O4} & O=C(O)CCCC(=O)O \\
  Hexanedioic acid & \ce{C6H10O4} & O=C(O)CCCCC(=O)O \\
  Heptanedioic acid & \ce{C7H12O4} & O=C(O)CCCCCC(=O)O \\
  Octanedioic acid & \ce{C8H14O4} & O=C(O)CCCCCCC(=O)O \\
  Nonanedioic acid & \ce{C9H16O4} & O=C(O)CCCCCCCC(=O)O \\
  Decanedioic acid & \ce{C10H18O4} & O=C(O)CCCCCCCCC(=O)O \\
  Undecanedioic acid & \ce{C11H20O4} & O=C(O)CCCCCCCCCC(=O)O \\
  Dodecanedioic acid & \ce{C12H22O4} & O=C(O)CCCCCCCCCCC(=O)O \\
  Tridecanedioic acid & \ce{C13H24O4} & O=C(O)CCCCCCCCCCCC(=O)O \\
  Tetradecanedioic acid & \ce{C14H26O4} & O=C(O)CCCCCCCCCCCCC(=O)O \\
  Pentadecanedioic acid & \ce{C15H28O4} & O=C(O)CCCCCCCCCCCCCC(=O)O \\
  Hexadecanedioic acid & \ce{C16H30O4} & O=C(O)CCCCCCCCCCCCCCC(=O)O \\
  \end{tabular}};
\node[head=no, text width=7.28cm, anchor=north west] (h9) at ($(b8.south west)+(0,-2.2mm)$)
  {\textbf{Amino acids}\hfill 9\,$\cdot$\,C$_{2}$--C$_{10}$};
\node[body=no, text width=7.28cm, anchor=north west] (b9) at (h9.south west) {%
  \begin{tabular}{@{}>{\raggedright\arraybackslash}p{3.17cm}@{}>{\raggedright\arraybackslash}p{1.38cm}@{}>{\raggedright\arraybackslash\tiny\ttfamily\bfseries}p{2.73cm}@{}}
  2-Aminoethanoic acid & \ce{C2H5NO2} & NCC(=O)O \\
  3-Aminopropanoic acid & \ce{C3H7NO2} & NCCC(=O)O \\
  4-Aminobutanoic acid & \ce{C4H9NO2} & NCCCC(=O)O \\
  5-Aminopentanoic acid & \ce{C5H11NO2} & NCCCCC(=O)O \\
  6-Aminohexanoic acid & \ce{C6H13NO2} & NCCCCCC(=O)O \\
  7-Aminoheptanoic acid & \ce{C7H15NO2} & NCCCCCCC(=O)O \\
  8-Aminooctanoic acid & \ce{C8H17NO2} & NCCCCCCCC(=O)O \\
  9-Aminononanoic acid & \ce{C9H19NO2} & NCCCCCCCCC(=O)O \\
  10-Aminodecanoic acid & \ce{C10H21NO2} & NCCCCCCCCCC(=O)O \\
  \end{tabular}};
\node[head=ox, text width=7.28cm, anchor=north west] (h10) at ($(b9.south west)+(0,-2.2mm)$)
  {\textbf{Complex multifunctional}\hfill 19\,$\cdot$\,C$_{7}$--C$_{12}$};
\node[body=ox, text width=7.28cm, anchor=north west] (b10) at (h10.south west) {%
  \begin{tabular}{@{}>{\raggedright\arraybackslash}p{3.17cm}@{}>{\raggedright\arraybackslash}p{1.38cm}@{}>{\raggedright\arraybackslash\tiny\ttfamily\bfseries}p{2.73cm}@{}}
  Heptane-1,7-diol & \ce{C7H16O2} & OCCCCCCCO \\
  Octane-1,8-diol & \ce{C8H18O2} & OCCCCCCCCO \\
  Nonane-1,9-diol & \ce{C9H20O2} & OCCCCCCCCCO \\
  Decane-3,9-diol & \ce{C10H22O2} & CCC(O)CCCCCC(C)O \\
  Decane-4,7-diol & \ce{C10H22O2} & CCCC(O)CCC(O)CCC \\
  Decane-1,10-diol & \ce{C10H22O2} & OCCCCCCCCCCO \\
  Undecane-1,11-diol & \ce{C11H24O2} & OCCCCCCCCCCCO \\
  Dodecane-1,6,9-triol & \ce{C12H26O3} & CCCC(O)CCC(O)CCCCCO \\
  Dodecane-1,6,11-triol & \ce{C12H26O3} & CC(O)CCCCC(O)CCCCCO \\
  Dodecane-1,4,7,10-tetraol & \ce{C12H26O4} & CCC(O)CCC(O)CCC(O)CCCO \\
  Dodecane-1,4,10,12-tetraol & \ce{C12H26O4} & OCCCC(O)CCCCCC(O)CCO \\
  Undecane-4,9-dione & \ce{C11H20O2} & CCCC(=O)CCCCC(=O)CC \\
  Dodecanedial & \ce{C12H22O2} & O=CCCCCCCCCCCC=O \\
  3,4-Dioxodecanal & \ce{C10H16O3} & CCCCCCC(=O)C(=O)CC=O \\
  3,7-Dioxononanedial & \ce{C9H12O4} & O=CCC(=O)CCCC(=O)CC=O \\
  3,8-Dioxodecanedial & \ce{C10H14O4} & O=CCC(=O)CCCCC(=O)CC=O \\
  9-Oxoundecanedial & \ce{C11H18O3} & O=CCCCCCCCC(=O)CC=O \\
  3,4,7-Trioxononanedial & \ce{C9H10O5} & O=CCC(=O)CCC(=O)C(=O)CC=O \\
  3,4,8-Trioxodecanedial & \ce{C10H12O5} & O=CCC(=O)CCCC(=O)C(=O)CC=O \\
  \end{tabular}};
\node[anchor=north, font=\scriptsize] at ($(current bounding box.south)+(0,-2.5mm)$) {%
  Header colour = heteroatoms present:\quad
  \tikz\node[fill=hc, draw=hc!60!black, minimum width=3mm, minimum height=2.4mm]{}; none\quad
  \tikz\node[fill=ox, draw=ox!60!black, minimum width=3mm, minimum height=2.4mm]{}; O\quad
  \tikz\node[fill=ni, draw=ni!60!black, minimum width=3mm, minimum height=2.4mm]{}; N\quad
  \tikz\node[fill=no, draw=no!60!black, minimum width=3mm, minimum height=2.4mm]{}; N and O};
\end{tikzpicture}
\caption{All 118 molecules of the GMD dataset with their molecular formulas and canonical SMILES, grouped by functional group. Each panel header gives the group name, the number of molecules, and the range of carbon chain lengths; its colour indicates which heteroatoms the group contains.}
\label{fig:dataset_molecules}
\end{figure}




\clearpage
\newpage
\section*{NeurIPS Paper Checklist}

\begin{enumerate}

\item {\bf Claims}
    \item[] Question: Do the main claims made in the abstract and introduction accurately reflect the paper's contributions and scope?
    \item[] Answer: \answerYes{} 
    \item[] Justification: The abstract and introduction state that GMD is a benchmark of five compositional-generalisation tasks for MLIPs, and that current state-of-the-art models, including fine-tuned foundation models, exhibit large in- to out-of-distribution gaps on four of the five tasks; these claims are substantiated by the empirical analysis in Section~\ref{sec:evaluation} and supporting appendices.

\item {\bf Limitations}
    \item[] Question: Does the paper discuss the limitations of the work performed by the authors?
    \item[] Answer: \answerYes{} 
    \item[] Justification: Limitations of the benchmark are discussed in the Conclusions (Section~\ref{sec:conclusions})

\item {\bf Theory assumptions and proofs}
    \item[] Question: For each theoretical result, does the paper provide the full set of assumptions and a complete (and correct) proof?
    \item[] Answer: \answerNA{} 
    \item[] Justification: The paper does not include theoretical results: it introduces a benchmark dataset and evaluates existing architectures empirically.

    \item {\bf Experimental result reproducibility}
    \item[] Question: Does the paper fully disclose all the information needed to reproduce the main experimental results of the paper to the extent that it affects the main claims and/or conclusions of the paper (regardless of whether the code and data are provided or not)?
    \item[] Answer: \answerYes{} 
    \item[] Justification: Training and fine-tuning details (optimiser settings, fine-tuning budget, train/validation split protocol) are reported in Section~\ref{section:experimental_setup} and Appendix~\ref{details:trainingDetails}; the per-architecture hyperparameters selected via Bayesian optimisation are listed in Appendix~\ref{details:hyperparameters}; and the dataset generation pipeline (FlashMD sampling, ORCA(PBE/def2-TZVP+D3BJ) relabelling) is fully specified in Appendix~\ref{app:toolkit}.

\item {\bf Open access to data and code}
    \item[] Question: Does the paper provide open access to the data and code, with sufficient instructions to faithfully reproduce the main experimental results, as described in supplemental material?
    \item[] Answer:  \answerYes{} 
    \item[] Justification: The GMD dataset and the toolkit used to generate it are released under the MIT licence, with the toolkit and dataset available at the URLs given in Appendix~\ref{app:toolkit}. The dataset is provided as both raw labelled trajectories and pre-computed task-specific splits, enabling reproduction of the evaluation protocol without re-running upstream data generation; fine-tuned model weights are also released.

\item {\bf Experimental setting/details}
    \item[] Question: Does the paper specify all the training and test details (e.g., data splits, hyperparameters, how they were chosen, type of optimizer) necessary to understand the results?
    \item[] Answer: \answerYes{} 
    \item[] Justification: Data splits, training protocols, and the rationale for the foundation-model fine-tuning budget are documented in Section~\ref{section:experimental_setup} and Appendix~\ref{details:trainingDetails}; the hyperparameter search domains and selected values are tabulated in Appendix~\ref{details:hyperparameters}.

\item {\bf Experiment statistical significance}
    \item[] Question: Does the paper report error bars suitably and correctly defined or other appropriate information about the statistical significance of the experiments?
    \item[] Answer: \answerYes{} 
    \item[] Justification: All reported metrics are averaged over three independent validation/test splits drawn from held-out trajectories, and the corresponding per-split standard deviations are tabulated in Appendix~\ref{details:numerical_results}; across almost all (model, task, metric) cells, these standard deviations are substantially smaller than the in- to out-of-distribution gaps that support the main claims of the paper.

\item {\bf Experiments compute resources}
    \item[] Question: For each experiment, does the paper provide sufficient information on the computer resources (type of compute workers, memory, time of execution) needed to reproduce the experiments?
    \item[] Answer: \answerYes{} 
    \item[] Justification: Hardware (NVIDIA GH200, RTX Ada A6000, RTX A100, RTX 5090, RTX 4090 GPUs, and CPU resources for DFT relabelling) and approximate per-model training times are reported in Appendix~\ref{details:trainingDetails} (Table~\ref{tab:estimated_training_times}).
    
\item {\bf Code of ethics}
    \item[] Question: Does the research conducted in the paper conform, in every respect, with the NeurIPS Code of Ethics \url{https://neurips.cc/public/EthicsGuidelines}?
    \item[] Answer: \answerYes{} 
    \item[] Justification: The research conforms with the NeurIPS Code of Ethics: it concerns benchmarking of machine learning models on synthetic computational-chemistry data, does not involve human subjects or personal data, and presents no foreseeable ethical concerns.

\item {\bf Broader impacts}
    \item[] Question: Does the paper discuss both potential positive societal impacts and negative societal impacts of the work performed?
    \item[] Answer: \answerNA{} 
    \item[] Justification: The paper introduces a diagnostic benchmark for evaluating the compositional-generalisation capabilities of MLIPs; its contribution is methodological and we do not foresee a direct path to either positive or negative societal impact beyond improved scientific evaluation practice in computational chemistry.
    
\item {\bf Safeguards}
    \item[] Question: Does the paper describe safeguards that have been put in place for responsible release of data or models that have a high risk for misuse (e.g., pre-trained language models, image generators, or scraped datasets)?
    \item[] Answer: \answerNA{} 
    \item[] Justification: The released assets are a benchmark dataset of computational-chemistry trajectories and an associated toolkit; they pose no foreseeable misuse risk.

\item {\bf Licenses for existing assets}
    \item[] Question: Are the creators or original owners of assets (e.g., code, data, models), used in the paper, properly credited and are the license and terms of use explicitly mentioned and properly respected?
    \item[] Answer: \answerYes{} 
    \item[] Justification: Existing software assets used in this work — ASE~\cite{larsen2017atomic}, RDKit, FlashMD~\cite{bigi2025flashmd}, ORCA~\cite{ORCA}, fairchem-core, mace-torch, and the official NequIP implementation — are cited at point of use, and pre-trained foundation-model checkpoints (MACE-MP-0~\cite{batatia2025foundation}, UMA~\cite{wood2025family}) are similarly credited. All assets were used in accordance with their licences: ORCA under its free academic licence, the UMA checkpoint under its model licence, and the remaining software under their respective open-source licences.

\item {\bf New assets}
    \item[] Question: Are new assets introduced in the paper well documented and is the documentation provided alongside the assets?
    \item[] Answer: \answerYes{} 
    \item[] Justification: The GMD dataset and accompanying toolkit are documented in Appendix~\ref{app:toolkit}, including the released file formats and instructions for extending the benchmark to new molecular families. Both are released under the MIT licence at the URLs given there.   

\item {\bf Crowdsourcing and research with human subjects}
    \item[] Question: For crowdsourcing experiments and research with human subjects, does the paper include the full text of instructions given to participants and screenshots, if applicable, as well as details about compensation (if any)? 
    \item[] Answer: \answerNA{} 
    \item[] Justification: The paper does not involve crowdsourcing or research with human subjects.

\item {\bf Institutional review board (IRB) approvals or equivalent for research with human subjects}
    \item[] Question: Does the paper describe potential risks incurred by study participants, whether such risks were disclosed to the subjects, and whether Institutional Review Board (IRB) approvals (or an equivalent approval/review based on the requirements of your country or institution) were obtained?
    \item[] Answer: \answerNA{} 
    \item[] Justification: The paper does not involve crowdsourcing or research with human subjects.

\item {\bf Declaration of LLM usage}
    \item[] Question: Does the paper describe the usage of LLMs if it is an important, original, or non-standard component of the core methods in this research? Note that if the LLM is used only for writing, editing, or formatting purposes and does \emph{not} impact the core methodology, scientific rigor, or originality of the research, declaration is not required.
    \item[] Answer: \answerNA{} 
    \item[] Justification: LLMs were not used as a core methodological component of this research; any use was limited to writing assistance and did not affect the scientific rigour, originality, or methodology of the work.

\end{enumerate}

\end{document}